\documentclass[]{assets/matterlab}

\usepackage{microtype}
\usepackage{graphicx}
\usepackage{booktabs}
\usepackage{float}
\usepackage{xurl}
\usepackage{hyperref}
\usepackage{titlesec}
\usepackage{titletoc}

\usepackage{amsmath}
\usepackage{amssymb}
\usepackage{mathtools}
\usepackage{amsthm}
\usepackage{bm}
\usepackage{physics}

\usepackage[noabbrev,nameinlink]{cleveref}

\usepackage{amsmath,amsfonts,bm}

\def\eqref#1{equation~\ref{#1}}

\def\1{\bm{1}}

\DeclareMathAlphabet{\mathsfit}{\encodingdefault}{\sfdefault}{m}{sl}
\SetMathAlphabet{\mathsfit}{bold}{\encodingdefault}{\sfdefault}{bx}{n}

\usepackage{orcidlink}
\usepackage{comment}
\usepackage[version=4]{mhchem}
\usepackage{longtable}
\usepackage{array}
\renewcommand{\arraystretch}{1.7}
\usepackage{multirow}           
\usepackage{amsfonts}           
\usepackage{nicefrac}
\usepackage{duckuments}         
\usepackage{thmtools,thm-restate}
\usepackage{enumitem}
\usepackage[normalem]{ulem} 

\newcommand{\stdv}[1]{{\scriptsize\color{gray!70!black}\,$\pm$#1}}
\usepackage{xcolor,colortbl}
\usepackage{tikz}
\usepackage{tikz-cd}
\usepackage{caption}
\usepackage{subcaption}
\usepackage{listings}
\usepackage{gensymb}
\usepackage{textcomp} 
\usepackage[inkscapelatex=false]{svg}
\usepackage[nolist]{acronym}
\usepackage{cinzel}
\newcommand{\EAF}{\texorpdfstring{{\cinzel El~Agente~Forjador}}{El Agente Forjador}}
\newcommand{\Forjador}{\texorpdfstring{{\cinzel Forjador}}{Forjador}}

\newcommand{\addressCHEM}{Department of Chemistry, University of Toronto,  80 St. George St., Toronto, ON M5S 3H6, Canada}
\newcommand{\addressAC}{Acceleration Consortium, 700 University Ave., Toronto, ON M7A 2S4, Canada}
\newcommand{\addressCS}{Department of Computer Science, University of Toronto, 40 St George St., Toronto, ON M5S 2E4, Canada}
\newcommand{\addressVECTOR}{Vector Institute for Artificial Intelligence, W1140-108 College St., Schwartz Reisman Innovation Campus, Toronto, ON M5G 0C6, Canada}
\newcommand{\addressMSE}{Department of Materials Science \& Engineering, University of Toronto, 184 College St., Toronto, ON M5S 3E4, Canada}
\newcommand{\addressCHEMENG}{Department of Chemical Engineering \& Applied Chemistry, University of Toronto, 200 College St., Toronto, ON M5S 3E5, Canada}
\newcommand{\addressCIFAR}{Canadian Institute for Advanced Research (CIFAR), 661 University Ave., Toronto,
ON M5G 1M1, Canada}
\newcommand{\addressNVIDIA}{NVIDIA, 431 King St W \#6th, Toronto, ON M5V 1K4, Canada}

\newcommand{\acknowAC}{This research is part of the University of Toronto’s Acceleration Consortium, which receives funding from the CFREF-2022-00042 Canada First Research Excellence Fund.}

\title{\EAF: Task-driven Agent Generation for Quantum Simulation}

\author[1,6,8\dagger]{Zijian Zhang}
\author[1\dagger]{Aiwei Yin}
\author[1,2]{Amaan Baweja}
\author[2,6]{Jiaru Bai}
\author[2]{Ignacio Gustin}
\author[1,2,5,*]{Varinia~Bernales}
\author[1,2,3,4,5,6,7,8,*]{Al\'an Aspuru-Guzik}

\renewcommand{\cite}{\citep}
\tcbset{textmarker/.style={%
    enhanced,
    breakable,
    frame hidden, 
    parbox=false,
    boxrule=0mm,
    boxsep=0mm,
    arc=3mm,          
    outer arc=3mm,    
    left=3mm,right=3mm,
    top=7pt,bottom=7pt,
    toptitle=1mm,bottomtitle=1mm,
    left skip=0.2cm
}}

\tcbset{textmarker2/.style={%
    enhanced,
    breakable,
    frame hidden, 
    parbox=false,
    boxrule=0mm,
    boxsep=0mm,
    arc=3mm,          
    outer arc=3mm,    
    left=3mm,right=3mm,
    top=7pt,bottom=7pt,
    toptitle=1mm,bottomtitle=1mm,
    left skip=0.8cm
}}
\newtcolorbox{hintBox}{textmarker,
    colback=blue!10!white}

\newtcolorbox{hintBox2}{textmarker2,
    colback=blue!5!white}

\newtcolorbox{importantBox}{textmarker,
    colback=red!10!white}

\newtcolorbox{noteBox}{textmarker,
    colback=matterbg}
\newcommand{\prompt}[1]{\begin{noteBox} \underline{\textbf{Prompt}} #1 \end{noteBox}}

\newtcolorbox{questionBox}{textmarker,
    colback=violet!10!white}

\affiliation[1]{\addressCS}
\affiliation[2]{\addressCHEM}
\affiliation[3]{\addressMSE}
\affiliation[4]{\addressCHEMENG}
\affiliation[5]{\addressAC}
\affiliation[6]{\addressVECTOR}
\affiliation[7]{\addressCIFAR}
\affiliation[8]{\addressNVIDIA}

\contribution[\dagger]{These authors contributed equally to this work.}

\abstract{

AI for science promises to accelerate the discovery process. The advent of large language models (LLMs) and agentic workflows enables the expediting of a growing range of scientific tasks. However, most of the current generation of agentic systems depend on static, hand-curated toolsets that hinder adaptation to new domains and evolving libraries. We present \EAF, a multi-agent framework in which universal coding agents autonomously forge, validate, and reuse computational tools through a four-stage workflow of tool analysis, tool generation, task execution, and iterative solution evaluation. Evaluated across 24 tasks spanning quantum chemistry and quantum dynamics on five coding agent setups, we compare three operating modes: zero-shot generation of tools per task, reuse of a curriculum-built toolset, and direct problem-solving with the coding agents as the baseline. We find that our tool generation and reuse framework consistently improves accuracy over the baseline. We also show that reusing a toolset built by a stronger coding agent can reduce API cost and substantially raises the solution quality for weaker coding agents. Case studies further demonstrate that tools forged for different domains can be combined to solve hybrid tasks.
Taken together, these results show that LLM-based agents can use their scientific knowledge and coding capabilities to autonomously build reusable scientific tools, pointing toward a paradigm in which agent capabilities are defined by the tasks they are designed to solve rather than by explicitly engineered implementations.
}

\date{\today}
\correspondence{
Al\'an Aspuru-Guzik at \email{alan@aspuru.com} \\ Varinia Bernales at \email{varinia@bernales.org}}

\begin{document}


\acrodef{mcp}[MCP]{Model Context Protocol}
\acrodef{hpc}[HPC]{high-performance computing}

\acrodef{dft}[DFT]{density functional theory}
\acrodef{tddft}[TD-DFT]{time-dependent density functional theory}

\acrodef{vqe}[VQE]{variational quantum eigensolver}
\acrodef{dmrg}[DMRG]{density matrix renormalization group}
\acrodef{tddmrg}[TD-DMRG]{time-dependent density matrix renormalization group}
\acrodef{heom}[HEOM]{hierarchical equations of motion}
\acrodef{fmo}[FMO]{Fenna-Matthews-Olson}

\maketitle


\section{Introduction}

\begin{figure}
    \centering
\includegraphics[width=1\linewidth]{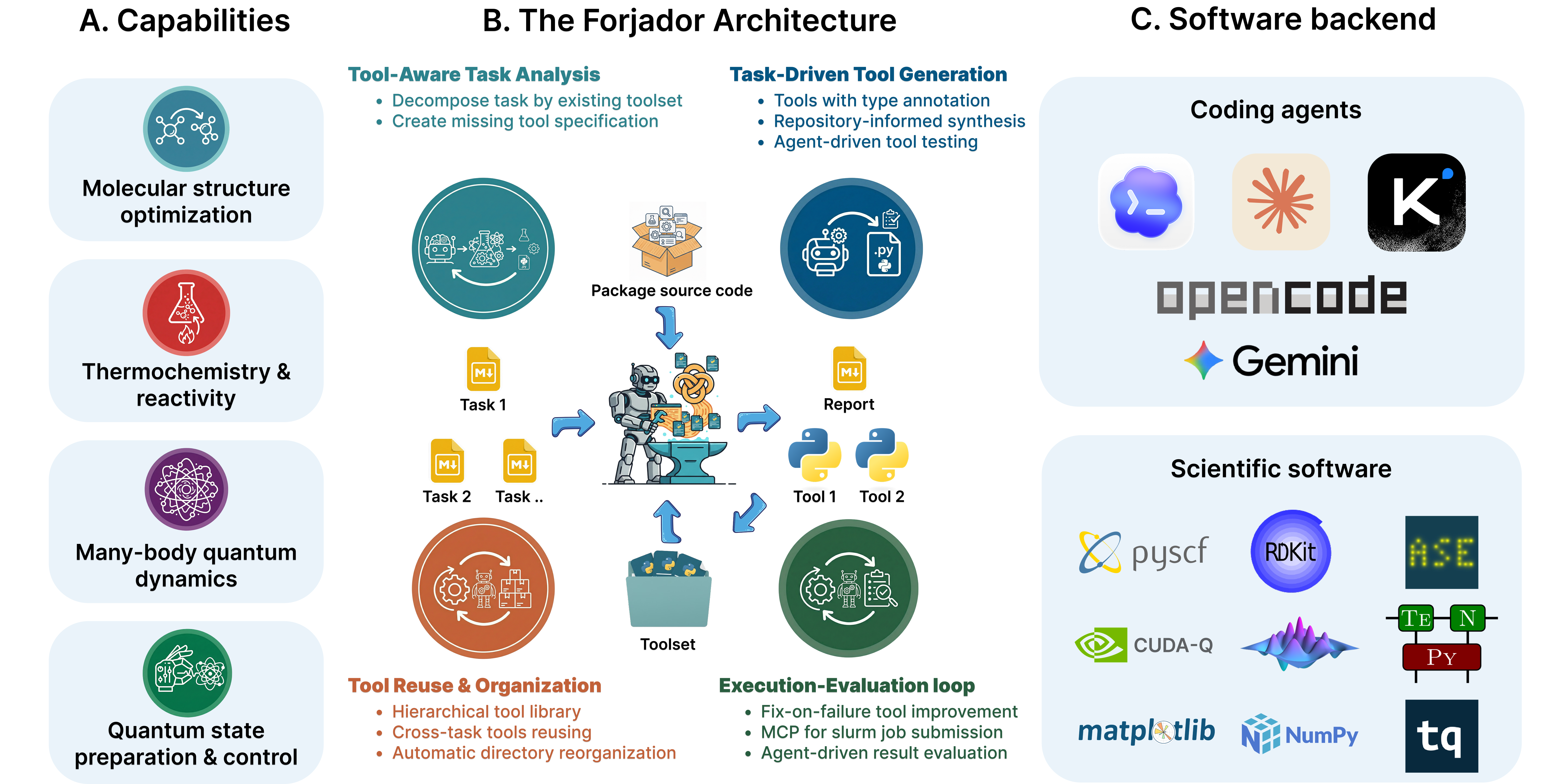}
    \caption{Overview of \EAF. \textbf{(A)}~We evaluate our system on quantum chemistry and quantum dynamics tasks, which span molecular structure optimization, thermochemistry \& reactivity, many-body quantum dynamics, and quantum state preparation \& control.
    \textbf{(B)}~The \Forjador{} workflow comprises four stages: \emph{tool analysis} (decompose tasks and search existing toolset), \emph{tool generation} (synthesize missing tools), \emph{task execution} (compose tools and run pipelines), and \emph{solution evaluation} (iteratively assess results). Tasks produce reports and reusable toolsets. \textbf{(C)}~Software backend: coding agents and scientific software used in this work.}
    \label{fig:overview}
\end{figure}

Automating scientific research has long been a central aspiration of the scientific community. Large language models (LLMs)~\cite{brown2020language,achiam2023gpt,jablonka2024leveraging,anthropic2024claude} and LLM-based agents~\cite{hong2023metagpt,wang2024survey, claudecode} have brought this vision closer to reality, demonstrating emergent physical and chemical reasoning~\cite{moreno2026ai, bran2023chemcrow}, autonomous scientific research capabilities~\cite{boiko2023emergent, lu2026end, perez2026quntur}, and sophisticated reasoning about experiment execution \cite{cao2025kagents,yoshikawa2023large, darvish2025organaroboticassistantautomated, song2025multiagent}. Building on these advances, domain-specific scientific agents now achieve strong performance on specialized tasks through expert-designed tools and  workflows~\cite{zou2025agente,wang2025dreams,pham2025chemgraph}. Our group has developed several agents of this type under the {\cinzel El Agente} umbrella which span quantum chemistry~\cite{zou2025agente,perez2026quntur, bai2026agente}, quantum dynamics simulations~\cite{gustin2025cuantico}, solid-state simulations~\cite{kumar2026solido}, and molecular structure manipulation and editing~\cite{choi2026estructural}. However, these systems require substantial human effort to design, configure, and maintain: each new domain or computational method necessitates significant engineering to adapt existing systems or build new ones from scratch.

The root cause is that traditional agent architectures treat their toolsets as static components. Tools must be manually implemented and integrated before deployment, and when new techniques emerge, software libraries evolve, or research questions shift to adjacent domains, human developers must intervene. This rigid paradigm creates a critical bottleneck: the pace of scientific agent development cannot keep up with the diversity and evolution of scientific research needs. What is needed is an architecture in which agents can autonomously generate, validate, organize, and compose tools, so that the toolset co-evolves with the research problems it serves.

We introduce \textbf{\EAF} (``The Forging Agent''), a multi-agent system that automates not only scientific problem-solving but also the generation of the specialized tools it requires (see Figure~\ref{fig:overview}). Rather than relying on a fixed toolkit, \EAF{} dynamically creates, tests, refines, and composes tools in response to task requirements, adapting to new challenges without human intervention in tool generation. The system is organized around a four-stage agentic workflow, namely (\emph{tool analysis}, \emph{tool generation}, \emph{task execution}, and \emph{solution evaluation}), which iterates until all task criteria are satisfied.

Our approach builds on the inherent power of coding agents~\cite{gao2025trae,codex,claudecode}, which, designed for code development, have several features that can aid scientific discovery. The framework introduces several key architectural choices. All subagents are \emph{universal}: every agent can generate scripts, execute computations, debug errors, and refine existing tools, enabling robust error recovery through mutual correction. 
A progressive disclosure mechanism and a dedicated toolset optimizer keep the growing library navigable by exposing only task-relevant tools and reorganizing tools into hierarchical categories. Finally, an automated solution evaluator drives iterative refinement by verifying bug-free execution, script completeness, simulation adequacy, and result completeness after each run.

We evaluate \EAF{} on two diverse benchmarks: (i)~13 undergraduate-level quantum chemistry tasks~\cite{zou2025agente, bai2026agente} covering molecular structure optimization and thermochemistry \& reactivity, and (ii)~11 quantum dynamics tasks~\cite{gustin2025cuantico} spanning many-body quantum dynamics and quantum state preparation \& control. 

\paragraph{Key Contributions.}

\begin{itemize}
    \item We present \EAF, a multi-agent framework that autonomously solves undergraduate-level quantum chemistry~\cite{zou2025agente, bai2026agente} and quantum dynamics~\cite{gustin2025cuantico} tasks by generating tools for these domains. We show that autonomous tool generation maintains or increases task accuracy relative to tool-free baselines, but incurs higher API cost due to the overhead of tool creation and refinement.
    \item We show that curriculum learning and tool reuse effectively mitigate this cost: reusing curriculum-built tools, underpinned by progressive disclosure, reduces API cost by 33\% to 78\% and wall-clock time by up to 88\% relative to zero-shot tool generation, while maintaining or improving accuracy. Because the toolset is built on a stronger model (Claude Opus~4.6) and reused by weaker ones, this constitutes a form of strong-to-weak knowledge transfer through tool interfaces: weaker models achieve up to 16.5 percentage points higher task success rate by composing well-tested tools that encode correct domain methodology.
    \item We demonstrate cross-domain composability through two case studies: tools originally forged for quantum chemistry and quantum dynamics are composed to solve hybrid quantum-chemistry/quantum-dynamics tasks. This points toward a future in which scientific agents with broad, multi-domain capabilities can be assembled by composing tools forged across disciplines by \EAF{}.
\end{itemize}

\section{Related Work}

\paragraph{Scientific Agents}

Recent benchmarks confirm that frontier LLMs already possess strong scientific knowledge. For example, Claude Opus~4.6~\cite{anthropic_claude_opus_46_2025} achieves 53.0\% accuracy on the Humanity's Last Exam (HLE)~\cite{phan2025humanity}, which is a collection of expert-level questions spanning science, mathematics, and engineering that were deemed very hard for LLMs to solve.
LLM-based scientific agents have been built on this foundation to automate scientific tasks. In quantum chemistry, {\cinzel El Agente Q}~\cite{zou2025agente} and {\cinzel El Agente Quntur}~\cite{perez2026quntur} use a hierarchical memory framework to decompose natural language prompts into executable simulation tasks. In quantum dynamics, {\cinzel El Agente Cu\'antico}~\cite{gustin2025cuantico} extends this paradigm to quantum simulation workflows spanning open-system dynamics and tensor-network methods. For solid-state simulations, {\cinzel El Agente S\'{o}lido}~\cite{kumar2026solido} translates high-level scientific objectives into Quantum ESPRESSO \cite{giannozzi2009quantum} pipelines integrating \ac{dft}, phonon calculations, and machine-learning potentials, while DREAMS~\cite{wang2025dreams} pairs a central LLM planner with specialized subagents for structure generation, \ac{dft} convergence testing, and \ac{hpc} scheduling. {\cinzel El Agente Gr\'{a}fico}~\cite{bai2026agente} takes a complementary approach by encoding scientific workflows as type-safe execution graphs backed by dynamic knowledge graphs, separating the execution layer from LLM reasoning to reduce token cost and hallucination. Beyond chemistry, \textit{k-agents}~\cite{cao2025kagents} automates quantum computing laboratory experiments, and \citet{lu2026end} demonstrates automation of machine learning research: the system generates research ideas, writes and executes experiments, and produces full papers that can pass peer-review by humans. Despite their success, most of these systems share a common constraint: their toolsets must be updated manually whenever new tasks, methods, or software versions arise.
\paragraph{Coding Agents}

Frontier LLMs have demonstrated strong software engineering capability, with mid-70\% resolution rates on SWE-bench Verified~\cite{jimenez2024swebench} (e.g., Claude Opus 4.6: 75.60\%, Gemini 3 Pro~\cite{google2026gemini31pro}: 69.60\%, GPT-5.2~\cite{singh2025openai}: 72.8\%, Kimi K2.5~\cite{team2026kimi}: 70.80\%). This capability has been composed into \emph{coding agents}---systems that pair an LLM with terminal access and file-system interaction to autonomously write, execute, and debug code~\cite{dong2025survey}. Multiple coding-agent frameworks have emerged: Claude Code~\cite{claudecode}, Codex~\cite{codex}, Cursor~\cite{cursor}, and OpenCode~\cite{opencode}.
Given the scientific knowledge of LLMs, coding agents are well-suited for scientific automation. They can reason about domain-specific computations while retaining their general-purpose coding and debugging capabilities.

\paragraph{Evolving Agents and Tool Creation}

A parallel line of research investigates agents that can extend their own capabilities by modifying themselves to achieve self-evolution. The self-evolving agent paradigm reframes agents as systems that iteratively update their memories, toolset, and architectures based on trajectory feedback to maximize long-term utility~\cite{gao2025survey}. Early tool-creation systems such as CRAFT and CREATOR used LLM prompting to synthesize new tools, with CRAFT discarding broken tools via execution feedback~\cite{yuan2023craft} and CREATOR enabling iterative refinement through manual execution~\cite{qian2023creator}. More recent systems, Alita~\cite{qiu2025alita} and ATLASS~\cite{haque2025advanced}, perform need-based tool generation with capability checking against an existing toolset, augmented by web search and code-execution feedback loops.

\section{Methods}
\label{sec:method}
\begin{figure}
    \centering
    \includegraphics[width=1\linewidth]{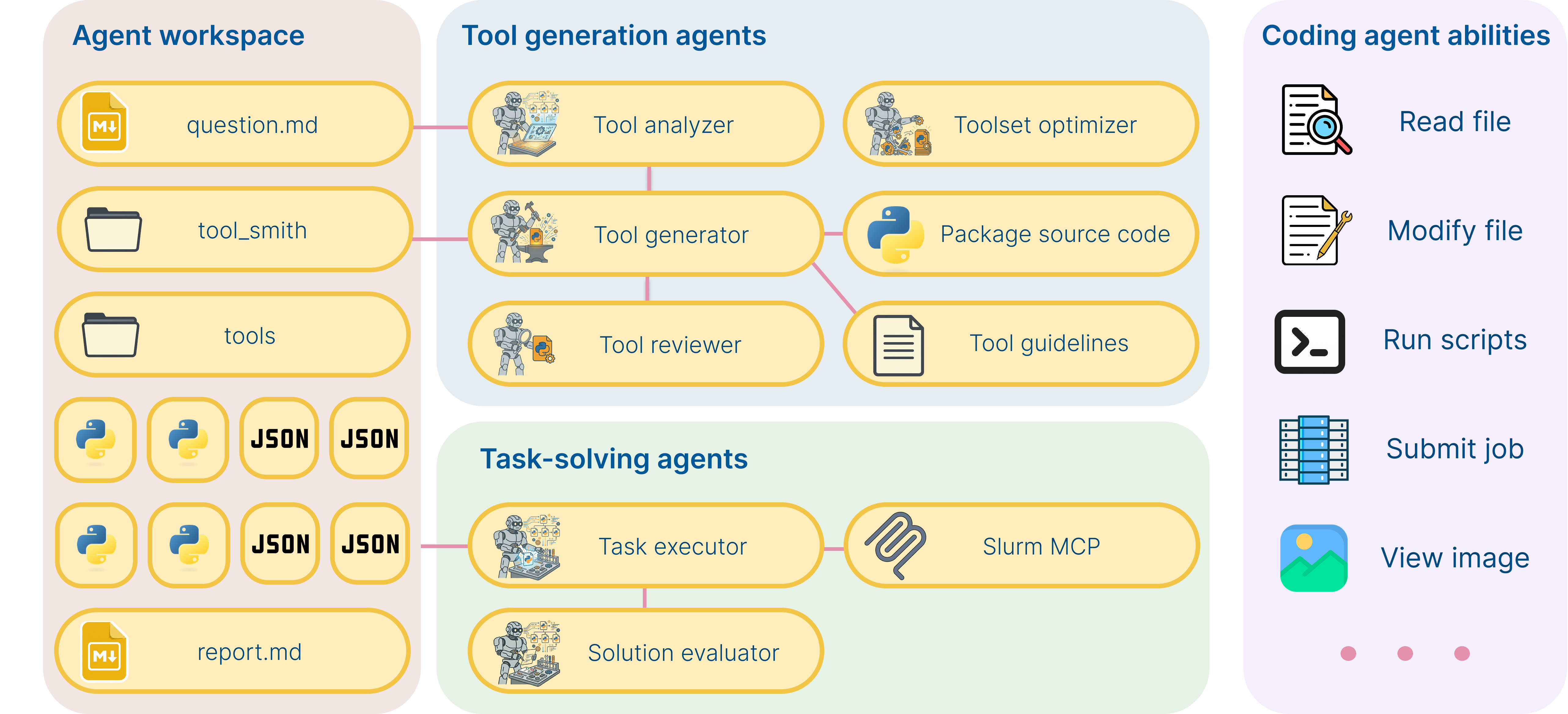}
    \caption{Detailed system architecture of \EAF. The \textbf{agent workspace} (left) maintains a structured directory containing the task specification (\texttt{question.md}), the \texttt{tool\_smith/} and \texttt{tools/} directories holding generated Python scripts and JSON metadata, and a \texttt{report.md} output file. \textbf{Tool generation agents} (centre top) comprise three roles: a \emph{tool analyzer} that assesses task requirements and existing tool coverage, a \emph{tool generator} that synthesizes new tools backed by package source code, and a \emph{tool reviewer} that reviews and refines tools for correctness and alignment; a \emph{toolset optimizer} refines tools. \textbf{Task-solving agents} (centre bottom) consist of a \emph{task executor} that composes tools into executable pipelines and dispatches jobs via an MCP server, and a \emph{solution evaluator} that validates outputs. \textbf{Coding agent abilities} (right) available to all agents include file read/modify, script execution, \ac{hpc} job submission, and image viewing.}
    \label{fig:system-architecture}
\end{figure}

\EAF{} is a task-driven agent framework in which each task is solved with a toolset that can be generated and modified one the fly to meet the requirement of the task (see Figure~\ref{fig:system-architecture}). Rather than relying on a fixed toolset, the system expands its capabilities on demand.  
The system operates through a four-stage workflow (tool analysis, tool generation, task execution, and solution evaluation). Supporting this workflow are a tool generator that creates, tests, and iteratively refines reusable Python functions, a toolset optimizer that maintains hierarchical organization of the growing toolset, a task executor that carries out script generation, execution, and report compilation, and a solution evaluator that assesses task completion and drives iterative refinement. 
We implement this workflow with general-purpose coding agents that share a common agent workspace (see Figure~\ref{fig:system-architecture}), enabling failures encountered in one stage to be repaired by later stages without human intervention.
We leverage state-of-the-art coding agents, specifically Claude Code~\cite{claudecode}, Codex~\cite{codex}, and OpenCode~\cite{opencode}, as the underlying building blocks of our system.

\subsection{The \Forjador{} workflow}

Each run of \EAF{} begins with a task specified in natural language in a \texttt{question.md} file in the agent workspace. When a new task is provided, \EAF{} executes a multi-stage workflow in which each stage spawns a fresh coding agent session in the agent workspace with a different prompt.

\begin{enumerate}
    \item \textbf{Tool analysis}: For each task, the agent surveys the existing toolset in the workspace \texttt{tools/} directory. It inspects available tools, reviews their source code, and determines whether they meet the task requirements. If gaps exist, the agent drafts a specification for the new tools that are needed.

    \item \textbf{Tool generation}: Given the specification of a new tool, the tool generator searches the source code of the scientific software required for the tool, implements the tool in Python, and writes test cases. It then iterates through a review--refinement loop until all tests pass. Details can be found in Section~\ref{sec:tool-generator}.

    \item \textbf{Task execution}: Once all necessary tools are available, the agent writes task-specific execution scripts that compose the tools from the toolset.
    Specifically, the agent drafts Python scripts that import the required tools, configure input parameters, and orchestrate the computational pipeline. It then executes these scripts, monitors for errors, and iteratively debugs until the pipeline completes. Finally, the agent compiles the results into a \texttt{report.md} file that addresses the task specification. Details are provided in Section~\ref{sec:task-executor}.
    
    \item \textbf{Solution evaluation}: After execution, an automated evaluator assesses whether the generated report satisfies all task requirements and determines whether additional iterations are needed. If any criterion is not met, the evaluator produces a structured next-step plan for the subsequent iteration. This loop repeats until all criteria are satisfied or the iteration budget is exhausted. The evaluation criteria and mechanism are detailed in Section~\ref{sec:solution-evaluator}.
\end{enumerate}

\begin{figure}
    \centering
    \includegraphics[width=0.9\linewidth]{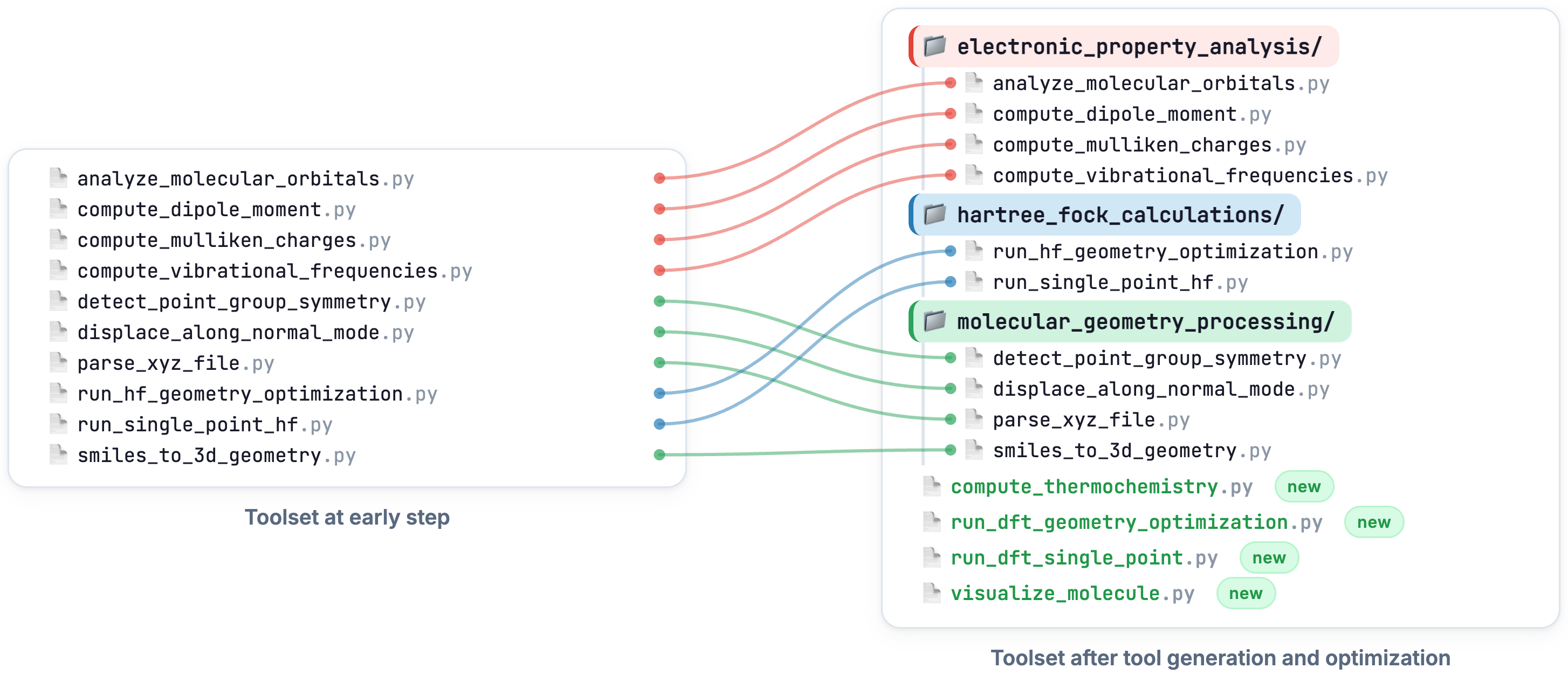}
    \caption{Toolset optimization and tool generation during a curriculum learning on quantum chemistry tasks. \textbf{Left:} A flat toolset at an early stage. \textbf{Right:} After optimization, existing tools (coloured lines) are reorganized into hierarchical subcategories based on functional similarity, and newly generated tools (green, labelled \emph{new}) are added at the root awaiting future reorganization.}
    \label{fig:toolset-optimization}
\end{figure}

\paragraph{Hierarchical Toolset with Progressive Disclosure.}
Each tool in \EAF{} is a Python file stored in a hierarchy of directories (see Figure~\ref{fig:toolset-optimization}). The hierarchical is built by the \emph{toolset optimizer}, which we will introduce in Section~\ref{sec:toolset-optimizer}. In this way, the agent is not given all existing tools upfront. Instead, it discovers tools through progressive disclosure within this hierarchy. For instance, an agent working on molecular geometry optimization navigates into relevant categories such as \texttt{molecular\_geometry\_processing/} and \texttt{hartree\_fock\_calculations/}, but does not see tools nested under unrelated branches like \texttt{electronic\_property\_analysis/} unless the task requires them.

\paragraph{Curriculum Learning Through Task Sequencing.}

The persistent toolset across tasks naturally enables curriculum learning~\cite{bengio2009curriculum, wang2023voyager}: when tasks are arranged from simple to complex, early agents generate foundational tools that later agents can directly reuse. For example:
An early task, such as ``optimize the geometry of methane'' generates foundational tools for converting molecular strings (e.g., SMILES (Simplified Molecular-Input Line-Entry System)~\cite{weininger1988smiles}) to 3D structures and performing geometry optimization using Hartree--Fock (HF). A subsequent task, such as ``compute the hydride abstraction energy of ethane'', reuses the structure generation and optimization tools but requires new tools to compute thermochemical properties. A later task such as ``compute the $pK_a$ of acetic acid in solution'' reuses all prior tools and additionally generates tools for implicit solvation and Gibbs free energy corrections.

The goal of curriculum learning is to progressively build a toolset that covers the space of tasks in a given domain, so that future tasks can be solved primarily through tool reuse rather than repeated generation. This staged approach ensures that agents build on previously discovered knowledge, amortizing the cost of tool generation across tasks and thereby creating an increasingly capable toolset. As the toolset matures through the curriculum, the proportion of tasks requiring the generation of new tools diminishes. Importantly, later agents not only benefit from earlier tools but also refine them: when bugs are discovered during reuse, the universal agent capability enables on-the-fly debugging, improving tool quality over time without human intervention.

\subsection{Tool Analyzer}
\label{sec:tool-analyzer}

The \textit{tool analyzer} is responsible for translating a natural-language task into a concrete plan of which tools should be reused and which need to be generated. Given the task description and the current contents of the shared toolset, the analyzer inspects available tools, identifies the computational capabilities required by the task, and produces a structured analysis with three components: (i) a \emph{task analysis} that decomposes the problem into its computational stages, links each stage to the tool that should perform it, and surfaces relevant domain considerations such as recommended methods, physical formulas, or common pitfalls; (ii) a list of \emph{existing tools to reuse}, drawn from the current toolset; and (iii) a list of \emph{new tool requirements}, each specified as an abstract contract --- a name, a description of the desired functionality and the underlying package or method to use, and the expected input and output types --- without committing to a concrete implementation.

\subsection{Tool Generator}
\label{sec:tool-generator}

The tool generator is responsible for creating new tools when the tool analyzer deems existing ones are insufficient for a given task. Each generated tool is implemented as a standalone Python function with strict interface requirements and validation mechanisms. The generator operates in an isolated \texttt{tool\_smith/task\_{\{id\}}/}  directory within the agent workspace, where it drafts implementations and runs tests without affecting the shared toolset. Once a tool passes all tests, it is copied into the root of the shared \texttt{tools/} directory, making it available for subsequent tasks. The generator follows an iterative refinement loop: it drafts an initial implementation based on task requirements, generates unit tests, executes these tests, analyzes failures, and regenerates the tool until all tests pass. A separate \emph{tool reviewer} agent then reviews the tool for correctness, adherence to interface requirements, and alignment with domain-specific generation guidelines. The generator revises the implementation if the reviewer deems it unsatisfactory.

\paragraph{Local Source Browsing.}
When generating a tool, the tool generator is prompted to directly browse the \texttt{site-packages} directory, where Python packages are installed. We provide coding agents with precomputed navigation indexes that summarize each library's module structure, import patterns, and common usage, enabling the agents to efficiently locate relevant functions without scanning entire packages. This design has two key advantages: (1) it enables tool generation against private or unreleased packages that have no public documentation, and (2) it guarantees that the agent references the exact interface of the installed version, avoiding outdated or version-mismatched information that is common in web-sourced documentation.

\paragraph{Strict Tool Interface Requirements.}
To ensure tools are reliable, composable, and debuggable, we enforce several constraints on all generated tools by prompts to the tool generator and tool reviewer:
\begin{itemize}
    \item \textbf{No ad-hoc tools.} Every tool must be a general-purpose reusable utility, not a task-specific script. The tool analyzer enforces this via an LLM-based validator that rejects proposals containing one-time scripts unlikely to be reused. This constraint ensures the toolset remains reusable across the curriculum and prevents redundant generation of near-identical single-purpose tools.

    \item \textbf{No silent failures or fallback behaviour.} Tools must raise explicit exceptions when encountering errors rather than returning default values, suppressing warnings, or attempting automatic recovery. This strict error propagation is essential for enabling downstream debugging: when a tool fails during execution, the calling agent receives clear diagnostic information and can inspect, modify, and repair the tool. Silent failures would mask bugs and prevent the self-healing behaviour central to our approach.

    \item \textbf{Pydantic validation for type safety.} All tool inputs and outputs are validated using \texttt{Pydantic}, a Python library that provides runtime type checking and data validation through declarative schemas. Each tool defines input parameters as \texttt{BaseModel} classes with explicit types, constraints, and documentation. This serves multiple purposes: (1) it enables agents to reason precisely about tool interfaces when composing multi-step workflows, (2) it catches type errors at runtime before they propagate through complex pipelines, and (3) it provides self-documenting code that agents can parse to understand tool capabilities.
\end{itemize}

\paragraph{Skill-Based Generation Guidelines.}
The tool generator can optionally leverage a collection of skills~\cite{claude_skills, scientific_agent_skills_2026}---curated markdown documents that encode domain-specific best practices, common patterns, and expert preferences. When generating tools, agents can read relevant skill documents to inform their implementations. For example, a skill document might direct the agent to use \texttt{TenPy} for \ac{dmrg} calculations instead of a from-scratch \texttt{NumPy} implementation, specify the correct pulse optimization interface in \texttt{QuTiP}, or provide a detailed reference for diagnosing imaginary vibrational frequencies in geometry optimizations. This mechanism serves as an important way to compensate for gaps in an LLM's domain knowledge: rather than relying solely on parametric knowledge from training data, agents can consult authoritative, user-curated guidance at generation time. This allows \EAF{} to generate tools aligned with established scientific practices. The generation guidelines used in our experiments are provided in Appendix~\ref{sec:generation-guidelines}.

\subsection{Toolset Optimizer}
\label{sec:toolset-optimizer}
As agents solve more tasks and the toolset grows, maintaining navigability and context efficiency becomes critical. The tool structure optimizer is a separate coding agent that is triggered at the beginning of each new task, before tool analysis and generation. It recursively scans the \texttt{tools/} directory and creates new subdirectories to group tools or directories when any directory exceeds a size threshold, ensuring that agents can quickly locate relevant tools without overwhelming their context windows.

The optimizer operates through a simple heuristic: when a tool category directory exceeds a threshold number of tools (10 in our configuration), it analyzes tool names, docstrings, and dependency patterns to identify natural subcategories. It then creates new subdirectories, moves tools accordingly, and updates import paths and documentation. Figure~\ref{fig:toolset-optimization} illustrates this process: an initially flat toolset is reorganized into subcategories, while newly generated tools are placed at the root awaiting future reorganization.

In addition to reorganizing, the optimizer can identify functionally overlapping tools and merge them into unified implementations, thereby preventing unnecessary expansion of the tool library as multiple agents independently generate similar utilities. This feature was not enabled in our benchmark runs, as the toolsets remained small enough that merging introduced unnecessary overhead. A full description is provided in Appendix~\ref{sec:tool-merger}.

\subsection{Task Executor}
\label{sec:task-executor}

Once all required tools are available, the task executor is invoked to generate and execute the solution pipeline. It receives the original task description, along with the tool analysis plan produced by the tool analyzer, and operates as a general-purpose coding agent within the agent workspace.

\paragraph{Script Generation and Execution.}
The coding agent is prompted with a structured instruction that includes the task specification, a navigation index of available Python packages, and the current contents of the \texttt{tools/} directory. Guided by these inputs, the agent designs a multi-phase scripting strategy: it decomposes the computation into logical stages (e.g., preprocessing and input validation, main simulation, and result aggregation), each implemented as a separate Python script. The agent is prompt to first tests its pipeline on a small subset of inputs (two to three molecules or parameter points) to catch errors early, then scales to the full dataset once correctness is verified. Scripts are executed via the \ac{mcp} execution interface (described below), which dispatches jobs to either a local subprocess or an \ac{hpc} scheduler depending on the environment. During execution, the agent monitors outputs and logs, iteratively debugging failures until the pipeline completes. Because every agent has full write access to the \texttt{tools/} directory, it can also debug and modify existing tools on the fly when a reused tool produces incorrect results or raises unexpected errors, enabling self-repair without re-entering the tool generation phase.
The agent is prompt to compiles results into a \texttt{report.md} in the workspace that addresses each aspect of the original task specification. The presence and completeness of this report serves as the primary indicator of success for the iteration.

\paragraph{Model Context Protocol for Command Execution.}

Scientific computing often requires executing long-running computational jobs on \ac{hpc} clusters with workload management systems like \texttt{SLURM} \cite{Yoo2003slurm}. To handle this seamlessly, we integrate an \ac{mcp} server~\cite{anthropic_mcp_2024} that provides agents with robust command-execution capabilities across heterogeneous computational environments.
In \ac{hpc} environments with \texttt{SLURM}, it automatically generates job submission scripts with appropriate resource requests, submits jobs to the scheduler and retrieves results. In local or cloud environments without job schedulers, it executes commands as subprocesses. This abstraction allows agents to focus on scientific logic rather than infrastructure details.

\subsection{Solution Evaluator}
\label{sec:solution-evaluator}

The solution evaluator is a coding agent that determines whether a generated report satisfies the task requirements; if it does not, the task executor is re-invoked with a structured next-step plan to refine the solution, enabling iterative improvement without human intervention. It reads both the task specification (\texttt{question.md}) and the generated report (\texttt{report.md}), and is prompted to browse the workspace to collect additional context to prevent the executor from cheating.
It outputs a structured \texttt{evaluation.json} with boolean flags across five dimensions: \texttt{bug\_need\_fix} (whether bugs are detected that require fixing), \texttt{script\_complete} (whether all required computations are implemented), \texttt{further\_simulation\_needed} (whether additional simulation is required), \texttt{result\_complete} (whether all required deliverables are present, such as data tables, visualizations, and interpretations), and \texttt{next\_step\_needed} (whether any of the above conditions are unmet). When further work is required, it also provides an actionable \texttt{next\_step\_plan} in text, which is incorporated into the updated task specification for the next iteration. This evaluation-refinement loop runs for a maximum of five iterations, after which the task is marked as failed if requirements are still unmet. The detailed prompt can be found in Appendix~\ref{app:solution-evaluator}.

\section{Results}
\label{sec:results}

We evaluate \EAF{} across two capability domains: quantum chemistry and quantum dynamics. For quantum chemistry tasks, agents are required to use \texttt{PySCF}~\cite{pyscf2018, pyscf2020} as the primary electronic structure backend, supplemented by \texttt{RDKit}~\cite{rdkit} and the Atomistic Simulation Environment (\texttt{ASE})~\cite{ase2017}. For quantum dynamics tasks, the required software includes \texttt{CUDA-Q}~\cite{cudaq2024} and \texttt{Qiskit}~\cite{qiskit2024} for quantum circuit simulation, \texttt{Tequila}~\cite{tequila2021} for variational quantum algorithms, \texttt{QuTiP}~\cite{qutip2012, qutip2013} for open quantum system dynamics, and \texttt{TenPy}~\cite{tenpy2018} for tensor network simulations.

\subsection{Quantum Chemistry}

We first benchmark our system on 13 quantum chemistry tasks, derived from the benchmark set of {\cinzel El Agente Q} and {\cinzel El Agente Gr\'afico}, consisting of undergraduate-level tasks, spanning six categories: organic compounds property calculation, inorganic compounds property calculation (both covering geometry optimization, HOMO--LUMO gap, point group symmetry, and Mulliken charges), relative carbocation stability ($\Delta H$ and $\Delta G$ for hydride abstraction), cycloalkane ring strain energies (C$_3$--C$_8$), $pK_a$ prediction with implicit solvation and proton calibration, and \ac{tddft} absorption spectra (S$_1$ energy, oscillator strength, S$_1$--T$_1$ gap) \cite{casida1995time}. 
Most categories are tested at two difficulty levels: Level~1 typically provides simpler or more scaffolded inputs, whereas Level~2 requires agents to handle more complex cases. Additionally, we introduce a Level~0 task for the organic compounds category, a single-molecule analysis of caffeine, to provide a simpler entry point for curriculum learning, allowing the agent to bootstrap foundational tools before tackling the harder tasks (see Table~\ref{tab:benchmark-exercises}).

\subsection{Quantum Dynamics}
The quantum dynamics benchmark comprises 11 tasks drawn from the {\cinzel El Agente Cuántico}. These tasks span three categories: (i) quantum circuit implementation and quantum control, including Bell state preparation, \ac{vqe} dissociation curves \cite{peruzzo2014variational, tilly2022variational}, noisy channel characterization, and GRAPE-based optimal control of a three-level system \cite{khaneja2005optimal}; (ii) open quantum system dynamics, including \ac{fmo} exciton dynamics simulated via \ac{heom} \cite{adolphs2006proteins, tanimura2020numerically}, transverse-field Ising model magnetization, and Lindblad evolution \cite{Lindblad1976}; and (iii) many-body simulations, including Floquet kicked Ising dynamics, Floquet dynamics with \ac{tddmrg} \cite{white2004real, vidal2004efficient}, and phase diagram computation via DMRG \cite{white1993density} (see Table~\ref{tab:cuantico_problems}).

\begin{table}[t]
\centering
\caption{Quantum chemistry benchmark: 13 tasks across 6 categories (\texttt{PySCF}), derived from the benchmark set of {\cinzel El Agente Q}~\cite{zou2025agente} and {\cinzel El Agente Gr\'afico}~\cite{bai2026agente}. Each task category is divided into levels of increasing difficulty. Full task prompts are provided in Appendix~\ref{sec:task-prompts}.}

\label{tab:benchmark-exercises}
\small
\begin{tabular}{@{}p{0.20\linewidth}cp{0.67\linewidth}@{}}
\toprule
\textbf{Category} & \textbf{Levels (L)} & \textbf{Task} \\
\midrule
Organic & 0/1/2 & Geometry optimization of molecules; report energy, point group, dipole, MO analysis, HOMO--LUMO gap, Mulliken charges, and Hessian check. L0: caffeine; L1: 3 organic molecules from SMILES; L2: 6 organic molecules from \texttt{XYZ} \\[3pt]
Inorganic & 1/2 & Same workflow as organic plus structure visualization. L1: Cr(CO)$_6$, ClF$_3$, fluorophosphate from SMILES; L2: 6 inorganic/organometallic species from \texttt{XYZ} \\[3pt]
Carbocations & 1/2 & Calculate $\Delta H$ and $\Delta G$ for the reaction R--H $\to$ R$^+$ + H$^-$ across 9 species (methane through toluene). L1: \texttt{XYZ} geometries provided; L2: agent generates 3D geometries from SMILES \\[3pt]
Ring strain & 1/2 & Calculate ring strain energies of cycloalkanes ($n\!=\!3$--$8$) via isodesmic reactions with frequency corrections. L1: reaction scheme given, cyclohexane reference; L2: agent determines scheme and reference \\[3pt]
$pK_a$ & 1/2 & Calculate $pK_a$ of acid in water. L1: acetic acid; L2: chlorofluoroacetic acid with proton solvation calibration against three known acids \\[3pt]
\ac{tddft} & 1/2 & \ac{tddft} on 3 chromophores; report S$_1$ energy, oscillator strength, S$_1$--T$_1$ gap. L1: \texttt{XYZ} files and methodology hint provided; L2: no methodology hint \\
\bottomrule
\end{tabular}
\end{table}

\begin{table}[t]
\centering
\caption{Quantum dynamics benchmark: 11 tasks across 3 categories, derived from the benchmark set of {\cinzel El Agente Cu\'antico}~\cite{gustin2025cuantico}. Full task prompts are provided in Appendix~\ref{sec:task-prompts}.}
\label{tab:cuantico_problems}
\small
\renewcommand{\arraystretch}{1.1}
\begin{tabular}{@{}p{0.25\linewidth}lp{0.55\linewidth}@{}}
\toprule
\textbf{Category} & \textbf{Main software} & \textbf{Task} \\
\midrule
State Prep.\ \& Control & \texttt{CUDA-Q} & Bell state preparation and verification from measurements  \\
  & \texttt{Tequila} & H$_2$ VQE dissociation curve compared against Full-CI\\
  & \texttt{Qiskit} & Decay of Bell states under depolarizing channels \\
  & \texttt{QuTiP} & GRAPE optimal control of a three-level system \\
\midrule
Open Quantum Systems & \texttt{QuTiP} & Unitary vs.\ Lindblad dynamics of a single qubit \\
  & \texttt{QuTiP} & \ac{fmo} exciton dynamics via \ac{heom} at $T = 300$\,K and $77$\,K \\
  & \texttt{TenPy} & Gibbs state of 1D Hubbard chain via imaginary-time evolution \\
  & \texttt{CUDA-Q} & Magnetization dynamics in transverse-field Ising model  \\
\midrule
Many-Body Systems & \texttt{CUDA-Q} & Dynamics of Floquet kicked Ising chain from a random state \\
  & \texttt{TenPy} & Dynamics of disordered Floquet spin chain via TD-DMRG \\
  & \texttt{TenPy} & Phase transition in the cluster model via DMRG \\
\bottomrule
\end{tabular}
\renewcommand{\arraystretch}{1.0}
\end{table}

\subsection{Evaluation}
To evaluate \EAF{}, we compare three operating modes across five coding agent setups to show the effect of tool generation and curriculum learning.
\begin{itemize}
    \item \textbf{Zero-shot (ZS):} Each problem is solved independently with an empty toolset; no prior tools are available. The four-stage agent workflow is active, so the agent forges an independent toolset for each task, exercising the tool generator and iterative evaluator from scratch.

    \item \textbf{Tool reuse (TR):} Each problem is solved with access to a toolset constructed via curriculum learning using Claude Opus~4.6 (via Claude Code) across all problems in the benchmark.
    The same toolset is then reused by in all the setups. Tool generation is disabled; the tool analyzer remains active, enabling the agent to discover relevant tools through progressive disclosure and compose them.
    \item \textbf{Evaluator only (EO):} The agent directly solves the problem without invoking the tool builder or accessing any existing toolset. Solutions are still verified by the automated solution evaluator (Section~\ref{sec:solution-evaluator}). This mode serves as a baseline that reflects the raw capability of the underlying coding agent without tool scaffolding.
\end{itemize}

Each mode--setup combination is repeated 3 times; for tool reuse (TR), each repetition draws from an independently curriculum-learned toolset from Claude Opus 4.6 (via Claude Code) with a fixed order of tasks (see Appendix~\ref{sec:per-task-scores}). Across 24 tasks, 3 operating modes, 5 coding agent setups, and 3 repetitions, the evaluation encompasses 1{,}080 total task runs.
Task success is evaluated against explicit criteria for each task, based on the numerical results and methodology adopted in the solutions.
Each task is scored along two dimensions, \emph{accuracy} and \emph{methodology}, using per-criterion rubrics detailed in Appendix Tables~\ref{tab:criterions-q-acc}, \ref{tab:methodology-q}, \ref{tab:criterions-cuantico-acc}, and~\ref{tab:methodology-cuantico}. The combined score for each task is the mean of the accuracy and methodology subscores. For example, accuracy criteria check whether computed energies fall within specified tolerances of reference values and whether required physical quantities are reported, while methodology criteria verify that the correct computational workflow was followed.

\begin{table}[t]
    \centering
    \caption{Average combined score (mean of accuracy + methodology/2 per question, averaged over 3 runs; $\pm$ denotes the sample standard deviation across the 3 runs; bold = best setup (ZS/TR/EO) for each model within each benchmark). Tool reuse (TR) yields the highest average on both benchmarks, with the largest gains for weaker models (e.g., Kimi: +16.5pp on quantum chemistry). Zero-shot (ZS) tool generation consistently outperforms the evaluator-only (EO) baseline, confirming that tool generation contributes beyond the evaluation loop alone.}
    \label{tab:score_comparison}
    \small
    \setlength{\tabcolsep}{4pt}
    \begin{tabular}{l  r r r  r r r}
    \toprule
    & \multicolumn{3}{c}{\textbf{Quantum Chemistry}} & \multicolumn{3}{c}{\textbf{Quantum Dynamics}} \\
    \cmidrule(lr){2-4} \cmidrule(lr){5-7}
    \textbf{Model} & ZS & TR & EO & ZS & TR & EO \\
    \midrule
    Claude Opus~4.6 (via Claude Code) & 86.6\%\stdv{4.9} & \textbf{87.9\%}\stdv{2.6} & 80.4\%\stdv{3.1} & \textbf{97.3\%}\stdv{1.5} & 97.1\%\stdv{2.0} & 96.4\%\stdv{1.1} \\
    Claude Sonnet~4.6 (via Claude Code) & \textbf{85.4\%}\stdv{1.5} & 84.4\%\stdv{1.7} & 84.2\%\stdv{1.6} & \textbf{98.5\%}\stdv{1.1} & 97.7\%\stdv{0.2} & 97.2\%\stdv{1.1} \\
    GPT-5.2-Codex (via Codex) & 83.2\%\stdv{3.0} & \textbf{88.1\%}\stdv{2.5} & 77.1\%\stdv{4.3} & 86.1\%\stdv{1.5} & \textbf{88.9\%}\stdv{1.2} & 86.8\%\stdv{2.1} \\
    Gemini~3.1~Pro (via OpenCode) & \textbf{86.7\%}\stdv{2.1} & 86.3\%\stdv{4.8} & 80.3\%\stdv{2.5} & 91.6\%\stdv{0.4} & 90.7\%\stdv{3.3} & \textbf{92.3\%}\stdv{3.6} \\
    Kimi~K2.5 (via OpenCode) & 65.7\%\stdv{15.0} & \textbf{82.2\%}\stdv{5.8} & 61.5\%\stdv{7.6} & 83.9\%\stdv{5.7} & \textbf{91.5\%}\stdv{4.9} & 84.0\%\stdv{4.7} \\
    \midrule
    \textit{Average} & 81.5\%\stdv{3.7} & \textbf{85.8\%}\stdv{2.4} & 76.7\%\stdv{1.5} & 91.5\%\stdv{1.0} & \textbf{93.2\%}\stdv{1.5} & 91.3\%\stdv{0.8} \\
    \bottomrule
    \end{tabular}
\end{table}

\paragraph{Overall Performance.}

Table~\ref{tab:score_comparison} summarizes overall performance across all five agentic setups and both benchmarks. On quantum chemistry, zero-shot scores range from 65.7\% (Kimi~K2.5 via OpenCode) to 86.7\% (Gemini~3.1~Pro via OpenCode), with an average of 81.5\% across models. On quantum dynamics, zero-shot scores are higher, ranging from 83.9\% (Kimi~K2.5 via OpenCode) to 98.5\% (Claude Sonnet~4.6 via Claude Code), with an average of 91.5\%, indicating that frontier coding agents already possess substantial scientific reasoning capability for quantum dynamics tasks out of the box.

The pattern across conditions differs between the two benchmarks. On quantum dynamics, all three setups achieve high scores (91.3--93.2\% average), and the evaluator-only (EO) baseline matches or exceeds zero-shot (ZS) for three of the five models (GPT-5.2-Codex via Codex, Gemini~3.1~Pro via OpenCode, and Kimi~K2.5 via OpenCode), suggesting that the iterative evaluation loop alone is largely sufficient when tasks are more self-contained. By contrast, on quantum chemistry---where tasks demand multi-step workflows, domain-specific methodology, and precise numerical outputs---EO scores drop substantially (76.7\% average vs.\ 81.5\% ZS), confirming that tool generation and reuse contribute meaningfully beyond the evaluation loop on harder, more complex tasks. 

Per-task breakdowns (see Appendix~\ref{sec:per-task-scores}) reveal systematic difficulty patterns. On quantum chemistry, Ring Strain L2 (66.4\% ZS average) is the hardest task: it requires agents to independently determine an isodesmic reaction scheme, perform geometry optimizations and frequency calculations for eight cycloalkanes plus reference molecules, and apply thermodynamic corrections. This is a multi-step pipeline in which a single methodological error (e.g., a missing zero-point energy correction) cascades through all downstream results. Tool reuse raises this to 76.1\%, because the curriculum-built tools already encode the correct thermochemical workflow. Similarly, $pK_a$ L2 (78.0\% ZS, 77.0\% TR) demands implicit solvation modeling and proton solvation energy calibration against known acids, a domain-specific technique that weaker models frequently fail to derive from scratch but execute reliably when encoded in reusable tools. In quantum dynamics, the Gibbs/Hubbard task (71.7\% ZS, 75.1\% TR) requires imaginary-time DMRG evolution with TenPy, where convergence depends on careful bond-dimension and time-step choices; tool reuse provides pre-validated parameter choices that reduce much of the trial-and-error exploration.

\paragraph{Tool-Editing in Execution.}
We designed the task executor to have full write access to tool files for on-the-fly self-repair. We found this mechanism is actively used in some setups: Kimi~K2.5 via OpenCode and Gemini~3.1~Pro via OpenCode edit or create tool files in over 40\% of iterations in the zero-shot setting, while the Claude Code models rarely need to (1--6\%), reflecting the higher quality of their initially generated tools (Appendix~\ref{sec:tool-edits}). In the tool-reuse setting, edits drop to near zero across all models because the pre-built library requires little correction.

\paragraph{Tool Reuse Consistently Reduces Cost.}
A central motivation of curriculum learning is to amortize the cost of tool generation across tasks: once foundational tools are forged by an early, stronger agent, later agents skip the expensive tool-generation loop entirely and proceed directly to tool composition. This design yields substantial LLM API cost reductions across both benchmarks and all five models. On quantum dynamics (see Table~\ref{tab:cuantico_comparison}), average cost reductions range from 62.7\% for Claude Opus~4.6 via Claude Code to 77.9\% for Kimi~K2.5 via OpenCode, with a mean reduction of 66.9\% across all models. On quantum chemistry (see Table~\ref{tab:q_comparison}), reductions are more modest but still consistent, ranging from 33.2\% for Claude Sonnet~4.6 via Claude Code to 64.2\% for Kimi~K2.5 via OpenCode, with a mean of 49.6\%. The effect is most pronounced for weaker models: Kimi~K2.5 via OpenCode achieves the largest cost savings on both benchmarks, suggesting that tool reuse disproportionately benefits models that would otherwise spend more tokens on exploratory reasoning and re-derivation, precisely the token-intensive tool generation and iterative testing phases that the curriculum library eliminates.

\paragraph{Tool Reuse Reduces Runtime for Most Models.}
Tool reuse also reduces wall-clock time for most model--benchmark combinations, because agents bypass the tool generation and unit-testing loop and proceed directly to task execution. The most dramatic speedup occurs for Claude Opus~4.6 via Claude Code on quantum dynamics: from an average of 128.4~minutes (ZS) to 15.2~minutes with tool reuse ($-$88.1\%), as the model can directly compose tools rather than re-discovering and re-implementing solutions. Kimi~K2.5 via OpenCode also benefits substantially, dropping from 453.8 to 75.4~minutes ($-$83.4\%) on quantum dynamics and from 276.2 to 97.3~minutes ($-$64.8\%) on quantum chemistry. Overall, runtime reductions in quantum chemistry range from 13.3\% for Claude Opus~4.6 via Claude Code to 64.8\% for Kimi~K2.5 via OpenCode, consistent with the harder generalization required. GPT-5.2-Codex via Codex is the sole exception, with tool reuse yielding a marginal increase of $+$0.4\% on quantum chemistry.

\begin{figure}[h]
    \centering
    \includegraphics[width=\linewidth]{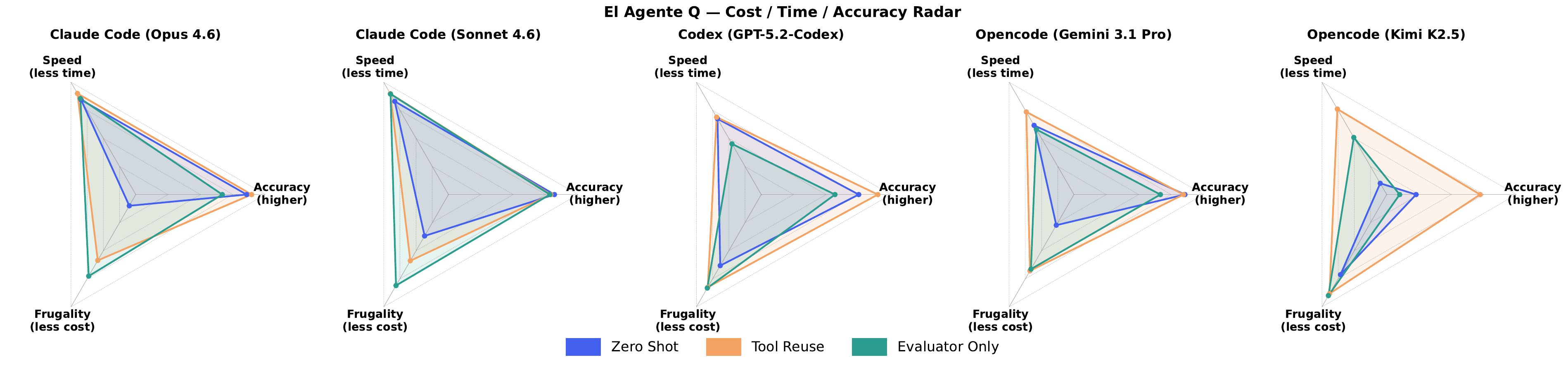}\\[6pt]
    \includegraphics[width=\linewidth]{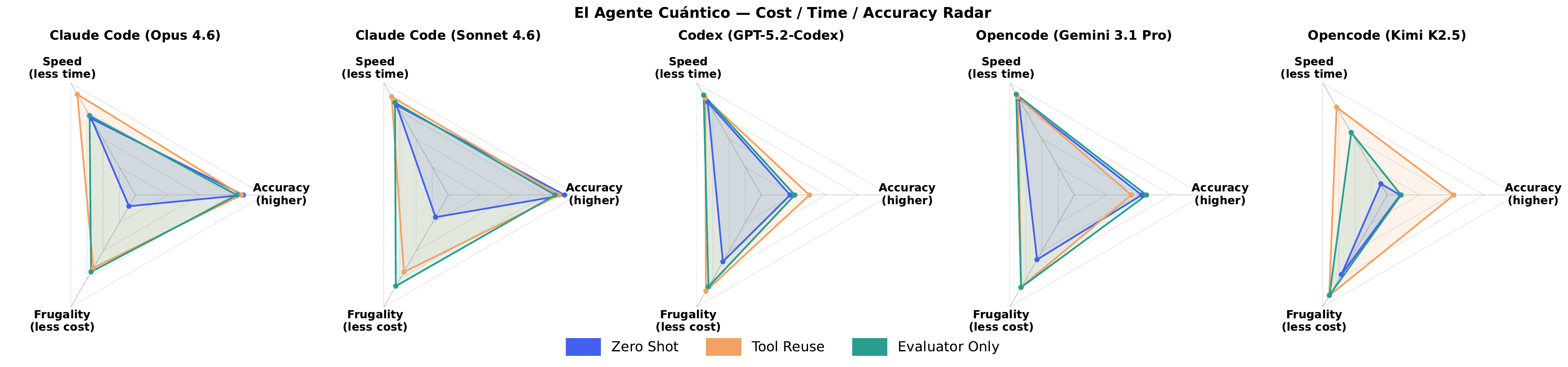}
    \caption{Relative Cost / Time / Accuracy radar plots for \EAF{} with five agent setups on the quantum chemistry tasks (top) and quantum dynamics tasks (bottom) with different tool generation strategies. Each axis is independently min--max normalized across all models and conditions, then rescaled to $[0.1, 0.9]$. Accuracy (higher = better); Speed (lower time = better); Frugality (lower cost = better). Each axis represents a model; the shaded area indicates the fraction of tasks solved. Tool reuse consistently expands the shaded area toward the frugality and speed axes while maintaining or improving accuracy, with the most pronounced gains for weaker models such as Kimi~K2.5 via OpenCode and GPT-5.2-Codex via Codex.}
    \label{fig:radar}
\end{figure}

\paragraph{Tool Reuse Maintains or Improves Accuracy.}
Zero-shot accuracy is already high across both benchmarks, with average scores (combining accuracy and methodology) ranging from 65.7\% to 98.5\% depending on model and benchmark (Table~\ref{tab:score_comparison}). Tool reuse either maintains or improves these scores. In quantum chemistry, tool reuse raises the average score from 81.5\% (ZS) to 85.8\% (TR); in quantum dynamics, from 91.5\% to 93.2\%. The improvement is most pronounced for weaker models: Kimi~K2.5 via OpenCode jumps from 65.7\% to 82.2\% on quantum chemistry and from 83.9\% to 91.5\% on quantum dynamics, a +16.5 and +7.6 percentage-point gains, respectively. GPT-5.2-Codex via Codex similarly improves from 83.2\% to 88.1\% on quantum chemistry and from 86.1\% to 88.9\% on quantum dynamics. These results demonstrate that the curriculum-built toolset not only reduces cost but also provides a \emph{scaffolding} effect: the tool generator (Section~\ref{sec:tool-generator}) encodes correct computational methodology into the tools themselves, allowing weaker models to achieve accuracy comparable to stronger ones by composing well-tested building blocks rather than re-deriving domain logic from scratch.

\subsection{Case Studies}
A key promise of task-driven tool generation is that the resulting toolsets are not confined to the domain in which they were created: tools forged for quantum chemistry should compose seamlessly with tools forged for quantum dynamics, enabling cross-domain workflows. Moreover, when the existing library lacks a required capability, the agent can forge new tools on the fly, extending its repertoire without human intervention. To test this, we present two representative case studies. Both problems require bridging classical electronic-structure theory and open-quantum-system dynamics, a combination that demands tools from multiple domains and, in each case, triggers the generation of new tools to fill gaps in the existing toolset. To set up the case studies, we seeded the agent's toolset by combining one curriculum-built toolset from the quantum chemistry benchmark with one from the quantum dynamics benchmark, giving the agent a cross-domain toolset to draw from at the outset. Traces of the case studies can be found in Appendix~\ref{sec:trace}.

\begin{figure}
    \centering
    \includegraphics[width=0.8\linewidth]{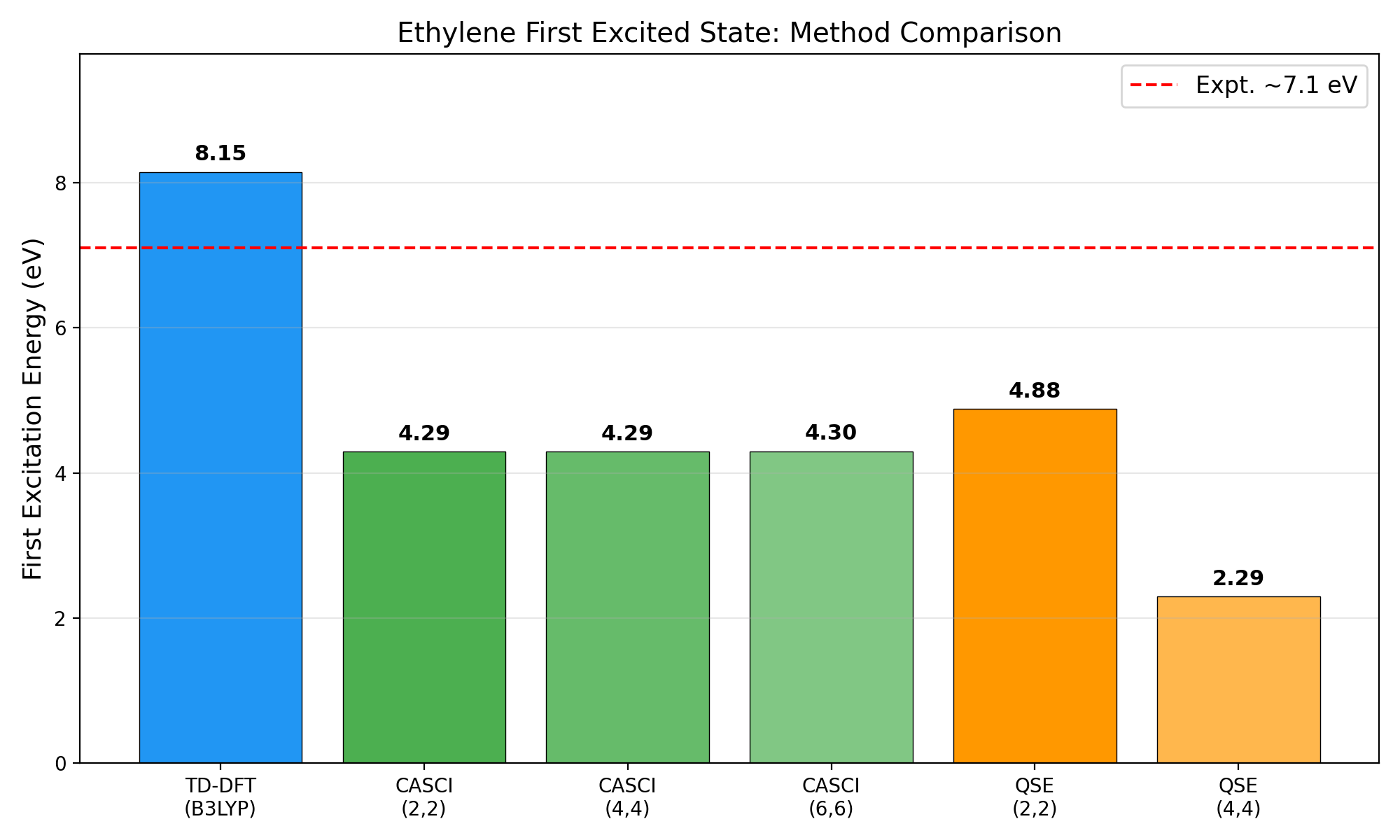}
    \caption{Figure generated by agents for case study 1: Comparison of first excitation energies from \ac{tddft} (B3LYP/def2-SVP), CASCI at three active space sizes (def2-SVP), QSE at two active space sizes (STO-3G, \texttt{CUDA-Q} simulation), and the experimental valence $\pi \to \pi^*$ excitation of ethylene (${\sim}7.66$~eV, dashed red line). CASCI results are consistent across active spaces (${\sim}4.29$~eV, corresponding to T$_1$). QSE(2,2) agrees well with CASCI, while QSE(4,4) deviates due to the larger subspace and minimal basis limitations. \ac{tddft} provides the closest estimate to the valence S$_1$ excitation among the methods shown.}
    \label{fig:cs1}
\end{figure}

\subsubsection{Case Study 1: Ethylene Excited States via \ac{tddft} and Quantum Subspace Expansion}

Ethylene is the prototypical benchmark for $\pi \to \pi^*$ excitations and a stringent test of electronic structure methods. Comparing \ac{tddft}, the workhorse of computational photochemistry, against a quantum algorithm---QSE~(Quantum Subspace Expansion)~\cite{mcclean2017hybrid} in this case---on the same molecule probes both the strengths and limitations of each approach. The problem additionally requires interoperability across three distinct software ecosystems: \texttt{PySCF}, \texttt{Tequila}, and \texttt{CUDA-Q}. The following prompt is provided to \EAF{}.

\prompt{For the molecule ethylene (C=C), compute the first excited state energy using TD-DFT, and estimate the first excited state energy using a quantum algorithm.
Optimize the geometry using B3LYP/def2-SVP and compute the S$_1$ energy using TD-DFT.
Identify an active space capturing the $\pi$ and $\sigma$ system.
Estimate the first excited state energy using Quantum Subspace Expansion with CUDA-Q simulation.
Analyze the nature of the $\pi \to \pi^*$ transition, including limitations of TD-DFT versus quantum algorithms and the effect of correlation and active space choice. Make a visualization of important molecular orbitals.}

The tool analyzer identified nine candidate tools, of which five were selected for the workflow, spanning geometry optimization, \ac{tddft}, and molecular orbital analysis (\texttt{PySCF}-based), as well as Hamiltonian construction (\texttt{Tequila}-based). Crucially, three capabilities were missing from the existing library, and the tool generator forged them on the fly:
\begin{itemize}
    \item \texttt{run\_pyscf\_casci.py} --- runs CASCI calculations for classical multireference benchmarking within a user-specified active space;
    \item \texttt{visualize\_molecular\_orbitals.py} --- renders 2D contour plots of molecular orbital wavefunctions on a grid, enabling the orbital visualizations requested in the prompt;
    \item \texttt{run\_quantum\_subspace\_expansion.py} --- implements the QSE algorithm with CUDA-Q simulation, building the excitation subspace from single and double excitation operators, and solving the generalized eigenvalue problem.
\end{itemize}
All three tools passed automated unit testing and review before being used in the workflow.

\begin{figure}
    \centering
    \includegraphics[width=0.48\linewidth]{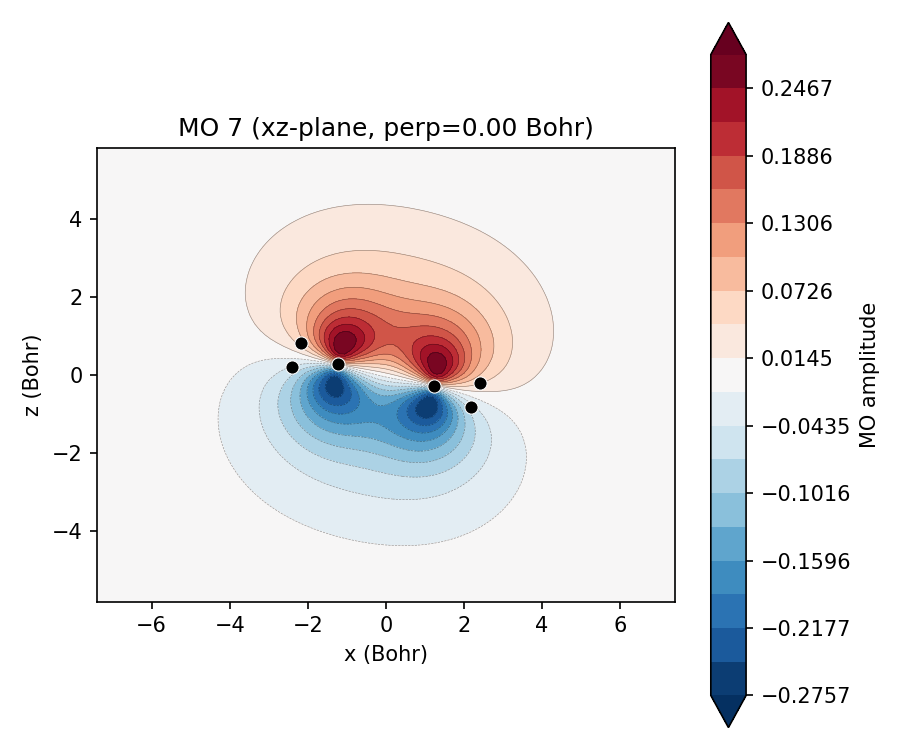}\hfill
    \includegraphics[width=0.48\linewidth]{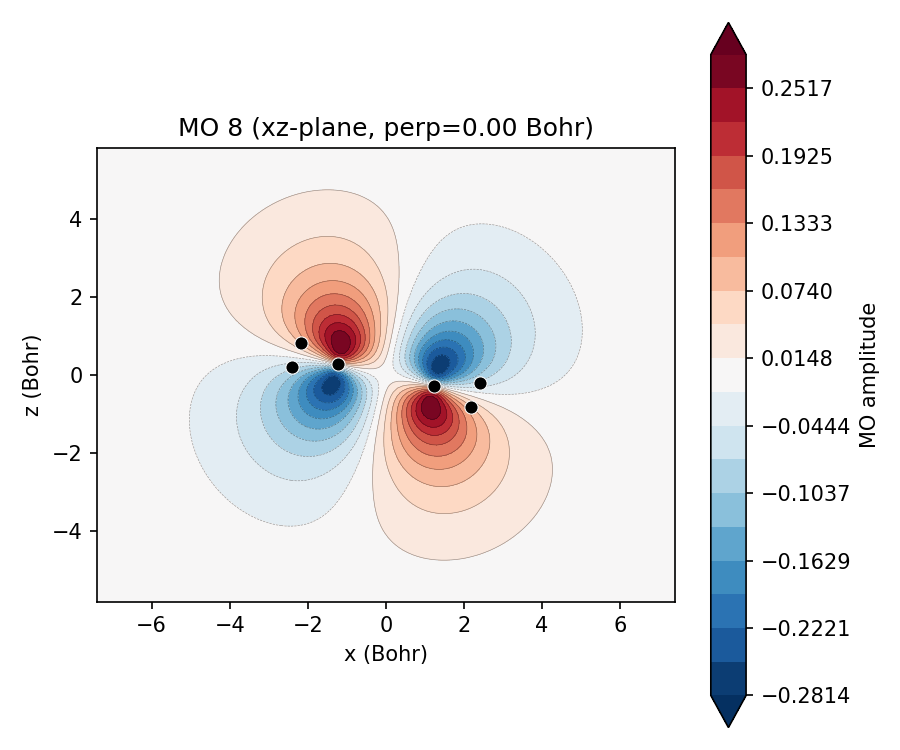}
    \caption{Frontier orbital visualizations generated by the newly forged \texttt{visualize\_molecular\_orbitals.py} tool: HOMO ($\pi_\text{CC}$, left) and LUMO ($\pi^*_\text{CC}$, right) of ethylene in the $xz$-plane. The $\pi \to \pi^*$ character of the dominant S$_1$ transition is clearly visible from the nodal structure.}
    \label{fig:cs1_orbitals}
\end{figure}

The agent executed the workflow as follows:
\begin{enumerate}
    \item \texttt{smiles\_to\_3d\_geometry.py} \emph{(reused from quantum chemistry)} generated the initial ethylene structure from the SMILES string \texttt{C=C}, followed by \texttt{run\_dft\_geometry\_optimization.py} \emph{(reused from quantum chemistry)} at B3LYP/def2-SVP ($E_0 = -78.5316$~Hartree).
    \item \texttt{run\_tddft\_calculation.py} \emph{(reused from quantum chemistry)} computed singlet and triplet excited states, identifying the $\pi \to \pi^*$ S$_1$ at 8.15~eV ($f = 0.375$) and T$_1$ at 4.09~eV.
    \item \texttt{analyze\_molecular\_orbitals.py} \emph{(reused from quantum chemistry)} mapped the frontier orbitals (HOMO at $-7.51$~eV, LUMO at $0.15$~eV), guiding active space selection.
    \item \texttt{run\_pyscf\_casci.py} \emph{(newly forged)} performed CASCI calculations at three active space sizes---CAS(2,2), CAS(4,4), and CAS(6,6)---providing classical multireference benchmarks. All three yield a consistent first excitation energy of ${\sim}4.29$~eV (the T$_1$ state), confirming convergence with respect to active space size within the def2-SVP basis.
    \item \texttt{tequila\_build\_molecular\_hamiltonian.py} \emph{(reused from quantum dynamics)} constructed second-quantized Hamiltonians within CAS(2,2) and CAS(4,4) active spaces (STO-3G basis), mapped to 4 and 8 qubits respectively via Jordan--Wigner transformation.
    \item \texttt{run\_quantum\_subspace\_expansion.py} \emph{(newly forged)} executed QSE on both active spaces: QSE(2,2) with a 6-dimensional subspace yielded a first excitation of 4.88~eV, while QSE(4,4) with a 53-dimensional subspace yielded 2.29~eV.
    \item \texttt{visualize\_molecular\_orbitals.py} \emph{(newly forged)} rendered contour plots of four frontier orbitals (HOMO$-$1, HOMO, LUMO, LUMO$+$1) in both the $xz$- and $xy$-planes (Figure~\ref{fig:cs1_orbitals}), clearly resolving the $\sigma_\text{CC}$, $\pi_\text{CC}$, $\pi^*_\text{CC}$, and $\sigma^*_\text{CC}$ character.
\end{enumerate}

Results are summarized in Figure~\ref{fig:cs1}. \ac{tddft} (B3LYP/def2-SVP) places the S$_1$ at 8.15~eV, somewhat above the experimental valence $\pi \to \pi^*$ excitation energy of ethylene (${\sim}7.66$~eV). The CASCI first excitations (${\sim}4.29$~eV, def2-SVP) and QSE(2,2) (4.88~eV, STO-3G) both correspond to the T$_1$ state. The QSE(4,4) result (2.29~eV) deviates more substantially, which the agent attributed to the enlarged excitation subspace and the limitations of the minimal STO-3G basis. The agent's report therefore treated the \ac{tddft} result as the closest estimate to the valence S$_1$ excitation among the methods considered here, while the quantum algorithm primarily captured static-correlation effects within the chosen active space.

\subsubsection{Case Study 2: Rubidium-87 Hyperfine Qubit}

Rb-87 is among the most widely used atomic qubits in quantum computing and quantum sensing~\cite{henriet2020quantum}. Quantitatively connecting the electron density at the nucleus, a quantity accessible from \emph{ab initio} calculation, to the qubit's Hamiltonian and decoherence properties demonstrates the kind of cross-domain reasoning that is important to quantum engineering. The problem also probes the limits of non-relativistic DFT on heavy atoms (Rb, $Z=37$), testing whether the agent can critically assess its own computational results against experimental benchmarks. The following prompt is provided to \EAF{}.

\prompt{Model a \textbf{rubidium-87} hyperfine qubit as a two-level open quantum system. Compute the ground-state electron density at the nucleus using \textbf{PySCF}, construct a two-level Hamiltonian ($H = (\omega/2)\,Z$) from it, and simulate Lindblad dynamics with \textbf{QuTiP}. Study the evolution from (a) the excited state ($|1\rangle$) and (b) a superposition ($(|0\rangle+|1\rangle)/\sqrt{2}$), then plot the excited-state population and coherence over time and provide a brief analysis. Your plots should be publication-quality plots.}

The task required integrating the quantum chemistry toolset (built with \texttt{PySCF}) with the \texttt{QuTiP} dynamics tools for quantum dynamics. The tool analyzer identified four reusable tools: \texttt{run\_dft\_single\_point.py} and \texttt{run\_single\_point\_hf.py} from quantum chemistry, plus \texttt{qutip\_lindblad\_evolution.py} and \texttt{plot\_line\_comparison.py} from quantum dynamics. One capability was missing: evaluating the electron density at nuclear positions from a converged SCF object. The tool generator forged a new tool, \texttt{compute\_electron\_density\_at\_nuclei.py}, which evaluates AO basis functions at specified nuclear coordinates via \texttt{PySCF}'s \texttt{numint.eval\_ao} and \texttt{eval\_rho} routines and returns the total electron density, supporting both restricted and unrestricted SCF objects.

\begin{figure}
    \centering
    \includegraphics[width=0.48\linewidth]{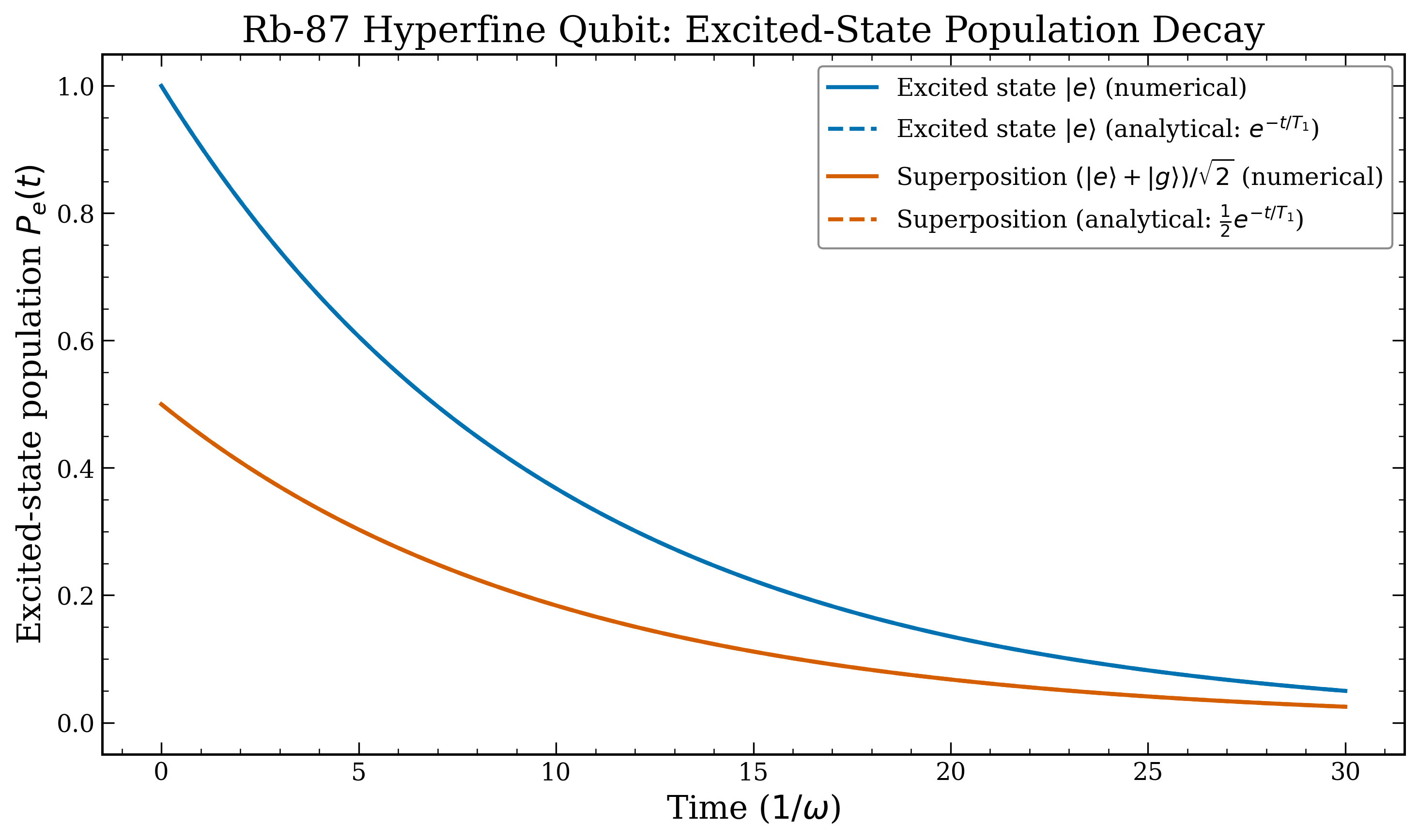}\hfill
    \includegraphics[width=0.48\linewidth]{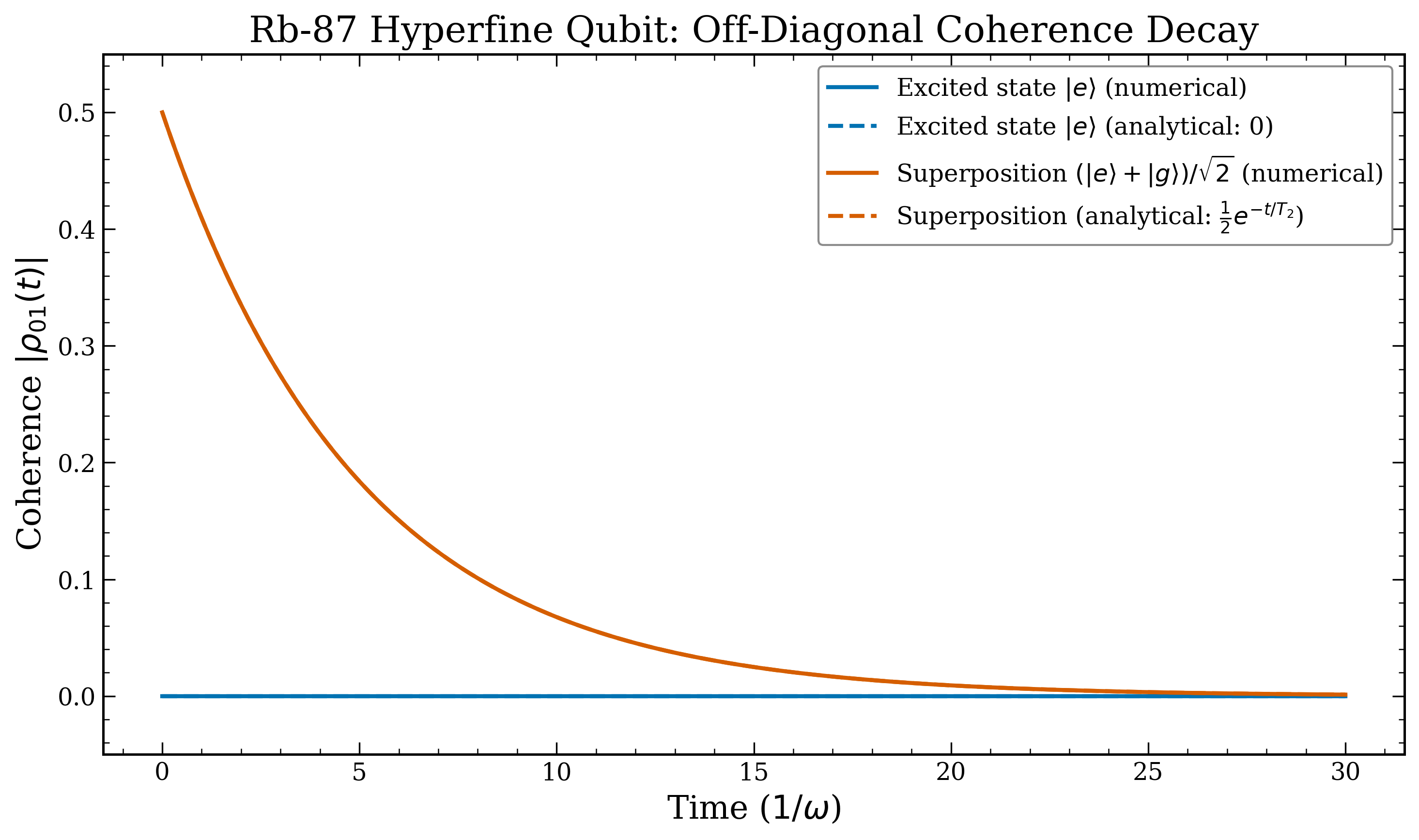}
    \caption{Figures generated by agents for case study 2: Lindblad dynamics of a $^{87}$Rb hyperfine qubit using the experimental hyperfine splitting, with the \texttt{PySCF} contact density serving as a diagnostic comparison rather than a quantitative input to the dynamics. \emph{Left:}~Excited-state population $P_e(t)$ decays exponentially with T$_1 = 10/\omega$ from both initial states ($|e\rangle$ and $(|e\rangle+|g\rangle)/\sqrt{2}$). \emph{Right:}~Coherence $|\rho_{01}|$ decays on the shorter T$_2 = 5/\omega$ timescale for the superposition state; the eigenstate $|e\rangle$ generates no coherence. Solid lines are numerical \texttt{QuTiP} results; dashed lines are analytical Bloch-equation solutions, showing agreement to machine precision.}
    \label{fig:cs2}
\end{figure}

The agent executed the workflow in a single consolidated script:
\begin{enumerate}
    \item \texttt{run\_dft\_single\_point.py} \emph{(reused from quantum chemistry)} converged a DFT/B3LYP/def2-SVP calculation for a neutral Rb atom (charge $= 0$, spin multiplicity $= 2$; $E_0 = -487.62$ Hartree). The agent first attempted a cc-pVDZ basis, found it unavailable for Rb, and switched to the def2-SVP basis.
    \item \texttt{compute\_electron\_density\_at\_nuclei.py} \emph{(newly forged)} evaluated the contact density at the nuclear origin, yielding $\rho(0) = 2.66\,e/\text{bohr}^3$. The resulting Fermi contact coupling ($A_F \approx 3.9 \times 10^{16}$~Hz) vastly overestimates the experimental hyperfine splitting ($6.83\,\text{GHz}$) because the total density is dominated by core $s$-electrons (1s--4s), not the valence 5s electron responsible for the physical coupling. The agent correctly identified this limitation and noted that isolating the valence contribution would require relativistic methods (ZORA or Douglas--Kroll--Hess). The experimental ground-state hyperfine splitting of $^{87}$Rb, $\omega_\text{hf} = 2\pi \times 6.8347\,\text{GHz}$, was adopted for the subsequent Lindblad simulation.
    \item \texttt{qutip\_lindblad\_evolution.py} \emph{(reused from quantum dynamics)} simulated the two-level system with decoherence parameters T$_1 = 10$ and T$_2 = 5$ (in units of $1/\omega$), using collapse operators for T$_1$ relaxation ($\sqrt{\gamma_1}\,\sigma_-$, $\gamma_1 = 0.1$) and pure dephasing ($\sqrt{\gamma_\varphi/2}\,\sigma_z$, $\gamma_\varphi = 0.15$). The simulation spanned $30/\omega$ to capture the full population and coherence decay. The agent initially obtained incorrect dynamics due to \texttt{QuTiP}'s basis convention ($|0\rangle$ is the $+1$ eigenstate of $\sigma_z$, i.e.\ the excited state), diagnosed the mismatch by comparing against analytical Bloch-equation solutions, and corrected the initial-state assignments and projectors accordingly.
    \item \texttt{plot\_line\_comparison.py} \emph{(reused from quantum dynamics)} generated publication-quality figures with both numerical (solid) and analytical (dashed) curves overlaid for validation (Figure~\ref{fig:cs2}).
\end{enumerate}

Results are shown in Figure~\ref{fig:cs2}. Population decays exponentially on the T$_1 = 10/\omega$ timescale from both initial states, while coherence decays on the shorter T$_2 = 5/\omega$ timescale for the superposition state; no coherence is generated from the eigenstate $|e\rangle$, consistent with Lindblad theory. The numerical \texttt{QuTiP} solutions match the analytical Bloch-equation predictions to machine precision, validating the implementation. The T$_2 < \text{T}_1$ hierarchy ($1/T_2 = 1/(2T_1) + \gamma_\varphi$) reflects the combined effect of relaxation and pure dephasing, a universal feature of open quantum systems. The agent's report discussed implications for real Rb-87 qubits, where T$_1$ can exceed seconds while T$_2$ is limited by magnetic-field fluctuations and collisions.

\section{Discussion}
\subsection{Strengths}

\paragraph{Task-Driven Agent Generation Reduces Human Involvement.}

Our approach significantly reduces the need for human intervention in building agent systems. Instead of relying on manually engineered, domain-specific prompts or hand-crafted tools, agents are generated in a task-driven manner: complex capabilities emerge through task decomposition and tool generation. This paradigm aligns well with the strengths of LLMs, whose training data already encode most of the relevant procedural and domain knowledge. In practice, describing a target task or simulation objective is often easier than designing expert-level prompts.

\paragraph{Tool Reuse Simultaneously Improves Cost, Speed, and Accuracy.}
In most machine learning settings, improvements in one metric come at the expense of another. Tool reuse breaks this trade-off: across five coding agent setups, we observe simultaneous reductions in API cost (33--78\%) and wall-clock time (up to 88\%), with equal or improved task accuracy. This is because curriculum-built tools amortize the exploration cost of tool generation: once a useful tool has been discovered and validated, it can be reused at negligible marginal cost. The result is a progressively more efficient system where per-task cost decreases as the toolset grows. Moreover, because the curriculum toolset is built by a stronger model (Claude Opus~4.6) and reused by weaker ones, tool reuse also serves as strong-to-weak knowledge transfer: correct domain methodology is encoded in the tools and can be used by weaker agents.

\paragraph{Universal Agents Enable Self-Healing Tool Ecosystems.}
Rather than assigning agents fixed context and roles, all agents in our framework are universal: any agent can code, debug, and browse the whole agent workspace. This design creates a self-healing ecosystem. Tools are often imperfect at the time of initial generation but are subsequently corrected by downstream agents. Because later agents depend on these tools to solve their own tasks, failures naturally trigger inspection, refinement, and repair, all without explicit human supervision.
In practice, bugs surface when tools are applied in new contexts, and the same agent that encounters the bug has the capability to fix it---an effect quantified by the tool-editing statistics in Appendix~\ref{sec:tool-edits}. The curriculum ordering amplifies this effect, as tools generated for simpler tasks are tested by increasingly complex downstream applications.

\subsection{Limitations \& Outlook}

We identify several limitations of the current system and outline directions for future work.

\paragraph{Unreliable Self-Verification and Reward Exploitation.}
Our system relies on agent self-verification and automated evaluation signals. As with many autonomous agent frameworks~\cite{hu2024automated,zhang2025dgm}, these signals are not always reliable. Agents may incorrectly assess correctness, fail to detect subtle errors, or converge to behaviours that exploit weaknesses in the evaluation function, optimizing for superficial success metrics rather than true task correctness. This risk is particularly acute in scientific computing, where numerically plausible but physically incorrect results (e.g., a geometry optimization converging to a saddle point rather than a minimum) may pass automated checks.

A promising mitigation is evaluation-time scaling: allocating more computation to verification and refinement, analogous to test-time compute scaling in language models~\cite{snell2024scaling}. Recent work has taken important steps toward verifiable and self-correcting AI agents for quantum many-body simulations~\cite{deng2026}. Domain-specific invariants (conservation laws, symmetry constraints, known asymptotic limits) could serve as physics-informed verification signals that are harder for agents to exploit than generic correctness metrics.

\paragraph{Restricted Software Ecosystem.}
The current implementation is limited to Python-based tools and a predefined set of repositories. While this simplifies execution and reuse, it excludes tasks that require input-file-driven software, such as ORCA~\cite{ORCA}, VASP~\cite{kresse1996}, or Gaussian~\cite{gaussian16}, external APIs, or non-Python ecosystems, which remain prevalent in computational chemistry. Extending the framework to wrap such software (generating input files, launching executables, and parsing output) is a natural next step that would substantially broaden the system’s applicability without abandoning the Python-based tool abstraction layer.

\paragraph{Toward Self-Directed Curriculum Generation.}
The current system requires human-specified task sequences to drive curriculum learning. Future versions of the system could integrate self-tasking mechanisms~\cite{10.1145/3690624.3709321,xia2025agent0,zhai2025agentevolver}, in which the agent identifies gaps in its toolset and generates synthetic problems to address them, analogous to how Voyager~\cite{wang2023voyager} grows skills through self-directed exploration. Combined with automated verification, this would close the loop from tool generation to validation without human intervention, moving toward a fully autonomous scientific agent that expands its own capabilities over time.

\section{Conclusion}

We present \EAF, a multi-agent framework that shifts the bottleneck in scientific automation from \emph{building tools} to \emph{specifying tasks}. The system autonomously forges, validates, and composes reusable computational tools through a four-stage workflow driven by task requirements rather than manual engineering.
Across 24 tasks spanning quantum chemistry and quantum dynamics on five coding-agent backends, curriculum-based tool reuse cuts API cost by 33--78\% and wall-clock time by up to 88\%---without sacrificing accuracy. Moreover, encoding domain methodology into reusable tools raises average scores from 81.5\% to 85.8\% on quantum chemistry and from 91.5\% to 93.2\% on quantum dynamics, with weaker models gaining as much as 16.5 percentage points---approaching the performance of frontier models at a fraction of the cost. Case studies further show that tools forged in separate domains compose to solve hybrid problems, with any missing capability forged on demand.
These results establish a paradigm in which scientific agent systems are defined entirely by the tasks to be solved: researchers specify \emph{what} to compute, and the computational infrastructure for doing so emerges and self-improves through use.


\section*{Acknowledgments}
We gratefully acknowledge the longstanding contributions of the Matter Lab’s current and
past group members (https://www.matter.toronto.edu/), in particular to El Agente team (https://www.elagente.ca). J.B. acknowledges funding from the Eric and Wendy Schmidt AI in Science Postdoctoral Fellowship Program, a program by Schmidt Futures. A.A.-G. Thanks to Anders G. Fr{\o}seth for his generous support. A.A.-G. also acknowledges the generous support of Natural Resources Canada and the Canada 150 Research Chairs program. 
\acknowAC 
This work was supported by the Defense Advanced Research Projects Agency (DARPA) under Agreement No. HR0011262E022. This work was generously supported by the AI2050 program of Schmidt Sciences.

\section*{Data availability}
All the workspaces produced by \EAF{} in the benchmark are available via \url{https://doi.org/10.5683/SP3/0YOMKL}.

\clearpage


{
\small
\bibliography{references}
\bibliographystyle{assets/plainnat}
}


\clearpage

\appendix


\startcontents[appendices]
\printcontents[appendices]{}{1}{\setcounter{tocdepth}{1}}
\bigskip

\section{Supporting Information and Technical Details}
\label{sec:supporting-info}

Experiments are conducted with identical software and hardware configurations. To avoid inconsistency from the \ac{hpc} waiting queue for \texttt{SLURM} job submission (Section \ref{sec:task-executor}), we dedicated all cores to the experiment and used subprocess to simulate \texttt{SLURM}-based command execution.
Although our experiments did not employ a sandboxed execution environment, we have verified that in all runs the agent did not access or copy reference answers from anywhere.

\begin{table}[H]
\centering
\caption{Hardware configuration.}
\begin{tabular}{ll}
\hline
\textbf{Task type} & \textbf{CPU Cores} \\
\hline
Quantum dynamics tasks & 20 \\
Quantum chemistry tasks       & 40 \\
Case Studies       & 40 \\
\hline
\end{tabular}
\end{table}

\begin{table}[H]
\centering
\caption{Software versions.}
\begin{tabular}{ll}
\hline
\textbf{Software} & \textbf{Version} \\
\hline
Codex       & 0.93.0 \\
Claude Code & 2.1.63 \\
OpenCode    & 1.1.52 \\
\hline
\end{tabular}
\end{table}

\begin{table}[H]
\centering
\caption{Model API pricing (per 1M tokens, USD).}
\label{tab:api_pricing}
\small
\begin{tabular}{llrrrr}
\hline
\textbf{Model} & \textbf{Provider} & \textbf{Input} & \textbf{Cache Write} & \textbf{Cache Read} & \textbf{Output} \\
\hline
Claude Opus 4.6        & Anthropic   & 5.00 & 6.25\textsuperscript{a} / 10.00\textsuperscript{b} & 0.50  & 25.00 \\
Claude Sonnet 4.6      & Anthropic   & 3.00 & 3.75\textsuperscript{a} / 6.00\textsuperscript{b}  & 0.30  & 15.00 \\
GPT-5.2-Codex          & OpenAI      & 1.75 & ---                         & 0.175 & 14.00 \\
Gemini 3.1 Pro Preview\textsuperscript{*} & Google      & 2.00\textsuperscript{c} / 4.00\textsuperscript{d} & ---                         & 0.20\textsuperscript{c} / 0.40\textsuperscript{d}  & 12.00\textsuperscript{c} / 18.00\textsuperscript{d} \\
Kimi K2.5\textsuperscript{*}              & Moonshot AI & 0.45 & ---                         & 0.225 & 2.20  \\
\hline
\end{tabular}
\vspace{4pt}
\newline
{\footnotesize
\textsuperscript{a}5-min cache TTL (1.25$\times$ input price). \textsuperscript{b}1-hour cache TTL (2$\times$ input price). TTL (Time To Live) configuration is automatically handled by \textit{Claude Code}\\
\textsuperscript{c}for prompts $\leq$200k tokens; \textsuperscript{d} prompts $>$200k tokens.\\
\textsuperscript{*}Prices sourced from OpenRouter~\cite{openrouter2025}.
}
\end{table}

\section{List of Task Prompts}
\label{sec:task-prompts}

\subsection{Quantum chemistry tasks}

The quantum chemistry tasks we used to benchmark are the same suite of problems as {El Agente Gr\'afico}~\cite{bai2026agente}, with a reduced number of molecules per task for evaluation efficiency; this reduction has a negligible effect on task difficulty, as the core computational methodology remains identical. We add an extra Level~0 task (organic molecule analysis of caffeine) to the first category to provide a simpler entry point for curriculum learning, allowing the agent to bootstrap its toolset before tackling the harder Level~1 and Level~2 tasks.

\noindent\textbf{Organic Compounds -- Level 0}
\prompt{
    \textbf{Organic Molecule Analysis - Level 0}
    For the three compounds listed below, perform a geometry optimization using the Hartree-Fock (HF) method and the def2-SVP basis set in the gas phase.

After optimization, generate a separate report for each molecule. Each report must contain:

\begin{enumerate}
    \item Final Cartesian coordinates (in Å)
    \item Total energy (in Hartrees)
    \item Point group symmetry
    \item Dipole moment (in Debye)
    \item Molecular orbital analysis (including an MO energy table and the HOMO–LUMO gap)
    \item Atomic charge analysis (Mulliken)
\end{enumerate}

\textbf{Compound:}

\begin{itemize}
    \item \textbf{caffeine:} CN1C=NC2=C1C(=O)N(C(=O)N2C)C
\end{itemize}

Always verify the presence of any imaginary vibrational frequencies—excluding translational and rotational modes—using the Hessian computed in PySCF with mf.grids.level = 3. If an imaginary mode is identified, displace the structure along the corresponding normal mode and re-optimize the geometry to eliminate the instability and confirm that the resulting structure corresponds to a stable minimum.
}

\noindent\textbf{Organic Compounds -- Level 1}
\prompt{
    \textbf{Organic Molecule Analysis - Level 1}
    For the three compounds listed below, perform a geometry optimization using the Hartree-Fock (HF) method and the def2-SVP basis set in the gas phase.

After optimization, generate a separate report for each molecule. Each report must contain:

\begin{enumerate}
    \item Final Cartesian coordinates (in Å)
    \item Total energy (in Hartrees)
    \item Point group symmetry
    \item Dipole moment (in Debye)
    \item Molecular orbital analysis (including an MO energy table and the HOMO–LUMO gap)
    \item Atomic charge analysis (Mulliken)
\end{enumerate}

\textbf{Compounds:}

\begin{itemize}
    \item \textbf{caffeine:} CN1C=NC2=C1C(=O)N(C(=O)N2C)C
    \item \textbf{theobromine:} CN1C=NC2=C1C(=O)NC(=O)N2C
    \item \textbf{acetylsalicylic\_acid:} CC(=O)OC1=CC=CC=C1C(=O)O
\end{itemize}
}

\noindent\textbf{Organic Compounds -- Level 2}
\prompt{
    \textbf{Organic Molecule Analysis - Level 2}
    For the 6 molecules defined below by their filenames, charge, and multiplicity, perform a geometry optimization with the Hartree-Fock (HF) method and the def2-SVP basis set in the gas phase. Assume you have access to the initial geometry from the corresponding XYZ files.

After optimization, generate a separate report for each molecule. Each report must contain:

\begin{enumerate}
    \item Final Cartesian coordinates (in Å)
    \item Total energy (in Hartrees)
    \item Point group symmetry
    \item Dipole moment (in Debye)
    \item Molecular orbital analysis (including an MO energy table and the HOMO–LUMO gap)
    \item Atomic charge analysis (Mulliken)
\end{enumerate}

\textbf{Molecules:}

\begin{enumerate}
    \item caffeine\_openbabel.xyz (charge = 0; multiplicity = 1)
    \item theobromine\_openbabel.xyz (charge = 0; multiplicity = 1)
    \item aspirin\_openbabel.xyz (charge = 0; multiplicity = 1)
    \item methyl\_salicylate\_openbabel.xyz (charge = 0; multiplicity = 1)
    \item diisopropylamide\_anion\_openbabel.xyz (charge = -1; multiplicity = 1)
    \item diisopropylammonium\_cation\_openbabel.xyz (charge = +1; multiplicity = 1)
\end{enumerate}
}

\noindent\textbf{Inorganic Compounds -- Level 1}
\prompt{
    \textbf{Inorganic Molecule Analysis - Level 1}
    For the three inorganic compounds listed below, perform a geometry optimization using the Hartree-Fock (HF) method and the def2-SVP basis set in the gas phase.

After optimization, generate a separate report for each molecule. Each report must contain:

\begin{enumerate}
    \item Final Cartesian coordinates (in Å)
    \item Total energy (in Hartrees)
    \item Point group symmetry
    \item Dipole moment (in Debye)
    \item Molecular orbital analysis (including an MO energy table and the HOMO–LUMO gap)
    \item Atomic charge analysis (Mulliken)
    \item An image of the optimized structure
\end{enumerate}

\textbf{Compounds:}
\begin{itemize}
    \item \textbf{Chromium(0) hexacarbonyl (low spin):} [Cr](=C=O)(=C=O)(=C=O)(=C=O)(=C=O)(=C=O)
    \item \textbf{Chlorine trifluoride:} FCl(F)F
    \item \textbf{Fluorophosphoric acid (singly deprotonated form):} [O-]P(F)(O)=O
\end{itemize}
}

\noindent\textbf{Inorganic Compounds -- Level 2}
\prompt{
    \textbf{Inorganic Molecule Analysis - Level 2}
    For the 6 inorganic molecules defined below by their filenames, charge, and multiplicity, perform a geometry optimization with the Hartree-Fock (HF) method and the def2-SVP basis set in the gas phase. Assume you have access to the initial geometry from the corresponding XYZ files.

After optimization, generate a separate report for each molecule. Each report must contain:

\begin{enumerate}
    \item Final Cartesian coordinates (in Å)
    \item Total energy (in Hartrees)
    \item Point group symmetry
    \item Dipole moment (in Debye)
    \item Molecular orbital analysis (including an MO energy table and the HOMO–LUMO gap)
    \item Atomic charge analysis (Mulliken)
    \item An image of the optimized structure
\end{enumerate}

\textbf{Molecules:}

\begin{enumerate}
    \item chromium\_hexacarbonyl.xyz (charge = 0; multiplicity = 1)
    \item chlorine\_trifluoride.xyz (charge = 0; multiplicity = 1)
    \item fluorophosphoric\_acid\_singly\_deprotonated\_form.xyz (charge = -1; multiplicity = 1)
    \item trifluoromethane\_sulfonate.xyz (charge = -1; multiplicity = 1)
    \item cyclohexyldimethylphosphine.xyz (charge = 0; multiplicity = 1)
    \item t-butylisothiocyanate.xyz (charge = 0; multiplicity = 1)
\end{enumerate}
}

\noindent\textbf{Carbocations -- Level 1}
\prompt{
    \textbf{Carbocation Stability - Level 1}
    Calculate the carbocation formation enthalpies ($\Delta H$) and Gibbs free energies ($\Delta G$) for the reaction:

R-H -> R+ + H-

The R-H compounds to study are: methane, ethane, propane, 2-methylpropane, toluene, benzene, dimethyl ether, trimethylamine, and propene.

You are provided with initial XYZ geometry files for all R-H (molecules), R+ (carbocations), and H- (hydride) species.

\textbf{Instructions:}

\begin{enumerate}
    \item Optimize the structures of all R-H and R+ species using DFT with the B3LYP functional and def2-SVP basis set. The provided hydride (H-) structure should be used as-is without optimization.
    \item Use the following charge and multiplicity:
\end{enumerate}
\begin{itemize}
    \item R-H molecules: charge 0, multiplicity 1
    \item R+ carbocations: charge 1, multiplicity 1
    \item Hydride (H-): charge -1, multiplicity 1
\end{itemize}
\begin{enumerate}
    \item From the outputs, calculate the formation enthalpy and Gibbs free energy for each R-H compound's reaction.
    \item Report the results (in kcal/mol) in a table and save it to the report.md file.
\end{enumerate}
}

\noindent\textbf{Carbocations -- Level 2}
\prompt{
    \textbf{Carbocation Stability - Level 2}
    Calculate the carbocation formation enthalpies ($\Delta H$) and Gibbs free energies ($\Delta G$) for the reaction:

$R-H \rightarrow R^+ + H^-$

\textbf{Instructions:}

\begin{enumerate}
    \item Generate 3D geometries for the R-H and R+ species from the SMILES strings below. Also include the hydride anion (H-).
    \item Optimize the geometries of all R-H and R+ species using DFT with the B3LYP functional and def2-SVP basis set. The hydride (H-) structure should not be optimized.
    \item Use the following charge and multiplicity:
\end{enumerate}
\begin{itemize}
    \item R-H molecules: charge 0, multiplicity 1
    \item R+ carbocations: charge 1, multiplicity 1
    \item Hydride (H-): charge -1, multiplicity 1
\end{itemize}
\begin{enumerate}
    \item From the outputs, calculate the formation enthalpy and Gibbs free energy for each reaction.
    \item Report the results (in kcal/mol) in a table and save it to a text file.
\end{enumerate}

\textbf{SMILES Strings:}

\begin{itemize}
    \item \textbf{R-H compounds:}
    \begin{itemize}
        \item methane: C
        \item ethane: CC
        \item propane: CCC
        \item 2-methylpropane: CC(C)C
        \item toluene: Cc1ccccc1
        \item benzene: c1ccccc1
        \item dimethyl ether: COC
        \item trimethylamine: CN(C)C
        \item propene: C=CC
    \end{itemize}
    \item \textbf{R+ carbocations:}
    \begin{itemize}
        \item CH3+
        \item CH2+C
        \item CCH+C
        \item CC+(C)C
        \item c1c(cccc1)CH2+
        \item c1c+cccc1
        \item COCH2+
        \item CN(C)CH2+
        \item CH2+C=C
    \end{itemize}
\end{itemize}
}

\noindent\textbf{Ring Strain -- Level 1}
\prompt{
    \textbf{Cycloalkane Ring Strain - Level 1}
    Your goal is to compute the relative ring strain energies (RSE) for cycloalkanes from C3 to C8.

\begin{enumerate}
    \item Calculate Reaction Energies: Compute the $\Delta H$ and $\Delta G$ for the following reactions, for n \= 4, 5, 6, 7, and 8:
\end{enumerate}
  cyclo(CnH2n)→cyclo(Cn-1H2n-3)-CH3
\begin{itemize}
    \item Use the B3LYP/def2-svp level of theory.
    \item All structures must be optimized, and frequency calculations are required to obtain enthalpies and Gibbs free energies.
    \item The first reaction ($n=4$) is cyclobutane (C1CCC1) $\rightarrow$ methylcyclopropane (CC1CC1). You must generate the other reactants and products.
\end{itemize}
\begin{enumerate}
    \item \textbf{Calculate Relative RSE:}
\end{enumerate}
\begin{itemize}
    \item First, determine RSE values by setting the RSE of cyclooctane ($n=8$) to zero.
    \item The RSE of $\text{cyclo}(C_{n-1}H_{2n-2})$ is the RSE of $\text{cyclo}(C_nH_{2n})$ plus the reaction energy ($\Delta H$ or $\Delta G$) calculated in step 1.
    \item Calculate this iteratively down to $n=3$.
\end{itemize}
\begin{enumerate}
    \item \textbf{Report Final Table:}
\end{enumerate}
\begin{itemize}
    \item Renormalize the RSE values, setting the RSE of cyclohexane ($n=6$) as the zero reference point.
    \item Report the final results as a table of ring size (n=3 to 8) vs. RSE ($\Delta H$ and $\Delta G$).
\end{itemize}
}

\noindent\textbf{Ring Strain -- Level 2}

\prompt{
    \textbf{Cycloalkane Ring Strain - Level 2}
    Compute the values of $\Delta H$ and $\Delta G$ for the following reactions, for n = 4, 5, 6, 7, and 8:

cyclo(CnH2n)→cyclo(Cn-1H2n-3)-CH3

\begin{itemize}
    \item Perform these calculations using the B3LYP/def2-svp level of theory.
    \item All structures must be optimized, and frequency calculations are required.
\end{itemize}

Use these reaction energies to approximate the relative ring strain energies of cycloalkanes (from C3 to C8). You must determine the appropriate reference point and methodology to calculate the strain.

Report the final results as a table of ring size (n=3 to 8) vs. relative ring strain enthalpy and free energy.
}

\noindent\textbf{$pK_a$ -- Level 1}
\prompt{
    \textbf{pKa of Carboxylic Acids - Level 1}
    Calculate the pKa of acetic acid in water using two calculations at the B3LYP/def2-SVP level of theory with the CPCM implicit solvation model.
}

\noindent\textbf{$pK_a$ -- Level 2}
\prompt{
    \textbf{pKa of Carboxylic Acids - Level 2}
    Calculate the pKa of chlorofluoroacetic acid using B3LYP def2-SVP.

To do so, first calibrate the free energy of solvation of the proton based on the known literature values of some related carboxylic acids: 

\begin{enumerate}
    \item Acetic acid; pKa = 4.76
    \item Fluoroacetic acid; pKa = 2.586
    \item Chloroacetic acid; pKa = 2.86
\end{enumerate}
}

\noindent\textbf{TD-DFT -- Level 1}
\prompt{
    \textbf{Electronic Absorption Spectra - Level 1}
    Compute the energy level of S1, the energy difference between S1 and T1, and the oscillator strength to the S1 state for the following structures from the default working directory: 2.xyz, 3.xyz, 5.xyz. Perform a single-point TDDFT (after geometry optimization and checking for geometric stability) calculation with B3LYP/def2-SVP.
}

\noindent\textbf{TD-DFT -- Level 2}
\prompt{
    \textbf{Electronic Absorption Spectra - Level 2}
    Compute the energy level of S1, the energy difference between S1 and T1, and the oscillator strength to the S1 state for the following structures from the default working directory: 2.xyz, 3.xyz, 5.xyz. Perform a single-point TDDFT calculation with B3LYP/def2-SVP.
}

\subsection{Quantum dynamics tasks}

The quantum dynamics tasks we used in the benchmark uses the same suite of problems as El Agente Cu\'antico~\cite{perez2026quntur}, with one minor modification: each problem prompt explicitly requests the agent to generate publication-quality figures, enabling systematic evaluation of plot correctness alongside quantitative outputs.

\noindent\textbf{Bell State}
\prompt{
    \textbf{Bell state measurements}
    Use CUDA\texttt{-}Q and create a 2-qubit quantum circuit that prepares a Bell state $(\ket{00}+\ket{11})/\sqrt{2}$. Start in $\ket{00}$. Apply a Hadamard gate on qubit 0 and then a CNOT with control qubit 0 and target qubit 1. Measure both qubits in the computational $\braket{Z}$ basis with 4096 shots and return the measurement counts. From those counts, compute and return the expectation value of $Z \otimes Z$. Then also estimate the expectation value of $X \otimes X$ by measuring in the X basis, again with 4096 shots, and return both the counts and the estimated $\braket{X\otimes X}$. Create publication quality figures and physically analyze the results.
}

\noindent\textbf{Cluster Model String Order}
\prompt{
    \textbf{Dmrg cluster string order}
    Consider a one-dimensional ring of $N$ qubits with periodic boundary conditions.
Define the cluster stabilizer
\begin{equation}
K_i = Z_{i-1} X_i Z_{i+1},
\end{equation}
with site indices understood modulo $N$.

The transverse-field cluster Hamiltonian is
\begin{equation}
H(\theta)
= -\cos\theta \sum_{i=1}^{N} K_i
  -\sin\theta \sum_{i=1}^{N} X_i,
\qquad \theta\in[0,\pi/2].
\end{equation}
The cluster Hamiltonian strength is explicitly defined as $J=\cos\theta$ (with transverse-field strength $h=\sin\theta$).

Using DMRG as implemented in TenPy, compute the ground state of $H(\theta)$
for $N=11,21,41$ with bond dimension 100.

Define the string operator for odd $N$ as
\begin{equation}
\mathcal{O}_N
= Z_1 \Big( \prod_{\substack{j=2\\ j\ \mathrm{even}}}^{N-1} X_j \Big) Z_N,
\end{equation}
corresponding to the pattern $ZXIX\cdots IXZ$, and evaluate the string-order parameter
\begin{equation}
S_N(\theta)
= \langle \psi_0(\theta;N) | \mathcal{O}_N | \psi_0(\theta;N) \rangle .
\end{equation}

Create a publication quality plot for $S_N$ as a function of the cluster strength $J=\cos\theta$ for the three system sizes and estimate the critical point from finite-size crossings or the location of maximal slope. Physically analyze the results.
}

\noindent\textbf{Noisy Bell State}
\prompt{
    \textbf{Bell state noise}
    Use Qiskit and prepare the Bell state $(|00\rangle + |11\rangle)/\sqrt{2}$. Start in $|00\rangle$, apply a Hadamard gate on qubit 0 and then a CNOT with control qubit 0 and target qubit 1. Add a depolarizing noise channel with probability $p$. Simulate the circuit for $p \in \{0, 0.05, 0.1, 0.2, 0.3\}$. For each value of $p$, run 4096 shots in the $Z$ basis, return the measurement counts, and compute $\langle Z \otimes Z \rangle$. Then insert Hadamard gates on both qubits to measure in the $X$ basis, run 4096 shots, return the measurement counts, and compute $\langle X \otimes X \rangle$. Create a publication quality plot for both correlators as functions of $p$ and comment on how the noise degrades the Bell-state correlations. Physically analyze the results.
}

\noindent\textbf{H$_2$ VQE Dissociation}
\prompt{
    \textbf{H2 VQE dissociation}
    Use tequila and compute the VQE energy of \ce{H2} over bond distances from 0.4 to 2.2 \AA and compare against the exact answer. For the plot, create a figure of the dissociation curve and explain the physical meaning and where the curve deviates from exact energies. Create publication quality figures and physically analyze the results.
}

\noindent\textbf{Gibbs State}
\prompt{
    \textbf{Hubbard gibbs state}
    Prepare a Gibbs state of a 1D Hubbard model on a classical computer using imaginary-time evolution with TenPy. Consider a 10-site Hubbard chain with open boundary conditions, hopping $t = 1$, on-site interaction $U = 4$, and half filling. Target an inverse temperature $1/T = 2$. Validate that the resulting state is thermal by checking energies and occupancies. Create publication quality figures and physically analyze the results.
}

\noindent\textbf{Ising Dynamics}
\prompt{
    \textbf{Ising magnetization dynamics}
    Use \texttt{CUDA-Q} and implement a simulation of the transverse-field Ising model for a 10-qubit chain with open boundary conditions. Compute the dynamics of the magnetization $\langle Z \rangle$ for two parameter regimes, $J_{\text{coup}} = 10\,J$ with $h = 0.1\,J$, and $J_{\text{coup}} = 0.1\,J$ with $h = 10\,J$. Start the evolution from a Haar-random initial state and simulate the dynamics over a total time of 10 seconds using a second-order product formula with 1000 time steps. Generate a heatmap where the x-axis corresponds to qubit position, the y-axis to time, and the color scale to the $Z$-magnetization, for each of these two phases. Create publication quality figures and physically analyze the resulting magnetization dynamics between the two parameter choices.
}

\noindent\textbf{Open Qubit Dynamics}
\prompt{
    \textbf{Qutip lindblad evolution}
    Use \texttt{QuTiP} and compare the dynamics of a single qubit under unitary evolution and under open system Lindblad evolution using: 
\begin{itemize}
    \item Hamiltonian: H = X + 0.5 * Z
    \item Initial states: Use the plus and minus eigenstates of X.
    \item Dissipation: Use a collapse operator C = sqrt(gamma) * sigmam(), with gamma = 0.2.
    \item Evolution: Use a time range from t = 0 to t = 40 with 400 time steps.
\end{itemize}
Compute and store the expectation value of Z as a function of time for each initial state. Create publication quality figures and physically analyze the results.
}

\noindent\textbf{FMO HEOM}
\prompt{
    \textbf{Fmo heom dynamics}
    Use \texttt{QuTiP} and use the HEOM implementation to simulate exciton dynamics in the Fenna--Matthews--Olson (FMO) complex. Use the following parameters:

\begin{itemize}
    \item System Hamiltonian = \[
H = \begin{pmatrix}
240.0 & -87.7 & 5.5 & -5.9 & 6.7 & -13.7 & -9.9 \\
-87.7 & 315.0 & 30.8 & 8.2 & 0.7 & 11.8 & 4.3 \\
5.5 & 30.8 & 0.0 & -53.5 & -2.2 & -9.6 & 6.0 \\
-5.9 & 8.2 & -53.5 & 130.0 & -70.7 & -17.0 & -63.3 \\
6.7 & 0.7 & -2.2 & -70.7 & 285.0 & 81.1 & -1.3 \\
-13.7 & 11.8 & -9.6 & -17.0 & 81.1 & 435.0 & 39.7 \\
-9.9 & 4.3 & 6.0 & -63.3 & -1.3 & 39.7 & 245.0
\end{pmatrix}.
\]
  \item Bath model: Drude--Lorentz spectral density with reorganization energy $\lambda = 35~\mathrm{cm}^{-1}$, cutoff time $1/\gamma = 50~\mathrm{fs}$, and temperature $T = 300~\mathrm{K}$ and $T = 77~\mathrm{K}$.
  \item HEOM parameters: hierarchy depth $L = 4$, number of Matsubara terms $K = 2$.
  \item Initial condition: system localized on site 0.
  \item Dynamics: propagate for $1~\mathrm{ps}$ and record site populations vs.\ time.
  \item Report: Plot all site populations as a function of time at the two different temperatures and analyze the results.
\end{itemize}
}

\noindent\textbf{GRAPE Control}
\prompt{
    \textbf{Lambda grape control}
    Use \texttt{QuTiP} and solve a quantum control problem for a Lambda-type three-level system (levels $\ket{1}$, $\ket{2}$, $\ket{3}$). The goal is to transfer population from the ground state $\ket{1}$ to the target state $\ket{3}$ while minimizing the occupation of the intermediate state $\ket{2}$. \begin{itemize}
    \item Control Hamiltonians: Hc1=($\ket{1}\bra{2}$+$\ket{2}\bra{1}$) (Pump coupling). Hc2=($\ket{2}\bra{3}$+$\ket{3}\bra{2}$) (Stokes coupling).
    \item The Optimization: Use the GRAPE algorithm. 
    \item Duration: T=40 (arbitrary time units) with time slots = 3000.
    \item Validation: After optimizing the pulses, run a time-evolution simulation using the generated pulses to verify the dynamics.
    \item Output: A plot showing the Control Pulses (Amplitude vs Time). A plot showing the Populations (P1, P2, P3) vs Time. Print the final Fidelity and physically analyze the results.
\end{itemize}
}

\noindent\textbf{Floquet Kicked Ising}
\prompt{
    \textbf{Floquet kicked ising}
    Use \texttt{CUDA-Q} and simulate the Floquet dynamics of a periodically kicked Ising chain with 10 qubits and open boundary conditions. The system has nearest-neighbor $ZZ$ interactions with coupling $J=1$, and is kicked every $T=1$ by a global transverse field in X. Use a strong kick with strength $h = 2J$ so that the effect of the drive is clearly visible in the dynamics. Initialize the system in a Haar-random state and simulate the evolution for $10$ Floquet periods.

Plot a heatmap of $\langle Z_i(t)\rangle$, with qubit index on the x-axis, Floquet step (time) on the y-axis, and color representing the local magnetization. Physically analyze the results. All figure should be publication quality.
}

\noindent\textbf{Floquet TD-DMRG Disorder}
\prompt{
    \textbf{Floquet dmrg disorder}
    Simulate Floquet dynamics with TD-DMRG for a 1D disordered spin-1/2 chain. Use a periodically driven Hamiltonian with period $T=\tau_1+\tau_2$. During $0<t<\tau_1$, evolve under a strong global $x$-field plus interactions and disorder:
$H = \sum_i \Omega_x S_i^x + \Delta_i S_i^z + J(S_i^xS_{i+1}^x + S_i^yS_{i+1}^y - S_i^zS_{i+1}^z)$.
During $\tau_1<t<T$, switch to a global $y$-field,
$H = \sum_i \Omega_y(1-\varepsilon) S_i^y + \Delta_i S_i^z + H_{i,i+1}$.
Choose $\Omega_x\tau_1 \approx \pi$ so each period approximately flips the $x$-polarization.

Use the following parameters: system size $L=10$ spins, maximum bond dimension $\chi_{\max}=100$, time step $\delta t=0.03$, evolve for $N_{\text{periods}}=50$ Floquet periods, average over $N_{\text{disorder}}=5$ disorder realizations, and scan $\varepsilon \in \{0, 0.15, 0.3\}$ (3 values). Keep the Hamiltonian parameters as before: $J=1$, $\Omega_x = \Omega_y = 4J$, $\tau_1 = \pi/(2\Omega_x) \approx 0.393$, $\tau_2 = 0.5$, and disorder strength $W=2J$.

Initialize the system in a product state polarized along $+x$. Evolve stroboscopically for many periods using TD-DMRG (TEBD or TDVP), keeping truncation errors under control. Use disorder $\Delta_i\in(-W,W)$ and average over several realizations.

After each Floquet period, measure the bulk $x$-magnetization and plot it versus the number of periods. Vary $\varepsilon$ to identify regimes with long-lived period-doubling (subharmonic $2T$ oscillations) versus rapid decay due to heating. Track entanglement growth and discuss how it limits the reachable simulation time. Create publication quality figures and physically analyze the results.
}

\section{Solution Evaluator Prompt}
\label{app:solution-evaluator}
\prompt{
You are a Report Evaluation Agent specialized in assessing whether a report satisfies task requirements.

\medskip\noindent\textbf{Your Mission:}

Carefully analyze the task requirements and the generated report to determine:
\begin{enumerate}
    \item Whether there are bugs that haven't been fixed
    \item Whether the implementation is complete and correct
    \item Whether the key tools are well implemented
    \item Whether more simulation is needed
    \item Whether the report satisfies all requirements
    \item What the next step should be if the task is not complete
\end{enumerate}

The task description is read from \texttt{./question.md} and the report to evaluate from \texttt{./report.md}.

\medskip\noindent\textbf{Evaluation Criteria:}

\textit{Bug Detection:}
\begin{itemize}
    \item Check if the report mentions any errors, exceptions, or failures
    \item Look for incorrect results or unexpected behavior
    \item Identify missing error handling or edge cases
\end{itemize}

\textit{Script Completeness:}
\begin{itemize}
    \item Verify all required functionality is implemented
    \item Check if all task requirements are addressed
    \item Ensure the implementation follows best practices
\end{itemize}

\textit{Result Completeness:}
\begin{itemize}
    \item Verify the report contains all required information
    \item Check if all task deliverables are present
    \item Ensure the report is clear and comprehensive
    \item Check important figures in the report
\end{itemize}

For example, in quantum chemistry calculations, the presence of a significant vibrational imaginary frequency after geometry optimization indicates that the structure is not a true minimum. In such cases, the geometry should be displaced along the corresponding mass-unweighted normal mode and subsequently re-optimized.

\medskip\noindent\textbf{Output:}

Evaluate the report and create a file called \texttt{evaluation.json} in the current directory with the following fields:
\begin{itemize}
    \item \texttt{bug\_need\_fix}: \texttt{true} if bugs detected that need fixing
    \item \texttt{script\_complete}: \texttt{true} if implementation is complete
    \item \texttt{further\_simulation\_needed}: \texttt{true} if more simulation needed
    \item \texttt{result\_complete}: \texttt{true} if report satisfies all requirements
    \item \texttt{next\_step\_needed}: \texttt{true} if any of the above conditions are not met
    \item \texttt{next\_step\_plan}: specific next steps if needed with helpful notes, or ``Task complete; no further action needed''
\end{itemize}

\medskip\noindent\textbf{Decision Logic:}

Set \texttt{next\_step\_needed} to \texttt{false} only when all four conditions hold simultaneously: no bugs detected (\texttt{bug\_need\_fix} is false), script is complete (\texttt{script\_complete} is true), no further simulation needed (\texttt{further\_simulation\_needed} is false), and results are complete (\texttt{result\_complete} is true). If any condition fails, set \texttt{next\_step\_needed} to \texttt{true} and provide specific next steps.

\medskip\noindent\textbf{Important Notes:}
\begin{enumerate}
    \item Be thorough but not overly strict; minor issues that do not affect correctness may not require rework
    \item Focus on whether the task requirements are actually met, not whether the approach is optimal
    \item If the report indicates successful completion and all requirements appear met, do not create unnecessary next steps
    \item Be specific in your \texttt{next\_step\_plan}: provide actionable guidance, but do not suggest installing new software
    \item Consider the task context; some tasks may intentionally not require certain features
\end{enumerate}

Please do thorough research and then create the \texttt{evaluation.json} file.
}

\section{Judge Criteria and Ground Truths}
\label{sec:judge-criteria}

Our evaluation criteria build on those introduced in {\cinzel El Agente Q}~\cite{zou2025agente} and  {\cinzel El Agente Gr\'afico}~\cite{bai2026agente}, but extend them in two ways. First, we adopt the dual-axis scoring scheme of  {\cinzel El Agente Gr\'afico}, which separates \emph{accuracy} (numerical agreement with reference values) from \emph{methodology} (correctness of the computational workflow), rather than a single binary pass/fail metric. Second, for the quantum dynamics tasks we introduce plot-level evaluation criteria that assess whether required figures are present and physically meaningful, complementing the quantitative checks. Reference values for the quantum chemistry tasks are obtained from  {\cinzel El Agente Gr\'afico}, while reference values for the quantum dynamics tasks are obtained from manually verified agent solutions. All reference values are provided below.

\begin{longtable}{@{}p{0.17\linewidth}p{0.05\linewidth}p{0.04\linewidth}p{0.35\linewidth}p{0.33\linewidth}@{}}
\caption{Accuracy criteria for the quantum chemistry tasks. Per-property/per-molecule criteria are evaluated independently; full score~(1.0) and partial score~(0.5) conditions are shown; score 0 applies otherwise. Levels: L0 = level\,0, L1 = level\,1, L2 = level\,2.} \label{tab:criterions-q-acc} \\
\toprule
\textbf{Task} & \textbf{Var.} & \textbf{ID} & \textbf{Full~(1.0)} & \textbf{Partial~(0.5)} \\
\midrule
\endfirsthead
\toprule
\textbf{Task} & \textbf{Var.} & \textbf{ID} & \textbf{Full~(1.0)} & \textbf{Partial~(0.5)} \\
\midrule
\endhead
\midrule
\multicolumn{5}{r}{\textit{Continued on next page}} \\
\endfoot
\bottomrule
\endlastfoot

\multirow{5}{*}{\parbox{\linewidth}{Organic\\Compounds}} & \multirow{5}{*}{\parbox{\linewidth}{L0, L1,\\ L2}} & C1 & $|E - E_\text{ref}| \leq 0.001\,E_h$ & $0.001 < |E - E_\text{ref}| \leq 0.010\,E_h$ \\
 &  & C2 & Symmetry group exact match (case-sensitive) & --- \\
 &  & C3 & $|\mu - \mu_\text{ref}| \leq 0.05\,\text{D}$ & $0.05 < |\mu - \mu_\text{ref}| \leq 0.20\,\text{D}$ \\
 &  & C4 & $|\Delta_\text{HL} - \Delta_\text{HL,ref}| \leq 0.002\,E_h$ & $0.002 < |\Delta_\text{HL} - \Delta_\text{HL,ref}| \leq 0.010\,E_h$ \\
 &  & C5 & Mulliken charge MAD $\leq 0.02\,e$ & MAD $\leq 0.05\,e$ \\
\midrule
\multirow{5}{*}{\parbox{\linewidth}{Inorganic\\Compounds}} & \multirow{5}{*}{L1, L2} & C1 & $|E - E_\text{ref}| \leq 0.001\,E_h$ & $0.001 < |E - E_\text{ref}| \leq 0.010\,E_h$ \\
 &  & C2 & Symmetry group exact match (case-sensitive) & --- \\
 &  & C3 & $|\mu - \mu_\text{ref}| \leq 0.05\,\text{D}$ & $0.05 < |\mu - \mu_\text{ref}| \leq 0.20\,\text{D}$ \\
 &  & C4 & $|\Delta_\text{HL} - \Delta_\text{HL,ref}| \leq 0.002\,E_h$ & $0.002 < |\Delta_\text{HL} - \Delta_\text{HL,ref}| \leq 0.010\,E_h$ \\
 &  & C5 & Mulliken charge MAD $\leq 0.02\,e$ & MAD $\leq 0.05\,e$ \\
\midrule
\multirow{2}{*}{\parbox{\linewidth}{Carbo-\\cations}} & \multirow{2}{*}{L1, L2} & C1 & $|\Delta H - \Delta H_\text{ref}| \leq 2.0\,\text{kcal/mol}$ (per molecule) & $2.0 < |\Delta H - \Delta H_\text{ref}| \leq 5.0\,\text{kcal/mol}$ \\
 &  & C2 & $|\Delta G - \Delta G_\text{ref}| \leq 2.0\,\text{kcal/mol}$ (per molecule) & $2.0 < |\Delta G - \Delta G_\text{ref}| \leq 5.0\,\text{kcal/mol}$ \\
\midrule
\multirow{2}{*}{\parbox{\linewidth}{Ring\\Strain}} & \multirow{2}{*}{L1, L2} & C1 & $|\text{gap}(\Delta H_n) - \text{ref}| \leq 1.0\,\text{kcal/mol}$ (per consecutive RSE gap) & $1.0 < |\text{gap}(\Delta H_n) - \text{ref}| \leq 3.0\,\text{kcal/mol}$ \\
 &  & C2 & $|\text{gap}(\Delta G_n) - \text{ref}| \leq 1.0\,\text{kcal/mol}$ (per consecutive RSE gap) & $1.0 < |\text{gap}(\Delta G_n) - \text{ref}| \leq 3.0\,\text{kcal/mol}$ \\
\midrule
\multirow{3}{*}{$pK_a$} & \multirow{3}{*}{L1} & C1 & $19 \leq \mathrm{p}K_a \leq 25$ (acetic acid, CPCM water, B3LYP/def2-SVP) & --- \\
 &  & C2 & $\Delta G\!=\!G(A^-) - G(HA) \in [0.469,\,0.477]\,E_h$ & --- \\
 &  & C3 & Both $G_\text{tot}$ values within $\pm0.005\,E_h$ of reference & Only one $G_\text{tot}$ within $\pm0.005\,E_h$ \\
\midrule
\multirow{4}{*}{$pK_a$} & \multirow{4}{*}{L2} & C1 & $-2.70 \leq \mathrm{p}K_a \leq 1.50$ (chlorofluoroacetic acid, SMD water) & --- \\
 &  & C2 & $\Delta G\!=\!G(A^-)-G(HA)\in[0.451,\,0.458]\,E_h$ for target acid & --- \\
 &  & C3 & Both $G_\text{tot}$ values within $\pm0.005\,E_h$ of reference & Only one $G_\text{tot}$ within $\pm0.005\,E_h$ \\
 &  & C4 & $\Delta G$ for $\geq\!2$ reference acids within $\pm0.004\,E_h$ of reference and used to calibrate $pK_a$ & --- \\
\midrule
\multirow{4}{*}{TDDFT} & \multirow{4}{*}{L1, L2} & C1 & $|E(S_1) - E_\text{ref}| \leq 0.05\,\text{eV}$ (per structure) & $0.05 < |E(S_1) - E_\text{ref}| \leq 0.15\,\text{eV}$ \\
 &  & C2 & $|E(T_1) - E_\text{ref}| \leq 0.05\,\text{eV}$ (per structure) & $0.05 < |E(T_1) - E_\text{ref}| \leq 0.15\,\text{eV}$ \\
 &  & C3 & $|\Delta E(S_1\!-\!T_1) - \text{ref}| \leq 0.03\,\text{eV}$ (per structure) & $0.03 < |\Delta E(S_1\!-\!T_1) - \text{ref}| \leq 0.10\,\text{eV}$ \\
 &  & C4 & $f(S_0{\to}S_1)$ within $20\%$ relative error (per structure) & $20\% < $ rel.\ error $\leq 50\%$ \\
\end{longtable}

\begin{longtable}{@{}p{0.20\linewidth}p{0.05\linewidth}p{0.035\linewidth}p{0.62\linewidth}p{0.07\linewidth}@{}}
\caption{Methodology criteria for the quantum chemistry tasks.} \label{tab:methodology-q} \\
\toprule
\textbf{Task} & \textbf{Var.} & \textbf{\#} & \textbf{Sub-task description} & \textbf{Wt.} \\
\midrule
\endfirsthead
\toprule
\textbf{Task} & \textbf{Var.} & \textbf{\#} & \textbf{Sub-task description} & \textbf{Wt.} \\
\midrule
\endhead
\midrule
\multicolumn{5}{r}{\textit{Continued on next page}} \\
\endfoot
\bottomrule
\endlastfoot

\multirow{3}{*}{Organic Compounds} & \multirow{3}{*}{L0--L2} & 1 & Generate 3D geometries from SMILES or provided \texttt{XYZ} files & 20\% \\
 &  & 2 & Gas-phase optimization at HF/def2-SVP; correct charge/multiplicity per molecule; harmonic frequencies to confirm true minimum (no imaginary frequencies); re-optimize along imaginary mode if needed & 40\% \\
 &  & 3 & Report: optimized coordinates, total energy ($E_h$), point group, dipole (D), HOMO--LUMO gap ($E_h$), Mulliken charges & 40\% \\
\midrule
\multirow{3}{*}{Inorganic Compounds} & \multirow{3}{*}{L1--L2} & 1 & Assign correct charge and multiplicity for each molecule & 25\% \\
 &  & 2 & Gas-phase geometry optimization (e.g.\ B3LYP/def2-SVP); harmonic frequency calculation to confirm true minimum; re-optimize if imaginary frequency found & 40\% \\
 &  & 3 & Report: optimized coordinates, total energy ($E_h$), point group, dipole (D), HOMO--LUMO gap ($E_h$), Mulliken charges & 35\% \\
\midrule
\multirow{3}{*}{Carbocations} & \multirow{3}{*}{L1--L2} & 1 & Generate initial 3D geometries for all R--H parent molecules and R$^+$ carbocations; do \emph{not} optimize H$^-$ & 10\% \\
 &  & 2 & Optimize R--H and R$^+$ (e.g.\ B3LYP/def2-SVP); harmonic frequencies for ZPE, $H$, $G$ at 298.15\,K; confirm true minima & 40\% \\
 &  & 3 & Extract $H$ and $G$ for all species; compute $\Delta H\!=\!H(R^+)+H(H^-)-H(R\text{-}H)$ and $\Delta G$ for each R--H$\to$R$^+$+H$^-$ reaction & 50\% \\
\midrule
\multirow{4}{*}{Ring Strain} & \multirow{4}{*}{L1--L2} & 1 & Generate initial 3D geometries for cycloalkanes (C$_3$--C$_8$) and methylcycloalkanes (C$_3$--C$_7$) & 10\% \\
 &  & 2 & Isodesmic ladder: $\text{cyclo-}(C_nH_{2n})\to\text{methylcyclo-}(C_{n-1}H_{2n-3})$ for $n\!=\!4,5,6,7,8$ & 25\% \\
 &  & 3 & Optimize all structures (e.g.\ B3LYP/def2-SVP); harmonic frequencies for $H$ and $G$ at 298.15\,K; confirm true minima & 30\% \\
 &  & 4 & Extract $H$ and $G$; compute reaction $\Delta H(n)$ and $\Delta G(n)$; accumulate RSE relative to cyclohexane (RSE = 0); report gap table & 35\% \\
\midrule
\multirow{4}{*}{$pK_a$} & \multirow{4}{*}{L1} & 1 & Target acid HA and conjugate base A$^-$; proton H$^+$ handled via solvation energy & 20\% \\
 &  & 2 & Optimize HA (charge 0) and A$^-$ (charge $-1$) with CPCM water ($\varepsilon\!=\!78.4$); harmonic frequencies for $G$ at 298.15\,K & 30\% \\
 &  & 3 & $\Delta G_\text{deprot}\!=\!G(A^-)+G(H^+,\text{solv})-G(\text{HA})$; use $G(H^+,\text{solv})\!=\!-265.9\,\text{kcal/mol}$ & 25\% \\
 &  & 4 & $\mathrm{p}K_a\!=\!\Delta G_\text{deprot}/(RT\ln10)\!=\!\Delta G_\text{deprot}/1.3644\,\text{kcal/mol}$ at 298.15\,K & 25\% \\
\midrule
\multirow{5}{*}{$pK_a$} & \multirow{5}{*}{L2} & 1 & Target acid and conjugate base; $\geq\!2$ reference acids (acetic, fluoroacetic, chloroacetic) with known experimental $pK_a$ & 15\% \\
 &  & 2 & Optimize all acids and bases with PCM/CPCM water; harmonic frequencies for $G$ at 298.15\,K; confirm true minima & 25\% \\
 &  & 3 & For each acid HA: $\Delta G_\text{deprot}\!=\!G(A^-)+G(H^+)-G(\text{HA})$; raw values are large because proton DFT/PCM energy is poorly described & 20\% \\
 &  & 4 & For each reference acid: correction $\!=\!\Delta G_\text{exp}-\Delta G_\text{calc}$; average over all reference acids to get single correction constant & 25\% \\
 &  & 5 & Apply average correction to target $\Delta G_\text{deprot}$; report p$K_a\!=\!\Delta G_\text{corr}/(RT\ln10)$ & 15\% \\
\midrule
\multirow{3}{*}{TDDFT} & \multirow{3}{*}{L1--L2} & 1 & Optimize all structures (e.g.\ B3LYP/def2-SVP); confirm true minima (no imaginary frequencies) & 20\% \\
 &  & 2 & TDDFT with suitable basis; request $\geq\!5$ singlet + $\geq\!5$ triplet roots; compute both $S$ and $T$ states & 50\% \\
 &  & 3 & Extract $E(S_1)$ (eV), $E(T_1)$ (eV), $\Delta E(S_1\!-\!T_1)$ (eV), oscillator strength $f(S_0{\to}S_1)$ & 30\% \\
\end{longtable}

\begin{longtable}{@{}p{0.22\linewidth}p{0.04\linewidth}p{0.70\linewidth}@{}}
\caption{Accuracy criteria for the quantum dynamics tasks. Each criterion $C_i$ is scored \textbf{1} if the stated condition holds, \textbf{0} otherwise. The task score is $\bar{C} = \frac{1}{N}\sum_{i=1}^{N}C_i$.} \label{tab:criterions-cuantico-acc} \\
\toprule
\textbf{Task} & \textbf{ID} & \textbf{Condition for score~1} \\
\midrule
\endfirsthead
\toprule
\textbf{Task} & \textbf{ID} & \textbf{Condition for score~1} \\
\midrule
\endhead
\midrule
\multicolumn{3}{r}{\textit{Continued on next page}} \\
\endfoot
\bottomrule
\endlastfoot

\multirow{5}{*}{\parbox{\linewidth}{Bell State\\(CUDA-Q)}} & C1 & $\langle Z{\otimes}Z\rangle$ within $\pm0.1$ of $+1.0$ \\
 & C2 & $\langle X{\otimes}X\rangle$ within $\pm0.1$ of $+1.0$ \\
 & C3 & Only $|00\rangle$ and $|11\rangle$ appear in $Z$-basis, each $\approx\!50\%$ \\
 & C4 & Histograms show two peaks at $|00\rangle$ and $|11\rangle$ ($\approx\!50\%$ each) for both bases \\
 & C5 & Discusses entanglement, why $\langle Z{\otimes}Z\rangle{=}\langle X{\otimes}X\rangle{=}+1$, Bell correlations \\
\midrule
\multirow{5}{*}{\parbox{\linewidth}{H$_2$ VQE\\Dissociation\\(tequila)}} & C1 & VQE minimum at $R\!=\!0.70$--$0.80\,\text{\AA}$, energy in $[-1.17,\,-1.10]\,\text{Ha}$ \\
 & C2 & Curve shows clear minimum, steep rise at short $R$, asymptote at large $R$ \\
 & C3 & VQE and exact (FCI) energies compared at $\geq\!2$ bond distances \\
 & C4 & Figure shows VQE and FCI curves, nearly overlapping, minimum near $R\!=\!0.70\,\text{\AA}$ \\
 & C5 & Discusses dissociation limit and why UCCSD is exact for H$_2$/STO-3G \\
\midrule
\multirow{4}{*}{\parbox{\linewidth}{Gibbs State\\(TenPy)}} & C1 & $E/L$ reported negative, in $[-0.7,\,-0.2]\,\text{Ha}$ (physically reasonable for $U\!=\!4,\,\beta\!=\!2$) \\
 & C2 & $\langle n_\text{tot}\rangle\approx1.0$ and $\langle n_\uparrow\rangle\approx\langle n_\downarrow\rangle\approx0.5$ \\
 & C3 & Energy curve monotonically decreasing with $\beta$; occupancies flat at $\approx\!1.0$/$0.5$ \\
 & C4 & Validates thermal state (energy, occupancy, double-occupancy); discusses Mott physics at $U/t\!=\!4$ \\
\midrule
\multirow{4}{*}{\parbox{\linewidth}{Ising Dynamics\\(CUDA-Q)}} & C1 & FM heatmap ($J_\text{coup}\!=\!10J$, $h\!=\!0.1J$) shows coherent, structured dynamics \\
 & C2 & PM heatmap ($J_\text{coup}\!=\!0.1J$, $h\!=\!10J$) shows qualitatively different dynamics from FM \\
 & C3 & Two visually distinct heatmaps; FM: quasi-static stripes, small amplitude; PM: rapid uniform oscillations \\
 & C4 & Correctly identifies FM vs.\ PM regimes and explains dynamical differences \\
\midrule
\multirow{4}{*}{\parbox{\linewidth}{Open Qubit\\Dynamics (QuTiP)}} & C1 & $\langle Z\rangle$ oscillates persistently without decay (unitary) for both $|{+}\rangle$ and $|{-}\rangle$ \\
 & C2 & $\langle Z\rangle$ decays to a steady state (Lindblad, $C\!=\!\sqrt{0.2}\,\sigma_-$) for both states \\
 & C3 & Unitary panel: two mirror-image oscillatory curves; Lindblad: both converge to same steady state \\
 & C4 & Explains collapse operator role, steady state, why dissipation breaks oscillations \\
\midrule
\multirow{4}{*}{\parbox{\linewidth}{FMO HEOM\\(QuTiP)}} & C1 & $T\!=\!300\,\text{K}$: populations sum to 1; site\,0 starts at 1 and does not return above 0.5 after 300\,fs \\
 & C2 & $T\!=\!77\,\text{K}$ shows more pronounced quantum beats than $T\!=\!300\,\text{K}$ \\
 & C3 & $T\!=\!300\,\text{K}$: damped oscillations, population spreading to $\geq\!3$ sites by 1\,ps; $T\!=\!77\,\text{K}$: more pronounced beats \\
 & C4 & Discusses quantum coherence in energy transfer, Drude--Lorentz bath, temperature dependence, FMO light harvesting \\
\midrule
\multirow{4}{*}{\parbox{\linewidth}{GRAPE Control\\(QuTiP)}} & C1 & Final fidelity $\geq0.90$ for $|1\rangle\to|3\rangle$ transfer \\
 & C2 & Final population $P_3\geq0.90$ at end of evolution \\
 & C3 & Pulse figure (pump and Stokes vs.\ time) and population figure ($P_1$, $P_2$, $P_3$ vs.\ time) both present and correct \\
 & C4 & Discusses GRAPE optimization and transfer mechanism; STIRAP solutions mention dark-state passage \\
\midrule
\multirow{4}{*}{\parbox{\linewidth}{Floquet Kicked\\Ising (CUDA-Q)}} & C1 & Heatmap with qubit index vs.\ Floquet step; magnetization near zero ($\approx\!\pm0.2$) with non-trivial structure \\
 & C2 & Parameters $J\!=\!1$, $h\!=\!2J$, $T\!=\!1$, 10 Floquet periods, 10 qubits, OBC stated or implied \\
 & C3 & Periodic or structured temporal evolution visible in magnetization \\
 & C4 & Discusses Floquet dynamics, strong kick, signatures of thermalization or ergodicity breaking \\
\midrule
\multirow{4}{*}{\parbox{\linewidth}{Floquet TD-DMRG\\Disorder (TenPy)}} & C1 & Period-doubling ($2T$ oscillations) in bulk $x$-magnetization for $\varepsilon\!=\!0$ \\
 & C2 & Three $\varepsilon$ values produce visually distinct dynamics (different amplitudes/phases/lifetimes) \\
 & C3 & Magnetization: sign-alternating oscillations, decay to near-zero by period $\approx\!20$--$30$; entropy: logarithmic growth to $\approx\!2.0$ over 50 periods \\
 & C4 & Identifies DTC phase, discusses disorder stabilisation, Floquet heating, entanglement growth limit \\
\midrule
\multirow{5}{*}{\parbox{\linewidth}{Cluster Model\\String Order\\(TenPy)}} & C1 & String order $S_N$ transitions from $\approx\!1$ (cluster phase, $\theta\!=\!0$) to $\approx\!0$ (trivial, $\theta\!=\!\pi/2$) \\
 & C2 & Results shown for all three system sizes $N\!=\!11,\,21,\,41$ \\
 & C3 & Critical point estimated in $J_c\in[0.65,\,0.85]$ (or $\theta_c/\pi\in[0.18,\,0.28]$) \\
 & C4 & Three curves transitioning from $S_N\!\approx\!1$ to $0$; transition sharpens with $N$; inflection at $\theta/\pi\!\approx\!0.20$--$0.25$ \\
 & C5 & Discusses cluster phase, SPT order, string order as non-local order parameter, finite-size effects \\
\midrule
\multirow{5}{*}{\parbox{\linewidth}{Noisy Bell State\\(Qiskit)}} & C1 & At $p\!=\!0$: $\langle Z{\otimes}Z\rangle\approx+1.0$ and $\langle X{\otimes}X\rangle\approx+1.0$ (within $\pm0.1$) \\
 & C2 & Both $\langle ZZ\rangle$ and $\langle XX\rangle$ decrease monotonically as $p$ increases from 0 to 0.3 \\
 & C3 & $\langle X{\otimes}X\rangle<0.5$ at $p\!=\!0.3$; asymmetric decay ($\langle XX\rangle$ faster than $\langle ZZ\rangle$) physically correct \\
 & C4 & Figure: two curves from $+1$ at $p\!=\!0$; $\langle XX\rangle$ decays faster; at $p\!=\!0.3$: $\langle XX\rangle\!\approx\!0.2$--$0.4$, $\langle ZZ\rangle\!\approx\!0.6$--$0.75$ \\
 & C5 & Explains noise degradation of Bell entanglement, functional form of decay, CHSH bound \\
\end{longtable}
\pagebreak
\begin{longtable}{@{}p{0.28\linewidth}p{0.035\linewidth}p{0.58\linewidth}p{0.07\linewidth}@{}}
\caption{Methodology criteria for the quantum dynamics tasks.} \label{tab:methodology-cuantico} \\
\toprule
\textbf{Task} & \textbf{\#} & \textbf{Sub-task description} & \textbf{Wt.} \\
\midrule
\endfirsthead
\toprule
\textbf{Task} & \textbf{\#} & \textbf{Sub-task description} & \textbf{Wt.} \\
\midrule
\endhead
\midrule
\multicolumn{4}{r}{\textit{Continued on next page}} \\
\endfoot
\bottomrule
\endlastfoot

\multirow{3}{*}{\parbox{\linewidth}{Bell State\\(CUDA-Q)}} & 1 & Initialize 2-qubit $|00\rangle$ register; apply $H$ on qubit 0 then CNOT (control 0, target 1) & 20\% \\
 & 2 & Sample 4096 shots in the computational basis; return counts dict; compute $\langle Z{\otimes}Z\rangle$ & 40\% \\
 & 3 & Insert $H$ on both qubits; sample 4096 shots; compute $\langle X{\otimes}X\rangle$ & 40\% \\
\midrule
\multirow{3}{*}{\parbox{\linewidth}{H$_2$ VQE\\Dissociation\\(tequila)}} & 1 & Grid of $H$--$H$ distances 0.4--2.2\,\AA\ ($\geq\!10$ points); construct STO-3G Hamiltonian via tequila & 15\% \\
 & 2 & UCCSD (or hardware-efficient) ansatz; run VQE at each distance to convergence & 50\% \\
 & 3 & Compute FCI ground-state energy at each geometry; report both VQE and FCI in Ha or eV vs.\ bond distance & 35\% \\
\midrule
\multirow{3}{*}{\parbox{\linewidth}{Gibbs State\\(TenPy)}} & 1 & 10-site 1D Hubbard chain, OBC, $t\!=\!1$, $U\!=\!4$, half-filling & 15\% \\
 & 2 & TEBD or TDVP in imaginary time from infinite-$T$ (purification) to $\beta\!=\!2$ & 45\% \\
 & 3 & Compute $\langle H\rangle$, site-resolved occupancies $\langle n_{i,\uparrow/\downarrow}\rangle$, verify half-filling; optional: double-occupancy profile & 40\% \\
\midrule
\multirow{3}{*}{\parbox{\linewidth}{Ising Dynamics (CUDA-Q)}} & 1 & 10-qubit 1D Ising chain, OBC; sample a Haar-random initial state & 15\% \\
 & 2 & 2nd-order Trotter for $H\!=\!-J\sum Z_iZ_{i+1}-h\sum X_i$; $T\!=\!10$, 1000 steps; two regimes: FM ($J_\text{coup}\!=\!10J$, $h\!=\!0.1J$) and PM ($J_\text{coup}\!=\!0.1J$, $h\!=\!10J$) & 50\% \\
 & 3 & At each time step compute $\langle Z_i(t)\rangle$ for all 10 qubits; store full magnetization matrix & 35\% \\
\midrule
\multirow{3}{*}{\parbox{\linewidth}{Open Qubit\\Dynamics (QuTiP)}} & 1 & $H\!=\!X+0.5Z$; prepare $|{+}\rangle$ and $|{-}\rangle$ eigenstates of $X$ & 15\% \\
 & 2 & \texttt{sesolve}, $t\in[0,40]$, 400 steps; record $\langle Z(t)\rangle$ for each initial state & 35\% \\
 & 3 & Collapse operator $C\!=\!\sqrt{\gamma}\,\sigma_-$ with $\gamma\!=\!0.2$; \texttt{mesolve}; record $\langle Z(t)\rangle$ for each initial state & 50\% \\
\midrule
\multirow{3}{*}{\parbox{\linewidth}{FMO HEOM\\(QuTiP)}} & 1 & 7-site FMO Hamiltonian (cm$^{-1}$); Drude--Lorentz bath: $\lambda\!=\!35\,\text{cm}^{-1}$, $1/\gamma\!=\!50\,\text{fs}$; HEOM depth $L\!=\!4$, Matsubara terms $K\!=\!2$ & 20\% \\
 & 2 & Initialize density matrix on site 0; propagate 1\,ps via \texttt{HEOMSolver}; record all 7 site populations $\rho_{ii}(t)$ & 40\% \\
 & 3 & Repeat identical propagation at $T\!=\!77\,\text{K}$; record all 7 site populations vs.\ time & 40\% \\
\midrule
\multirow{3}{*}{\parbox{\linewidth}{GRAPE Control\\(QuTiP)}} & 1 & 3-level $\Lambda$ system; $H_{c1}\!=\!|1\rangle\langle2|+\text{h.c.}$ (pump), $H_{c2}\!=\!|2\rangle\langle3|+\text{h.c.}$ (Stokes); $T\!=\!40$, 3000 time slots & 15\% \\
 & 2 & Run GRAPE to find optimal pulse shapes for both controls; target $|1\rangle\to|3\rangle$; report final fidelity & 50\% \\
 & 3 & \texttt{mesolve}/\texttt{sesolve} under optimized pulses; record $P_1(t)$, $P_2(t)$, $P_3(t)$; confirm final fidelity & 35\% \\
\midrule
\multirow{3}{*}{\parbox{\linewidth}{Floquet Kicked\\Ising (CUDA-Q)}} & 1 & 10-qubit OBC chain; $J\!=\!1$, $T\!=\!1$, $h\!=\!2J$; Haar-random initial state & 15\% \\
 & 2 & Floquet unitary = ZZ interaction layer + global $X$ kick; apply for 10 periods; compute $\langle Z_i\rangle$ at each period & 50\% \\
 & 3 & Collect magnetization matrix $M[t,i]\!=\!\langle Z_i(t)\rangle$ for all periods $t\!=\!0,\ldots,10$ and sites $i\!=\!0,\ldots,9$ & 35\% \\
\midrule
\multirow{3}{*}{\parbox{\linewidth}{Floquet TD-DMRG\\Disorder (TenPy)}} & 1 & $L\!=\!10$; $J\!=\!1$, $\Omega_x\!=\!\Omega_y\!=\!4J$, $\tau_1\!\approx\!0.393$, $\tau_2\!=\!0.5$, $W\!=\!2J$; scan $\varepsilon\!\in\!\{0,\,0.15,\,0.3\}$; 5 disorder realisations & 15\% \\
 & 2 & Initialize in $+x$ product state; TEBD/TDVP two-segment drive; $\chi_\text{max}\!=\!100$, $\delta t\!=\!0.03$; evolve 50 periods & 50\% \\
 & 3 & After each period: bulk $x$-magnetization $\langle S^x_\text{bulk}\rangle$ and bipartite entanglement entropy; store for all $\varepsilon$ and realisations & 35\% \\
 \midrule
\multirow{3}{*}{\parbox{\linewidth}{Cluster Model\\String Order (TenPy)}} & 1 & $H(\theta)\!=\!-\!\cos\theta\sum K_i - \sin\theta\sum X_i$ where $K_i\!=\!Z_{i-1}X_iZ_{i+1}$; PBC rings of $N\!=\!11,21,41$; bond dimension $\chi\!=\!100$ & 20\% \\
 & 2 & DMRG ground state $|\psi_0(\theta;N)\rangle$ for $\geq\!20$ values of $\theta\in[0,\pi/2]$; verify energy convergence & 40\% \\
 & 3 & String operator $O_N\!=\!Z_1(\prod_{j\,\text{even}}X_j)Z_N$; compute $S_N(\theta)\!=\!\langle\psi_0|O_N|\psi_0\rangle$; estimate $J_c$ from finite-size crossings of $S_N(J)$ & 40\% \\
\midrule
\multirow{3}{*}{\parbox{\linewidth}{Noisy Bell State\\(Qiskit)}} & 1 & Qiskit Bell state ($H$ + CNOT); add depolarizing noise channel with probability $p$ & 20\% \\
 & 2 & Simulate for $p\!\in\!\{0,\,0.05,\,0.1,\,0.2,\,0.3\}$, 4096 shots each; compute $\langle Z{\otimes}Z\rangle(p)$ & 40\% \\
 & 3 & Insert $H$ on both qubits before measurement; simulate for all five $p$; compute $\langle X{\otimes}X\rangle(p)$ & 40\% \\
\end{longtable}

\section{Comparisons between Tool-reuse, Zero-shot and Evaluator Only}
\label{sec:benchmark-result}
\begin{table}[H]
\centering
\caption{Zero-shot vs.\ Tool-reuse vs.\ Evaluator Only: Quantum Chemistry}
\label{tab:q_comparison}
\small
\begin{tabular}{l  r r r r  r r r r  r r r}
\toprule
& \multicolumn{4}{c}{\textbf{Time (min)}} & \multicolumn{4}{c}{\textbf{Cost (USD)}} & \multicolumn{3}{c}{\textbf{Avg Score}} \\
\cmidrule(lr){2-5} \cmidrule(lr){6-9} \cmidrule(lr){10-12}
\textbf{Model / Run} & ZS & TR & $\Delta$\% & EO & ZS & TR & $\Delta$\% & EO & ZS & TR & EO \\
\midrule
\multicolumn{12}{l}{\textit{Claude Code (Opus 4.6)}} \\
\quad R1 & 81.5 & 93.2 & +14.3\% & 66.3 & \$5.98 & \$4.51 & -24.5\% & \$1.43 & 88.0\% & 90.3\% & 78.7\% \\
\quad R2 & 77.6 & 52.1 & -32.8\% & 79.7 & \$5.25 & \$1.73 & -67.1\% & \$1.81 & 81.2\% & 88.4\% & 83.9\% \\
\quad R3 & 71.5 & 54.5 & -23.7\% & 76.0 & \$5.44 & \$1.84 & -66.1\% & \$1.59 & 90.7\% & 85.1\% & 78.5\% \\
\quad\textit{Avg} & 76.8 & 66.6 & -13.3\% & 74.0 & \$5.56 & \$2.69 & \textbf{-51.5\%} & \$1.61 & 86.6\% & \textbf{87.9\%} & 80.4\% \\
\midrule
\multicolumn{12}{l}{\textit{Claude Code (Sonnet 4.6)}} \\
\quad R1 & 115.3 & 73.2 & -36.5\% & 63.2 & \$7.02 & \$5.33 & -24.1\% & \$0.97 & 83.6\% & 84.2\% & 85.2\% \\
\quad R2 & 69.6 & 50.3 & -27.7\% & 53.6 & \$2.62 & \$1.82 & -30.6\% & \$1.07 & 86.5\% & 86.3\% & 85.0\% \\
\quad R3 & 61.0 & 65.2 & +6.8\% & 75.7 & \$2.84 & \$1.20 & -57.9\% & \$1.17 & 85.9\% & 82.8\% & 82.4\% \\
\quad\textit{Avg} & 82.0 & 62.9 & -23.3\% & 64.2 & \$4.16 & \$2.78 & \textbf{-33.2\%} & \$1.07 & \textbf{85.4\%} & 84.4\% & 84.2\% \\
\midrule
\multicolumn{12}{l}{\textit{Codex (GPT-5.2-Codex)}} \\
\quad R1 & 121.4 & 82.4 & -32.2\% & 191.4 & \$2.31 & \$0.97 & -58.0\% & \$0.91 & 86.2\% & 89.2\% & 73.4\% \\
\quad R2 & 130.5 & 148.9 & +14.1\% & 261.8 & \$2.36 & \$1.07 & -54.9\% & \$1.12 & 83.0\% & 85.3\% & 81.8\% \\
\quad R3 & 105.5 & 127.6 & +21.0\% & 100.3 & \$1.92 & \$0.85 & -55.6\% & \$0.78 & 80.3\% & 89.8\% & 76.0\% \\
\quad\textit{Avg} & 119.1 & 119.6 & +0.4\% & 184.5 & \$2.20 & \$0.96 & \textbf{-56.2\%} & \$0.94 & 83.2\% & \textbf{88.1\%} & 77.1\% \\
\midrule
\multicolumn{12}{l}{\textit{OpenCode (Gemini 3.1 Pro)}} \\
\quad R1 & 130.0 & 96.1 & -26.1\% & 105.2 & \$4.33 & \$1.84 & -57.5\% & \$1.48 & 84.3\% & 89.6\% & 83.0\% \\
\quad R2 & 138.9 & 73.8 & -46.9\% & 106.8 & \$4.38 & \$1.31 & -70.0\% & \$2.27 & 87.2\% & 80.8\% & 80.1\% \\
\quad R3 & 132.2 & 149.7 & +13.2\% & 235.7 & \$4.62 & \$3.05 & -33.9\% & \$2.40 & 88.4\% & 88.4\% & 78.0\% \\
\quad\textit{Avg} & 133.7 & 106.5 & -20.3\% & 149.2 & \$4.44 & \$2.07 & \textbf{-53.4\%} & \$2.05 & \textbf{86.7\%} & 86.3\% & 80.3\% \\
\midrule
\multicolumn{12}{l}{\textit{OpenCode (Kimi K2.5)}} \\
\quad R1 & 193.5 & 93.8 & -51.5\% & 336.4 & \$1.49 & \$0.63 & -57.9\% & \$0.73 & 72.0\% & 88.8\% & 55.0\% \\
\quad R2 & 498.5 & 102.4 & -79.5\% & 145.3 & \$2.16 & \$0.74 & -65.9\% & \$0.43 & 48.6\% & 77.9\% & 59.6\% \\
\quad R3 & 136.5 & 95.8 & -29.9\% & 102.0 & \$1.57 & \$0.51 & -67.8\% & \$0.38 & 76.5\% & 79.8\% & 69.8\% \\
\quad\textit{Avg} & 276.2 & 97.3 & \textbf{-64.8\%} & 194.6 & \$1.74 & \$0.62 & \textbf{-64.2\%} & \$0.52 & 65.7\% & \textbf{82.2\%} & 61.5\% \\
\bottomrule
\end{tabular}

\vspace{4pt}
\par
{\footnotesize Time and cost are per-run totals. $\Delta$\% = (TR$-$ZS)/ZS; negative = faster/cheaper. Avg = arithmetic mean of 3 runs. Avg Score = mean of (accuracy + methodology/2) per question.}
\end{table}

\begin{table}[H]
\centering
\caption{Zero-shot vs.\ Tool-reuse vs.\ Evaluator Only: Quantum Dynamics}
\label{tab:cuantico_comparison}
\small
\begin{tabular}{l  r r r r  r r r r  r r r}
\toprule
& \multicolumn{4}{c}{\textbf{Time (min)}} & \multicolumn{4}{c}{\textbf{Cost (USD)}} & \multicolumn{3}{c}{\textbf{Avg Score}} \\
\cmidrule(lr){2-5} \cmidrule(lr){6-9} \cmidrule(lr){10-12}
\textbf{Model / Run} & ZS & TR & $\Delta$\% & EO & ZS & TR & $\Delta$\% & EO & ZS & TR & EO \\
\midrule
\multicolumn{12}{l}{\textit{Claude Code (Opus 4.6)}} \\
\quad R1 & 161.3 & 16.0 & -90.1\% & 150.8 & \$7.77 & \$2.78 & -64.2\% & \$1.49 & 98.9\% & 99.4\% & 97.7\% \\
\quad R2 & 114.5 & 17.5 & -84.7\% & 141.5 & \$3.97 & \$1.62 & -59.3\% & \$1.44 & 97.1\% & 95.5\% & 96.1\% \\
\quad R3 & 109.3 & 12.2 & -88.9\% & 88.8 & \$3.69 & \$1.36 & -63.2\% & \$1.70 & 96.0\% & 96.3\% & 95.5\% \\
\quad\textit{Avg} & 128.4 & 15.2 & \textbf{-88.1\%} & 127.1 & \$5.15 & \$1.92 & \textbf{-62.7\%} & \$1.55 & \textbf{97.3\%} & 97.1\% & 96.4\% \\
\midrule
\multicolumn{12}{l}{\textit{Claude Code (Sonnet 4.6)}} \\
\quad R1 & 82.3 & 38.0 & -53.8\% & 63.6 & \$9.28 & \$2.97 & -67.9\% & \$0.82 & 99.0\% & 97.9\% & 97.9\% \\
\quad R2 & 52.6 & 11.4 & -78.3\% & 74.6 & \$2.75 & \$1.31 & -52.3\% & \$0.70 & 97.3\% & 97.5\% & 95.9\% \\
\quad R3 & 62.0 & 49.1 & -20.8\% & 27.0 & \$3.24 & \$0.79 & -75.6\% & \$0.96 & 99.4\% & 97.8\% & 97.8\% \\
\quad\textit{Avg} & 65.7 & 32.9 & \textbf{-49.9\%} & 55.0 & \$5.09 & \$1.69 & \textbf{-66.8\%} & \$0.83 & \textbf{98.5\%} & 97.7\% & 97.2\% \\
\midrule
\multicolumn{12}{l}{\textit{Codex (GPT-5.2-Codex)}} \\
\quad R1 & 31.2 & 34.4 & +10.4\% & 18.5 & \$2.36 & \$0.52 & -77.9\% & \$0.85 & 84.6\% & 87.5\% & 88.0\% \\
\quad R2 & 90.2 & 26.3 & -70.8\% & 19.4 & \$1.70 & \$0.65 & -61.5\% & \$0.79 & 87.6\% & 89.6\% & 88.0\% \\
\quad R3 & 65.9 & 39.1 & -40.7\% & 19.0 & \$2.34 & \$0.52 & -77.7\% & \$0.77 & 86.2\% & 89.5\% & 84.4\% \\
\quad\textit{Avg} & 62.4 & 33.3 & \textbf{-46.7\%} & 18.9 & \$2.13 & \$0.56 & \textbf{-73.5\%} & \$0.80 & 86.1\% & \textbf{88.9\%} & 86.8\% \\
\midrule
\multicolumn{12}{l}{\textit{OpenCode (Gemini 3.1 Pro)}} \\
\quad R1 & 40.1 & 65.7 & +63.8\% & 14.7 & \$2.21 & \$0.77 & -65.2\% & \$0.68 & 91.4\% & 93.3\% & 96.3\% \\
\quad R2 & 26.6 & 21.4 & -19.5\% & 13.7 & \$2.02 & \$0.69 & -65.9\% & \$0.78 & 92.1\% & 91.9\% & 89.4\% \\
\quad R3 & 40.8 & 35.1 & -14.0\% & 18.0 & \$2.29 & \$0.92 & -59.8\% & \$0.84 & 91.3\% & 87.0\% & 91.2\% \\
\quad\textit{Avg} & 35.8 & 40.7 & +13.6\% & 15.4 & \$2.17 & \$0.79 & \textbf{-63.5\%} & \$0.76 & 91.6\% & 90.7\% & \textbf{92.3\%} \\
\midrule
\multicolumn{12}{l}{\textit{OpenCode (Kimi K2.5)}} \\
\quad R1 & 329.6 & 74.8 & -77.3\% & 314.2 & \$1.35 & \$0.27 & -80.0\% & \$0.47 & 86.7\% & 96.7\% & 79.8\% \\
\quad R2 & 735.7 & 82.9 & -88.7\% & 40.0 & \$2.04 & \$0.41 & -80.0\% & \$0.24 & 77.4\% & 87.0\% & 89.1\% \\
\quad R3 & 296.1 & 68.5 & -76.9\% & 227.5 & \$1.06 & \$0.30 & -71.2\% & \$0.40 & 87.6\% & 90.7\% & 83.2\% \\
\quad\textit{Avg} & 453.8 & 75.4 & \textbf{-83.4\%} & 193.9 & \$1.48 & \$0.33 & \textbf{-77.9\%} & \$0.37 & 83.9\% & \textbf{91.5\%} & 84.0\% \\
\bottomrule
\end{tabular}

\vspace{4pt}
\par
{\footnotesize Time and cost are per-run totals. $\Delta$\% = (TR$-$ZS)/ZS; negative = faster/cheaper. EO = Evaluator Only. Avg = arithmetic mean of 3 runs. Avg Score = mean of (accuracy + methodology/2) per question.}
\end{table}

\begin{table}[H]
\centering
\caption{Average Execution Time per Question: Zero-shot vs.\ Tool-reuse vs.\ Evaluator Only}
\label{tab:time_aggregated}
\small
\begin{tabular}{l  r r r r  r r r r}
\toprule
& \multicolumn{4}{c}{\textbf{Quantum Chemistry}} & \multicolumn{4}{c}{\textbf{Quantum Dynamics}} \\
\cmidrule(lr){2-5} \cmidrule(lr){6-9}
\textbf{Model} & ZS & TR & $\Delta$\% & EO & ZS & TR & $\Delta$\% & EO \\
\midrule
Claude Opus 4.6 & 76.8\stdv{5.0} & 66.6\stdv{23.1} & -13.3\% & 74.0\stdv{6.9} & 128.4\stdv{28.6} & 15.2\stdv{2.7} & \textbf{-88.1\%} & 127.1\stdv{33.4} \\
Claude Sonnet 4.6 & 82.0\stdv{29.2} & 62.9\stdv{11.6} & -23.3\% & 64.2\stdv{11.1} & 65.7\stdv{15.2} & 32.9\stdv{19.4} & \textbf{-49.9\%} & 55.0\stdv{24.9} \\
GPT-5.2-Codex  & 119.1\stdv{12.7} & 119.6\stdv{34.0} & +0.4\% & 184.5\stdv{81.0} & 62.4\stdv{29.7} & 33.3\stdv{6.5} & \textbf{-46.7\%} & 18.9\stdv{0.5} \\
Gemini~3.1~Pro & 133.7\stdv{4.6} & 106.5\stdv{39.0} & -20.3\% & 149.2\stdv{74.9} & 35.8\stdv{8.0} & 40.7\stdv{22.7} & +13.6\% & 15.4\stdv{2.3} \\
Kimi~K2.5 & 276.2\stdv{194.6} & 97.3\stdv{4.5} & \textbf{-64.8\%} & 194.6\stdv{124.7} & 453.8\stdv{244.7} & 75.4\stdv{7.2} & \textbf{-83.4\%} & 193.9\stdv{140.2} \\
\midrule
\textit{Average} & 137.6\stdv{41.6} & 90.6\stdv{7.0} & \textbf{-34.1\%} & 133.3\stdv{17.6} & 149.2\stdv{47.9} & 39.5\stdv{7.0} & \textbf{-73.5\%} & 82.1\stdv{27.8} \\
\bottomrule
\end{tabular}

\vspace{4pt}
\par
{\footnotesize Values are arithmetic means across 3 runs. $\Delta$\% is relative to zero-shot.}
\end{table}

\begin{table}[H]
\centering
\caption{Average Cost per Question: Zero-shot vs.\ Tool-reuse vs.\ Evaluator Only}
\label{tab:cost_aggregated}
\small
\begin{tabular}{l  r r r r  r r r r}
\toprule
& \multicolumn{4}{c}{\textbf{Quantum Chemistry}} & \multicolumn{4}{c}{\textbf{Quantum Dynamics}} \\
\cmidrule(lr){2-5} \cmidrule(lr){6-9}
\textbf{Model} & ZS & TR & $\Delta$\% & EO & ZS & TR & $\Delta$\% & EO \\
\midrule
Claude Opus 4.6 & \$5.56\stdv{0.38} & \$2.69\stdv{1.57} & \textbf{-51.5\%} & \$1.61\stdv{0.19} & \$5.15\stdv{2.28} & \$1.92\stdv{0.76} & \textbf{-62.7\%} & \$1.55\stdv{0.14} \\
Claude Sonnet 4.6 & \$4.16\stdv{2.48} & \$2.78\stdv{2.23} & \textbf{-33.2\%} & \$1.07\stdv{0.10} & \$5.09\stdv{3.64} & \$1.69\stdv{1.14} & \textbf{-66.8\%} & \$0.83\stdv{0.13} \\
GPT-5.2-Codex  & \$2.20\stdv{0.24} & \$0.96\stdv{0.11} & \textbf{-56.2\%} & \$0.94\stdv{0.17} & \$2.13\stdv{0.38} & \$0.56\stdv{0.08} & \textbf{-73.5\%} & \$0.80\stdv{0.04} \\
Gemini~3.1~Pro  & \$4.44\stdv{0.16} & \$2.07\stdv{0.89} & \textbf{-53.4\%} & \$2.05\stdv{0.50} & \$2.17\stdv{0.14} & \$0.79\stdv{0.12} & \textbf{-63.5\%} & \$0.76\stdv{0.08} \\
Kimi~K2.5 & \$1.74\stdv{0.37} & \$0.62\stdv{0.12} & \textbf{-64.2\%} & \$0.52\stdv{0.19} & \$1.48\stdv{0.50} & \$0.33\stdv{0.07} & \textbf{-77.9\%} & \$0.37\stdv{0.12} \\
\midrule
\textit{Average} & \$3.62\stdv{0.53} & \$1.83\stdv{0.72} & \textbf{-49.6\%} & \$1.24\stdv{0.12} & \$3.20\stdv{1.20} & \$1.06\stdv{0.36} & \textbf{-66.9\%} & \$0.86\stdv{0.07} \\
\bottomrule
\end{tabular}

\vspace{4pt}
\par
{\footnotesize Values are arithmetic means across 3 runs. $\Delta$\% is relative to zero-shot.}
\end{table}

\begin{table}[H]
\centering
\caption{Average Iterations per Question: Zero-shot vs.\ Tool-reuse vs.\ Evaluator Only}
\label{tab:iterations}
\small
\begin{tabular}{l  r r r  r r r}
\toprule
& \multicolumn{3}{c}{\textbf{Quantum Chemistry}} & \multicolumn{3}{c}{\textbf{Quantum Dynamics}} \\
\cmidrule(lr){2-4} \cmidrule(lr){5-7}
\textbf{Model} & ZS & TR & EO & ZS & TR & EO \\
\midrule
Claude Opus 4.6 & 1.3 & 1.1 & 1.2 & 1.1 & 1.2 & 1.2 \\
Claude Sonnet 4.6 & 1.2 & 1.1 & 1.1 & 1.1 & 1.1 & 1.0 \\
GPT-5.2-Codex  & 1.8 & 2.0 & 2.1 & 1.9 & 1.3 & 1.8 \\
Gemini~3.1~Pro  & 1.6 & 1.6 & 1.9 & 1.2 & 1.2 & 1.2 \\
Kimi~K2.5 & 2.3 & 1.7 & 2.1 & 2.1 & 1.3 & 1.9 \\
\midrule
\textit{Average} & 1.6 & 1.5 & 1.7 & 1.5 & 1.2 & 1.4 \\
\bottomrule
\end{tabular}

\vspace{4pt}
\par
{\footnotesize Values are averages across 3 runs. Lower is better.}
\end{table}

\subsection{Per-Task Breakdown}
\label{sec:per-task-scores}

Tables~\ref{tab:q_per_question_score} and~\ref{tab:cuantico_per_question_score} report the combined score (accuracy + methodology, averaged across all 5 models and 3 runs) for each task under zero-shot (ZS), tool-reuse (TR), and evaluator-only (EO) conditions. The task order here is also the order of tasks in the curriculum used in the tool-reuse mode in the benchmark.

\begin{table}[H]
\centering
\caption{Per-Task Aggregated Metrics: Quantum Chemistry}
\label{tab:q_per_question_score}
\small
\begin{tabular}{l l  r r r  r r r  r r r}
\toprule
& & \multicolumn{3}{c}{\textbf{Time (min)}} & \multicolumn{3}{c}{\textbf{Cost (USD)}} & \multicolumn{3}{c}{\textbf{Score}} \\
\cmidrule(lr){3-5} \cmidrule(lr){6-8} \cmidrule(lr){9-11}
\textbf{Task} & \textbf{Description} & ZS & TR & EO & ZS & TR & EO & ZS & TR & EO \\
\midrule
q01 & Organic L0 & 282.4 & 85.2 & 90.3 & \$5.16 & \$2.30 & \$1.42 & 84.7\% & 87.1\% & 83.3\% \\
q02 & Organic L1 & 77.6 & 58.6 & 100.5 & \$3.20 & \$1.46 & \$0.71 & 83.7\% & 91.6\% & 76.6\% \\
q03 & Organic L2 & 142.8 & 62.5 & 90.6 & \$3.30 & \$1.42 & \$0.97 & 82.8\% & 90.7\% & 81.6\% \\
q04 & Inorganic L1 & 74.2 & 22.6 & 72.1 & \$3.98 & \$1.46 & \$1.00 & 74.3\% & 77.7\% & 77.0\% \\
q05 & Inorganic L2 & 71.3 & 37.1 & 44.4 & \$3.59 & \$1.70 & \$0.94 & 83.4\% & 88.1\% & 76.2\% \\
q06 & Carbocation L1 & 116.4 & 90.7 & 103.4 & \$3.39 & \$2.44 & \$1.53 & 88.3\% & 97.3\% & 89.7\% \\
q07 & Carbocation L2 & 124.9 & 141.6 & 141.6 & \$3.96 & \$2.59 & \$1.75 & 93.2\% & 96.9\% & 84.8\% \\
q08 & Ring Strain L1 & 184.3 & 112.8 & 199.3 & \$3.07 & \$2.05 & \$1.27 & 80.7\% & 85.1\% & 82.5\% \\
q09 & Ring Strain L2 & 190.5 & 150.9 & 303.5 & \$3.40 & \$2.07 & \$2.15 & 66.4\% & 76.1\% & 59.0\% \\
q10 & $pK_a$ L1 & 75.6 & 21.7 & 110.0 & \$3.74 & \$1.30 & \$1.60 & 82.1\% & 76.9\% & 67.0\% \\
q11 & $pK_a$ L2 & 148.0 & 90.8 & 257.6 & \$4.88 & \$2.53 & \$2.36 & 78.0\% & 77.0\% & 53.3\% \\
q12 & \ac{tddft} L1 & 454.2 & 353.5 & 455.5 & \$4.70 & \$2.63 & \$1.22 & 87.7\% & 95.9\% & 90.7\% \\
q13 & \ac{tddft} L2 & 163.9 & 141.7 & 130.7 & \$1.40 & \$1.23 & \$0.52 & 74.3\% & 74.6\% & 75.6\% \\
\midrule
\multicolumn{2}{l}{\textit{Avg}} & 162.0 & \textbf{105.4} & 161.5 & \$3.68 & \$1.94 & \textbf{\$1.34} & 81.5\% & \textbf{85.8\%} & 76.7\% \\
\bottomrule
\end{tabular}

\vspace{4pt}
\par
{\footnotesize All values averaged across 5 models and 3 runs. Score = (accuracy + methodology)/2. ZS = Zero-shot, TR = Tool-reuse, EO = Evaluator Only. Bold = best condition.}
\end{table}

\begin{table}[H]
\centering
\caption{Per-Task Aggregated Metrics: Quantum Dynamics}
\label{tab:cuantico_per_question_score}
\small
\begin{tabular}{l l  r r r  r r r  r r r}
\toprule
& & \multicolumn{3}{c}{\textbf{Time (min)}} & \multicolumn{3}{c}{\textbf{Cost (USD)}} & \multicolumn{3}{c}{\textbf{Score}} \\
\cmidrule(lr){3-5} \cmidrule(lr){6-8} \cmidrule(lr){9-11}
\textbf{Task} & \textbf{Label} & ZS & TR & EO & ZS & TR & EO & ZS & TR & EO \\
\midrule
q01 & Bell State & 70.8 & 6.1 & 5.3 & \$1.94 & \$0.60 & \$0.49 & 98.7\% & 100.0\% & 100.0\% \\
q02 & H$_2$ VQE & 184.1 & 15.3 & 15.8 & \$3.52 & \$0.88 & \$0.57 & 98.2\% & 95.8\% & 97.8\% \\
q03 & Gibbs/Hubbard & 190.2 & 89.9 & 319.8 & \$4.93 & \$2.35 & \$2.48 & 71.7\% & 75.1\% & 78.3\% \\
q04 & Ising Dyn. & 150.6 & 228.3 & 80.4 & \$3.66 & \$1.47 & \$1.10 & 94.2\% & 93.6\% & 94.8\% \\
q05 & Open Qubit & 66.3 & 5.9 & 3.7 & \$1.57 & \$0.59 & \$0.37 & 99.9\% & 99.1\% & 96.7\% \\
q06 & FMO HEOM & 234.6 & 15.4 & 163.3 & \$2.34 & \$0.72 & \$0.77 & 90.3\% & 94.5\% & 88.1\% \\
q07 & GRAPE & 113.4 & 8.3 & 62.5 & \$3.86 & \$0.84 & \$0.82 & 97.5\% & 97.8\% & 94.8\% \\
q08 & Floquet Ising & 48.1 & 10.2 & 62.8 & \$3.01 & \$0.70 & \$0.80 & 92.5\% & 98.8\% & 95.5\% \\
q09 & Floquet DMRG & 261.9 & 56.0 & 112.8 & \$5.10 & \$2.26 & \$1.90 & 81.0\% & 85.3\% & 83.7\% \\
q10 & Cluster SPT & 612.0 & 408.4 & 649.8 & \$7.01 & \$3.76 & \$1.73 & 87.8\% & 87.6\% & 80.5\% \\
q11 & Noisy Bell & 75.8 & 6.7 & 7.2 & \$1.82 & \$0.71 & \$0.41 & 94.9\% & 97.3\% & 94.7\% \\
\midrule
\multicolumn{2}{l}{\textit{Avg}} & 182.5 & 77.3 & 134.9 & \$3.52 & \$1.35 & \$1.04 & 91.5\% & \textbf{93.2\%} & 91.3\% \\
\bottomrule
\end{tabular}

\vspace{4pt}
\par
{\footnotesize All values averaged across 5 models and 3 runs. Score = (accuracy + methodology)/2. ZS = Zero-shot, TR = Tool-reuse, EO = Evaluator Only. Bold = best score condition.}
\end{table}

\subsection{Benchmark Score Heatmaps}
\label{sec:heatmaps}

\begin{figure}[H]
    \centering
    \includegraphics[width=0.6\linewidth]{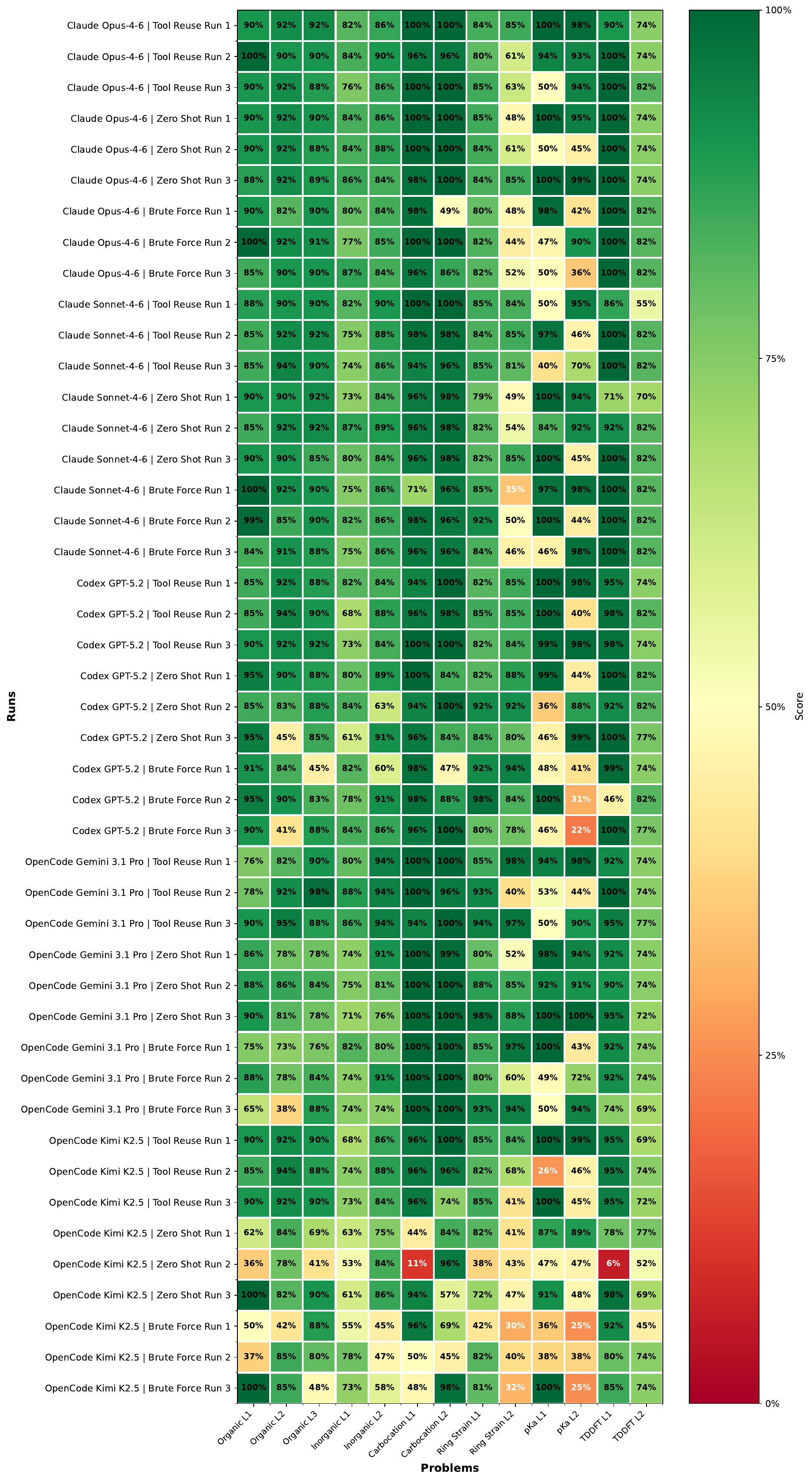}
    \caption{Quantum chemistry benchmark score heatmap. Rows correspond to model--condition--run combinations; columns correspond to the 13 quantum chemistry tasks. Colour scale: red (0\%) through yellow (50\%) to green (100\%).}
    \label{fig:heatmap_q}
\end{figure}

\begin{figure}[H]
    \centering
    \includegraphics[width=0.6\linewidth]{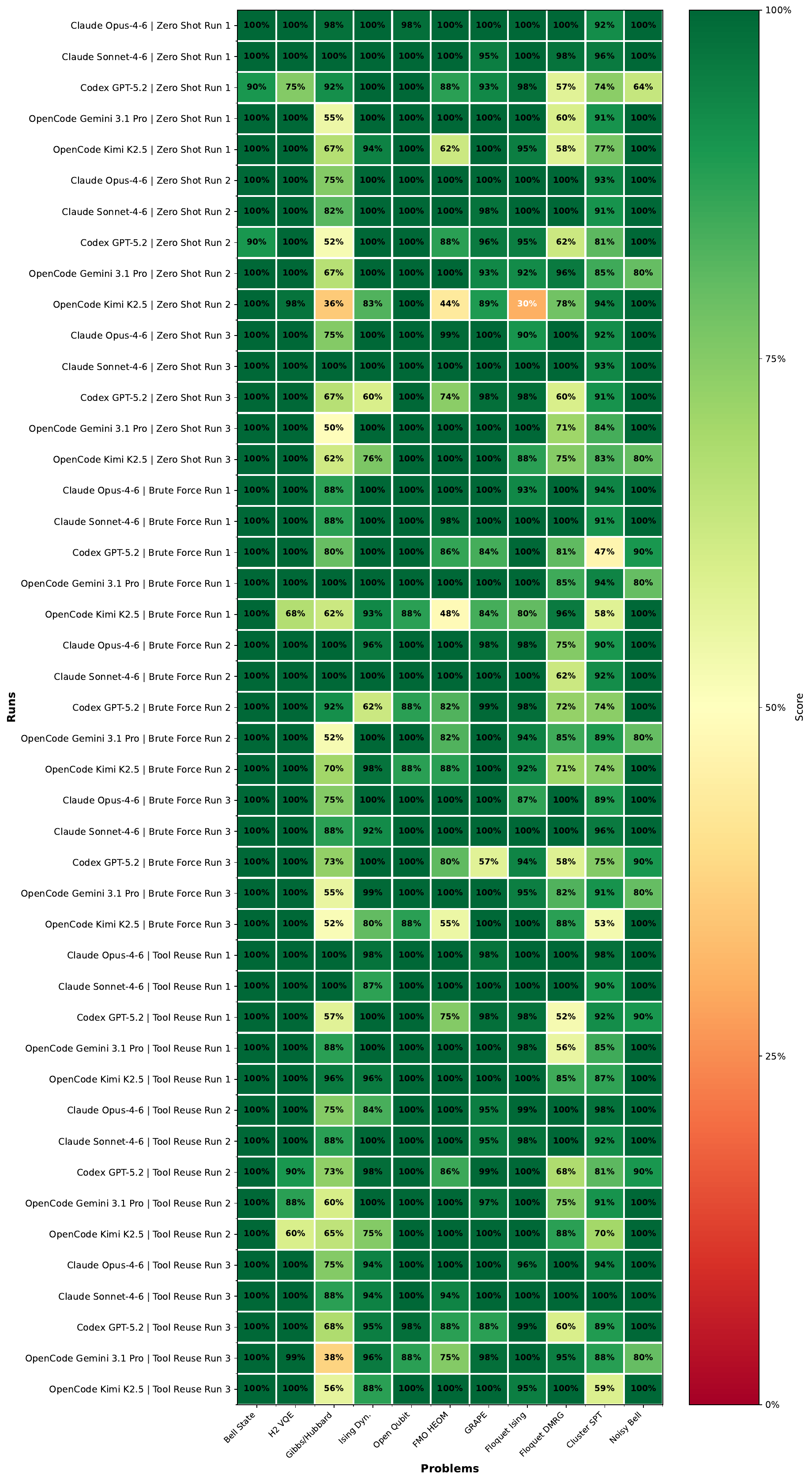}
    \caption{Quantum dynamics benchmark score heatmap. Rows correspond to model--condition--run combinations; columns correspond to the 11 quantum dynamics tasks. Colour scale: red (0\%) through yellow (50\%) to green (100\%).}
    \label{fig:heatmap_cuantico}
\end{figure}

Figures~\ref{fig:heatmap_q} and~\ref{fig:heatmap_cuantico} provide a visual overview of combined scores (accuracy + methodology, averaged over 3 runs) across all models, conditions, and tasks. Each cell is colour-coded on a red--yellow--green scale from 0\% to 100\%.

\vspace{-0.5em}

\section{Benchmark Evaluation Procedure}
\label{sec:benchmark-procedure}

Each agent run is evaluated along two independent axes, \emph{accuracy} and \emph{methodology}, using automated LLM-as-judge pipelines.
The procedure is identical across both benchmark suites with minor adaptations described below.

\paragraph{Accuracy evaluation.}
For the quantum chemistry tasks, the evaluator first extracts numerical results from the agent's output into a structured JSON file, which is then scored by a deterministic Python script that compares values against ground-truth references with appropriate tolerances.
For the quantum dynamics tasks, a separate Claude Code instance (with Claude Sonnet 4.6) reads the task specification and the agent's output files, then judges each predefined criterion as \texttt{true} or \texttt{false}.
The accuracy score is the fraction of satisfied criteria.

\paragraph{Methodology evaluation.}
For both benchmarks, each task is accompanied by a set of weighted methodology stages that describe the expected workflow (e.g.\ Hamiltonian construction, solver selection, post-processing).
A Claude Code-based evaluator (with Claude Sonnet 4.6) reads the agent's workspace and scores each stage on a 0--10 scale based on whether the agent followed correct domain methodology, independent of numerical accuracy.
The final methodology score is the weighted average across all stages.
Valid alternative approaches that achieve the same scientific objective receive full credit.
\section{Tool Merger}
\label{sec:tool-merger}

Over time, given enough different tasks, \EAF{} ends up creating sets of tools that are similar or identical in functionality. These excess tools negatively affect navigability and context efficiency. Therefore, we designed a \textit{tool merger}, a part of the toolset optimizer that merges and deletes existing tools to reduce clutter.

Tool merging follows a simple and efficient pipeline. Docstrings and function signatures of all tools are extracted and embedded before being sorted into clusters via embedding similarity. Identified clusters are then passed to Tool Merger, which decides if and how they should be merged. Tools that require merging are stored in a temporary backup directory, and new merged tools are drafted. These are then passed to the tool verification agent to ensure their correctness before deleting the original tools. If a merged tool fails verification, feedback is passed back to the tool merging agent along with the original tools to try again. Upon any re-failure, the original set is restored to the main directory. The tool structure optimizer is called following any tool merging to ensure optimal organization.

In this way the smallest fully functional tool set possible can be maintained to ensure optimal navigability and context efficiency.

Here we present and demonstration run in which the tool merger works on a set of tools created by solving the quantum chemistry tasks. Tool merger identified two similarity clusters containing six tools in total. The first cluster grouped four DFT tools into two merge pairs: the gas-phase and solvation variants of both single-point energy calculations and geometry optimizations. In each pair, neither tool was a strict superset of the other: the solvation variants added implicit solvent support (CPCM/DDCOSMO) but lacked features such as density fitting and auxiliary basis selection present in the base tools, while the geometry optimization pair additionally suffered from inconsistent input and output field names (\texttt{spin\_multiplicity} vs.\ \texttt{spin}, \texttt{energy} vs.\ \texttt{final\_energy}). Tool merger unified each pair into a single tool that makes solvation an optional parameter while standardizing all interfaces.

The second cluster paired a thermochemistry tool with a vibrational frequency tool. Both internally computed the molecular Hessian via the same \texttt{PySCF} call, but the thermochemistry tool returned only thermodynamic quantities (ZPE, enthalpy, Gibbs free energy, entropy) filtered to real positive frequencies, while the vibrational tool returned the full frequency spectrum including imaginary modes and normal-mode displacement vectors. The unified replacement computes the Hessian once and returns both the complete vibrational analysis and all thermodynamic properties, eliminating a redundant and expensive matrix computation.

All three unified tools passed automated review by a separate verification agent on the first attempt with zero errors or retries, reducing the tool inventory from 18 to 15. The remaining 12 tools were found to be sufficiently distinct from one another and were left unchanged. Original source files were preserved in a backup directory to allow rollback if needed.

\section{Tool File Edits During Execution}
\label{sec:tool-edits}

\EAF's iterative problem-solving loop gives the coding agent full write access to tool source files, allowing it to revise tools on the fly when execution results suggest a bug or a needed improvement. We expected this self-repair mechanism to be exercised frequently, yet our experiments show that in-place tool edits during execution were surprisingly rare for most setups.

Table~\ref{tab:tool-edits} reports how often tools were modified during execution. We distinguish two kinds of modification: \emph{pre-generated tool edits}, where the agent rewrites an existing tool file, and \emph{new tool creation}, where the agent writes an entirely new tool file that did not previously exist. In the zero-shot setting, Kimi~K2.5 (via OpenCode) created new tools in 29.9\% of its 144 iterations and edited pre-existing tools in 13.9\%, while Gemini~3.1~Pro (via OpenCode) created new tools in 30.6\% of 98 iterations and edited pre-existing ones in 12.2\%. GPT-5.2-Codex (via Codex) edited pre-existing tools in 12.8\% of iterations and created new ones in 4.5\%. By contrast, Opus~4.6 (via Claude Code) edited tools in only 6.0\% of iterations and Sonnet~4.6 in 1.2\%, with neither model ever creating new tools during execution. In the tool reuse setting, only Gemini~3.1~Pro and GPT-5.2-Codex edited tools by 3.9\% and 3.3\% of iterations, respectively; every other configuration left its tools untouched. These results suggest that the quality of initially generated tools differs greatly across models: stronger models produce tools that work correctly on first use, while weaker models must fall back on runtime patching, or even ad-hoc tool creation, to compensate for deficiencies in their initial implementations. The near-absence of edits in the tool reuse setting reinforces this interpretation, as the pre-existing library was generated by a stronger model (Opus~4.6) and thus requires little correction regardless of which model consumes it.

\begin{table}[H]
\centering
\caption{Tool file modifications during execution, reported per iteration. \emph{Edit Existing Tools}: iterations in which a pre-existing tool file was modified. \emph{Created New Tools}: iterations in which at least one new tool file was written.}
\label{tab:tool-edits}
\begin{tabular}{llcc}
\hline
\textbf{Model} & \textbf{Total Iters} & \textbf{Edit Existing Tools} & \textbf{Created New Tools} \\
\hline
\multicolumn{4}{l}{\emph{Zero-shot}} \\
\hline
Opus 4.6 (Claude Code)         & 84  & 5 (6.0\%)   & 0 (0\%)    \\
Sonnet 4.6 (Claude Code)       & 81  & 1 (1.2\%)   & 0 (0\%)    \\
GPT-5.2-Codex (Codex)          & 133 & 17 (12.8\%) & 6 (4.5\%)  \\
Gemini 3.1 Pro (OpenCode)      & 98  & 12 (12.2\%) & 30 (30.6\%) \\
Kimi K2.5 (OpenCode)           & 144 & 20 (13.9\%) & 43 (29.9\%) \\
\hline
\multicolumn{4}{l}{\emph{Tool reuse}} \\
\hline
Opus 4.6 (Claude Code)         & 81  & 0  & 0  \\
Sonnet 4.6 (Claude Code)       & 78  & 0  & 0  \\
GPT-5.2-Codex (Codex)          & 122 & 4 (3.3\%)  & 0  \\
Gemini 3.1 Pro (OpenCode)      & 102 & 4 (3.9\%)  & 0  \\
Kimi K2.5 (OpenCode)           & 112 & 0  & 0  \\
\hline
\end{tabular}
\end{table}

\section{Tool Generation and Usage in Zero-Shot Runs}
\label{sec:tool-usage}

\begin{figure}[H]
    \centering
    \includegraphics[width=\linewidth]{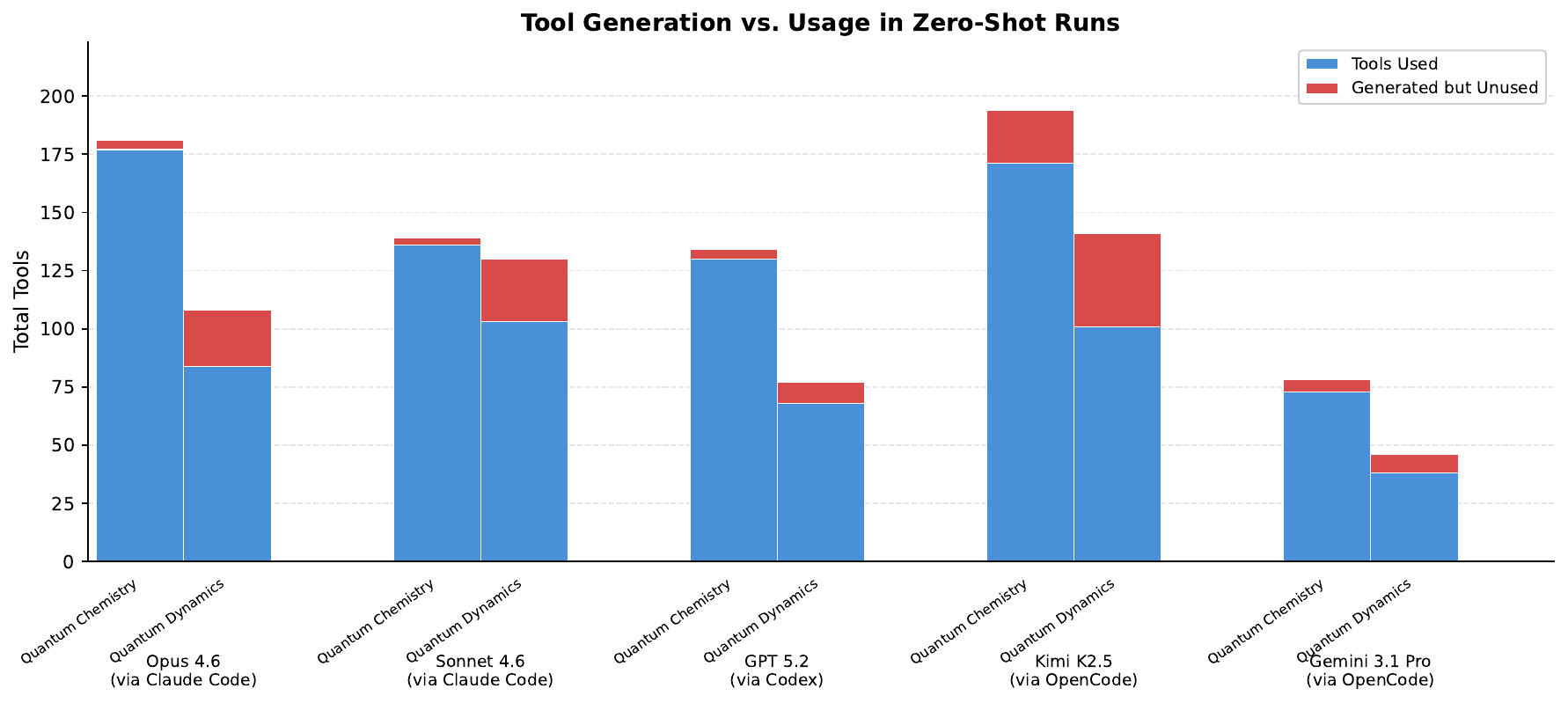}
    \caption{Tool generation versus usage in three zero-shot runs, broken down by model and question set. Blue bars indicate tools that were both generated and used; red segments indicate tools that were generated but never invoked.}
    \label{fig:tool-usage}
\end{figure}

Figure~\ref{fig:tool-usage} reports the total number of tools each model generated and invoked during zero-shot problem solving in the three runs of the two question sets.

Overall, agents use the large majority of the tools they create: aggregated use rates range from 81.2\% (Kimi~K2.5) to 93.8\% (GPT-5.2-Codex). Use rates are systematically higher on the quantum chemistry tasks (88--98\%) than on quantum dynamics tasks (72--88\%). Among the models, Kimi~K2.5 generates the most tools in total (335) but leaves the largest share unused (81.2\% use rate), while Gemini~3.1~Pro generates the fewest (124) yet maintains a comparable use rate (89.5\%). GPT-5.2-Codex achieves the highest overall use rate (93.8\%), suggesting that its tool generation strategy is the most targeted. Claude Opus~4.6 and Sonnet~4.6 fall between these extremes (90.3\% and 88.8\%, respectively), with both models showing a pronounced jump in use rate from quantum dynamics tasks ($\sim$78--79\%) to quantum chemistry tasks ($\sim$98\%). These results indicate that while all models occasionally generate tools they do not end up invoking, the waste is modest and concentrated in the simpler question set, where fewer tools are needed but agents generate them proactively.
\section{Tool Generation Guidelines}
\label{sec:generation-guidelines}

The following are example skill documents (generation guidelines) provided to the tool generator agent during our experiments. Each guideline encodes domain-specific best practices for a particular computational task. For El Cu\'antico tasks, the LLM often possesses broad knowledge of quantum computing concepts but tends to implement everything from scratch using \texttt{NumPy} rather than leveraging established domain-specific libraries. The DMRG and pulse optimization guidelines therefore steer the agent toward using dedicated packages (\texttt{TenPy} and \texttt{QuTiP}, respectively) that provide optimized, well-tested implementations. For quantum chemistry tasks, the agent generally knows which computational chemistry packages to use but sometimes lacks the domain knowledge needed to diagnose and resolve specific numerical issues. The vibrational frequency guideline, for example, provides a detailed reference on how to properly analyze imaginary frequencies and verify projection quality, knowledge that is critical for obtaining physically meaningful results but is not always reliably captured in the LLM's parametric knowledge.

\subsection{DMRG}

\prompt{
Always use \texttt{TenPy} to do DMRG.
}

\subsection{Pulse Optimization}

\prompt{
You should use the pulse optimizer provided by \texttt{qutip\_qtrl.pulseoptim}.
}

\subsection{Vibrational Frequency Calculation (PySCF)}

\prompt{
\textbf{Theory overview (what is being projected and why)}

For a molecule with $N$ atoms:

\begin{itemize}
    \item Total Cartesian degrees of freedom: $3N$
    \item Translational modes: \textbf{3}
    \item Rotational modes:
    \begin{itemize}
        \item \textbf{3} for non-linear molecules
        \item \textbf{2} for linear molecules
    \end{itemize}
    \item Vibrational modes:
    \begin{itemize}
        \item $3N - 6$ (non-linear)
        \item $3N - 5$ (linear)
    \end{itemize}
\end{itemize}

\textbf{Vibrational frequencies come from}

$$\mathbf{F} = \mathbf{M}^{-1/2} \mathbf{H} \mathbf{M}^{-1/2}$$

where:
\begin{itemize}
    \item $\mathbf{H}$: Cartesian Hessian (second derivatives of energy)
    \item $\mathbf{M}$: diagonal mass matrix
    \item $\mathbf{F}$: mass-weighted Hessian
\end{itemize}

The \textbf{translation and rotation subspace must be projected out} before diagonalization, otherwise you will see near-zero or imaginary frequencies.

PySCF provides this projection \textbf{automatically} if you use its harmonic analysis utilities correctly.

\textbf{Hessian calculation}

Choose the Hessian module consistent with your SCF type.

{\ttfamily\small
\noindent
from pyscf import gto, scf\\
from pyscf.hessian import rhf, thermo\\[6pt]
mol = gto.M(\\
\hspace*{2em}atom = \textquotesingle\textquotesingle\textquotesingle\\
\hspace*{2em}O 0 0 0\\
\hspace*{2em}H 0 -0.757 0.587\\
\hspace*{2em}H 0~~0.757 0.587\\
\hspace*{2em}\textquotesingle\textquotesingle\textquotesingle,\\
\hspace*{2em}basis = \textquotesingle cc-pvdz\textquotesingle\\
)\\[6pt]
mf = scf.RHF(mol).run()\\
hess = rhf.Hessian(mf).kernel()\\[6pt]
res = thermo.harmonic\_analysis(mol, hess)\\
print(res[\textquotesingle freq\_wavenumber\textquotesingle])
}

\begin{itemize}
    \item \texttt{hess} has shape \texttt{(3N, 3N)}
    \item Units: \textbf{Hartree / Bohr\textsuperscript{2}}
\end{itemize}

\textbf{Verifying projection quality (best practice)}

Check that:
\begin{itemize}
    \item No frequencies below ${\sim}10$ cm$^{-1}$ (except numerical noise)
    \item Number of modes equals:
    \begin{itemize}
        \item \texttt{3N - 6} (non-linear)
        \item \texttt{3N - 5} (linear)
    \end{itemize}
\end{itemize}

\textbf{Linear vs non-linear molecules}

PySCF \textbf{automatically detects linearity} from geometry and handles:
\begin{itemize}
    \item 5 projected modes for linear molecules
    \item 6 projected modes for non-linear molecules
\end{itemize}

No manual intervention needed unless geometry is ill-conditioned.

\textbf{Common pitfalls}

\begin{tabular}{lll}
\toprule
Issue & Cause & Fix \\
\midrule
Imaginary frequencies & Not optimized & Re-optimize geometry \\
Small non-zero translational modes & Loose SCF / grid & Tighten convergence \\
Wrong number of modes & Linear molecule misdetected & Check geometry \\
Large imaginary modes & Saddle point & Optimize further \\
\bottomrule
\end{tabular}
}

\section{Example Traces of Coding Agents}
\label{sec:trace}

\subsection{Case Study Question 1}
\prompt{
For the molecule ethylene (C=C), compute the first excited state energy using TD-DFT, and estimate the first excited state energy using a quantum algorithm.
Optimize the geometry using B3LYP/def2-SVP and compute the S$_1$ energy using TD-DFT.
Identify an active space capturing the $\pi$ and $\sigma$ system.
Estimate the first excited state energy using Quantum Subspace Expansion with CUDA-Q simulation.
Analyze the nature of the $\pi \to \pi^*$ transition, including limitations of TD-DFT versus quantum algorithms and the effect of correlation and active space choice. Make a visualization of important molecular orbitals.
}

\subsection*{Tool Analysis Report}

\subsubsection*{Task Analysis}

The task requires a multi-step workflow for ethylene:

\begin{enumerate}
  \item \textbf{Geometry generation \& optimization}: Use \texttt{smiles\_to\_3d\_geometry} to convert `C=C' SMILES to initial 3D coordinates, then use \texttt{run\_dft\_geometry\_optimization} with B3LYP/def2-SVP to get the optimized geometry.
  \item \textbf{TD-DFT excited state}: Use \texttt{run\_tddft\_calculation} on the converged DFT mean-field object to compute the S$_1$ excitation energy ($\pi \to \pi^*$ transition), getting excitation energies and oscillator strengths.
  \item \textbf{MO analysis}: Use \texttt{analyze\_molecular\_orbitals} to identify the HOMO ($\pi$) and LUMO ($\pi^*$) orbital indices and energies. Use the new \texttt{visualize\_molecular\_orbitals} tool to create contour plots of the $\pi$ and $\pi^*$ orbitals.
  \item \textbf{Active space setup}: Use \texttt{run\_pyscf\_casci} to validate the active space choice (e.g., 2 electrons in 2 orbitals for minimal $\pi/\pi^*$, or larger spaces including $\sigma/\sigma^*$). This provides a classical reference for the quantum algorithm.
  \item \textbf{Quantum Hamiltonian construction}: Use \texttt{tequila\_build\_molecular\_hamiltonian} with the chosen active space to generate the qubit Hamiltonian for ethylene.
  \item \textbf{Quantum Subspace Expansion}: Use the new \texttt{run\_quantum\_subspace\_expansion} tool with CUDA-Q to prepare the HF ground state, build excitation subspace, and compute the first excited state energy via QSE.
  \item \textbf{Comparison \& analysis}: Compare TD-DFT vs.\ QSE vs.\ CASCI energies. Analyze the $\pi \to \pi^*$ transition character, TD-DFT limitations, quantum algorithm advantages, and active space effects.
\end{enumerate}

\subsubsection*{Existing Tools (9)}

\begin{itemize}
  \item \texttt{smiles\_to\_3d\_geometry.py}
  \item \texttt{run\_dft\_geometry\_optimization.py}
  \item \texttt{run\_tddft\_calculation.py}
  \item \texttt{analyze\_molecular\_orbitals.py}
  \item \texttt{tequila\_build\_molecular\_hamiltonian.py}
  \item \texttt{tequila\_compute\_exact\_energy.py}
  \item \texttt{tequila\_run\_vqe.py}
  \item \texttt{visualize\_molecule.py}
  \item \texttt{plot\_line\_comparison.py}
\end{itemize}

\subsubsection*{New Tool Requirements (3)}

\begin{itemize}
  \item \texttt{visualize\_molecular\_orbitals} --- Generate 2D contour plots of molecular orbital wavefunctions from a converged PySCF mean-field object. Uses PySCF to evaluate MO wavefunctions on a grid, then takes a 2D slice and plots filled contour maps showing the orbital phase using matplotlib.
  \item \texttt{run\_quantum\_subspace\_expansion} --- Implement QSE for computing excited state energies using CUDA-Q. Constructs the QSE subspace by applying single and double excitation operators to the reference state, builds the Hamiltonian and overlap matrices, and solves the generalized eigenvalue problem.
  \item \texttt{run\_pyscf\_casci} --- Run a CASCI calculation using PySCF. Given a converged mean-field object and active space specification, performs CASCI to obtain ground and excited state energies within the active space.
\end{itemize}

\subsection*{iteration\_1 / Task Execution}

\begin{quote}
Let me start by understanding the problem and the workspace structure, then plan and execute the solution step by step.
\end{quote}

\begin{quote}
Let me explore the workspace and available tools first.
\end{quote}

\noindent\textbf{Step 1:} \texttt{Bash} --- `ls /`

\noindent\textbf{Step 2:} \texttt{Bash} --- `ls /tools/ 2>/dev/null || echo "no tools dir"`

\noindent\textbf{Step 3:} \texttt{Bash} --- `ls /tools/electronic\_structure\_dft\_calculations/ ...`

\noindent\textbf{Step 4:} \texttt{Read} --- `/tools/INDEX.md`

\begin{quote}
Let me read the key tools I'll need.
\end{quote}

\noindent\textbf{Step 5:} \texttt{Agent} --- \{"description": "Read all key tools", "subagent\_type": "Explore", "prompt": "Read the following files completely and report their function signatures, parameters, and return values:\textbackslash\{\}n\textbackslash\{\}n1. /...

\noindent\textbf{Step 6:} \texttt{Read} --- `/tools/molecular\_geometry\_visualization\_tools/molecular\_geometry\_processing/smiles\_to\_3d\_geometry.py`

\noindent\textbf{Step 7:} \texttt{Read} --- `/tools/electronic\_structure\_dft\_calculations/dft\_energy\_calculations/run\_dft\_geometry\_optimization.py`

\noindent\textbf{Step 8:} \texttt{Read} --- `/tools/electronic\_structure\_dft\_calculations/run\_tddft\_calculation.py`

\noindent\textbf{Step 9:} \texttt{Read} --- `/tools/electronic\_structure\_dft\_calculations/electronic\_property\_analysis/analyze\_molecular\_orbitals.py`

\noindent\textbf{Step 10:} \texttt{Read} --- `/tools/visualize\_molecular\_orbitals.py`

\noindent\textbf{Step 11:} \texttt{Read} --- `/tools/run\_pyscf\_casci.py`

\noindent\textbf{Step 12:} \texttt{Read} --- `/tools/run\_quantum\_subspace\_expansion.py`

\noindent\textbf{Step 13:} \texttt{Read} --- `/tools/electronic\_structure\_dft\_calculations/tequila\_molecular\_energy\_computation/tequila\_build\_molecular\_hamiltonian.py`

\noindent\textbf{Step 14:} \texttt{Read} --- `/tools/molecular\_geometry\_visualization\_tools/scientific\_data\_visualization\_tools/plot\_line\_comparison.py`

\begin{quote}
Now let me check what iterations/workspace content exists already.
\end{quote}

\noindent\textbf{Step 15:} \texttt{Bash} --- `ls /iterations/ 2>/dev/null; ls /...`

\noindent\textbf{Step 16:} \texttt{Bash} --- `mkdir -p /\{logs,img\}`

\begin{quote}
Now let me create the scripts. I'll break this into logical phases:

\begin{enumerate}
\item \textbf{script\_1.py} - Geometry generation, optimization, MO analysis, TD-DFT
\item \textbf{script\_2.py} - CASCI and Quantum Subspace Expansion
\item \textbf{script\_3.py} - Visualization and report generation
\end{enumerate}

Let me read the tequila tool more carefully for the Hamiltonian building part, and also the QSE tool.
\end{quote}

\noindent\textbf{Step 17:} \texttt{Read} --- `/tools/run\_quantum\_subspace\_expansion.py`

\noindent\textbf{Step 18:} \texttt{Read} --- `/tools/electronic\_structure\_dft\_calculations/tequila\_molecular\_energy\_computation/tequila\_build\_molecular\_hamiltonian.py`

\begin{quote}
Now let me create the scripts. I'll start with script\_1.py for geometry optimization, MO analysis, TD-DFT, and CASCI.
\end{quote}

\noindent\textbf{Step 19:} \texttt{Write} --- `/script\_1.py` (8,116 chars)

\begin{quote}
Now let me load the MCP tool schema and run script\_1.
\end{quote}

\noindent\textbf{Step 20:} \texttt{ToolSearch} --- \{"query": "select:mcp\_submit\_job", "max\_results": 1\}

\noindent\textbf{Step 21:} \texttt{mcp\_submit\_job} --- \{"workspace\_dir": "", "step": \{"command": "python script\_1.py"\}\}

\noindent\textbf{Step 22:} \texttt{Read} --- `/logs/20260414\_163936\_script\_1.out`

\noindent\textbf{Step 23:} \texttt{Bash} --- `grep -E "\textasciicircum{}(===|---|HOMO|LUMO|S1|T1|Ground|State|Active|Optimized energy|Converged|MO |  S|  T|  MO)" ...`

\begin{quote}
Script 1 succeeded. Now let me create script\_2 for the quantum subspace expansion using tequila + CUDA-Q.
\end{quote}

\noindent\textbf{Step 24:} \texttt{Write} --- `/script\_2.py` (4,956 chars)

\noindent\textbf{Step 25:} \texttt{mcp\_submit\_job} --- \{"workspace\_dir": "", "step": \{"command": "python script\_2.py"\}\}

\noindent\textbf{Step 26:} \texttt{Bash} --- `tail -30 /logs/20260414\_164059\_script\_2.out`

\noindent\textbf{Step 27:} \texttt{Bash} --- `tail -30 /logs/20260414\_164059\_script\_2.err`

\begin{quote}
The tequila Hamiltonian iteration is different. Let me check how to convert.
\end{quote}

\noindent\textbf{Step 28:} \texttt{Write} --- `/test\_tequila.py` (765 chars)

\noindent\textbf{Step 29:} \texttt{mcp\_submit\_job} --- \{"workspace\_dir": "", "step": \{"command": "python test\_tequila.py"\}\}

\noindent\textbf{Step 30:} \texttt{Read} --- `/logs/20260414\_164233\_test\_tequila.out`

\begin{quote}
Now I see the format. The keys are tuples of qubit-pauli pairs, and values are coefficients. Let me use \texttt{paulistrings} instead.
\end{quote}

\noindent\textbf{Step 31:} \texttt{Write} --- `/test\_tequila.py` (969 chars)

\noindent\textbf{Step 32:} \texttt{mcp\_submit\_job} --- \{"workspace\_dir": "", "step": \{"command": "python test\_tequila.py"\}\}

\noindent\textbf{Step 33:} \texttt{Bash} --- `tail -5 /logs/20260414\_164254\_test\_tequila.err`

\noindent\textbf{Step 34:} \texttt{Write} --- `/test\_tequila.py` (834 chars)

\noindent\textbf{Step 35:} \texttt{mcp\_submit\_job} --- \{"workspace\_dir": "", "step": \{"command": "python test\_tequila.py"\}\}

\noindent\textbf{Step 36:} \texttt{Read} --- `/logs/20260414\_164312\_test\_tequila.out`

\begin{quote}
Now I understand the format. \texttt{naked()} returns a string like \texttt{Z(0)}, \texttt{X(0)Z(1)X(2)}, and \texttt{coeff} is the coefficient. Let me fix script\_2.
\end{quote}

\noindent\textbf{Step 37:} \texttt{Edit} --- `/script\_2.py`

\noindent\textbf{Step 38:} \texttt{mcp\_submit\_job} --- \{"workspace\_dir": "", "step": \{"command": "python script\_2.py"\}\}

\noindent\textbf{Step 39:} \texttt{Read} --- `/logs/20260414\_164333\_script\_2.out`

\begin{quote}
Good, QSE is working. The CAS(2,2) gives 4.88 eV first excitation which is close to the CASCI(2,2) value of 4.29 eV. Note the nuclear repulsion energy looks odd (-75.9 Ha) --- that's likely because tequila with active\_orbitals folds core energy into the constant. The CAS(4,4) gives 2.29 eV which seems low. The sto-3g basis is quite small; let me note this is expected behavior for a minimal basis.

Now let me create script\_3 for visualization and report generation.
\end{quote}

\noindent\textbf{Step 40:} \texttt{Write} --- `/script\_3.py` (11,697 chars)

\noindent\textbf{Step 41:} \texttt{mcp\_submit\_job} --- \{"workspace\_dir": "", "step": \{"command": "python script\_3.py"\}\}

\noindent\textbf{Step 42:} \texttt{Read} --- `/logs/20260414\_164457\_script\_3.out`

\begin{quote}
All three scripts completed successfully. Let me fix the image paths in the report since the visualize tool used different naming than expected.
\end{quote}

\noindent\textbf{Step 43:} \texttt{Edit} --- `/script\_3.py`

\noindent\textbf{Step 44:} \texttt{mcp\_submit\_job} --- \{"workspace\_dir": "", "step": \{"command": "python script\_3.py"\}\}

\begin{quote}
Let me verify the images and report are in place.
\end{quote}

\noindent\textbf{Step 45:} \texttt{Bash} --- `ls /img/`

\begin{quote}
Let me check the MO images look reasonable.
\end{quote}

\noindent\textbf{Step 46:} \texttt{Read} --- `/img/mo\_xz\_7.png`

\noindent\textbf{Step 47:} \texttt{Read} --- `/img/mo\_xz\_8.png`

\begin{quote}
The MO plots look excellent --- MO 7 (HOMO) shows the classic $\pi$ bonding orbital with one nodal plane, and MO 8 (LUMO) shows the $\pi$* antibonding with two lobes of opposite sign on each carbon. Let me check the comparison plot and verify the report.
\end{quote}

\noindent\textbf{Step 48:} \texttt{Read} --- `/img/method\_comparison.png`

\begin{quote}
Everything looks good. All three scripts ran successfully:

\textbf{Summary of results:}

\begin{enumerate}
\item \textbf{Script 1} - Geometry optimization (B3LYP/def2-SVP), MO analysis, TD-DFT, and CASCI
\end{enumerate}
   - S$_1$ ($\pi$$\rightarrow$$\pi$*) via TD-DFT: \textbf{8.15 eV} (f=0.37)
   - HOMO-LUMO gap: 7.66 eV
   - CASCI(2,2)/(4,4)/(6,6) first excitation (T$_1$): \textasciitilde{}4.29 eV

\begin{enumerate}
\item \textbf{Script 2} - Tequila qubit Hamiltonian + QSE with CUDA-Q
\end{enumerate}
   - QSE(2,2): \textbf{4.88 eV} first excitation (4 qubits)
   - QSE(4,4): \textbf{2.29 eV} first excitation (8 qubits, 53-dim subspace)
\ldots{} (+391 chars)
\end{quote}

\subsection*{iteration\_1 / Evaluation}

\begin{quote}
Let me read the question and report files.
\end{quote}

\noindent\textbf{Step 1:} \texttt{Read} --- `/question.md`

\noindent\textbf{Step 2:} \texttt{Read} --- `/report.md`

\begin{quote}
Let me check if the key files exist.
\end{quote}

\noindent\textbf{Step 3:} \texttt{Glob} --- `*.py` in ``

\noindent\textbf{Step 4:} \texttt{Glob} --- `img/*` in ``

\begin{quote}
The report and all referenced files exist. Let me verify the key requirements against the report:

\begin{enumerate}
\item \textbf{Geometry optimization (B3LYP/def2-SVP)} - Done $\checkmark$
\item \textbf{S$_1$ energy via TD-DFT} - Done (8.15 eV) $\checkmark$
\item \textbf{Active space identification ($\pi$ and $\sigma$ system)} - Done (CAS(2,2), CAS(4,4), CAS(6,6)) $\checkmark$
\item \textbf{QSE with CUDA-Q simulation} - Done $\checkmark$
\item \textbf{Analysis of $\pi$$\rightarrow$$\pi$* transition, TD-DFT vs quantum limitations, correlation/active space effects} - Done $\checkmark$
\item \textbf{MO visualization} - Done, images exist $\checkmark$
\end{enumerate}

The rep\ldots{} (+230 chars)
\end{quote}

\noindent\textbf{Step 5:} \texttt{Write} --- `/evaluation.json` (225 chars)

\begin{quote}
Created \texttt{evaluation.json}. The report satisfies all task requirements:

\begin{itemize}
\item \textbf{Geometry optimization} at B3LYP/def2-SVP completed
\item \textbf{TD-DFT S$_1$ energy} computed (8.15 eV, f=0.375)
\item \textbf{Active space} identified with CAS(2,2), (4,4), (6,6) covering $\pi$/$\sigma$ systems
\item \textbf{QSE with CUDA-Q} performed for CAS(2,2) and CAS(4,4)
\item \textbf{Analysis} covers $\pi$$\rightarrow$$\pi$* nature, TD-DFT limitations, quantum algorithm advantages, and active space/correlation effects
\item \textbf{MO visualizations} present (8 orbital images + method com
\end{itemize}\ldots{} (+13 chars)
\end{quote}

\subsection{Case Study Question 2}
\prompt{Model a \textbf{rubidium-87} hyperfine qubit as a two-level open quantum system. Compute the ground-state electron density at the nucleus using \textbf{PySCF}, construct a two-level Hamiltonian ($H = (\omega/2)\,Z$) from it, and simulate Lindblad dynamics with \textbf{QuTiP}. Study the evolution from (a) the excited state ($|1\rangle$) and (b) a superposition ($(|0\rangle+|1\rangle)/\sqrt{2}$), then plot the excited-state population and coherence over time and provide a brief analysis. Your plots should be publication-quality.}

\subsection*{Tool Analysis Report}

\begin{quote}
The task requires three main computational steps:

\textbf{Step 1: PySCF electron density computation.}
Run a DFT calculation on a rubidium atom (Rb, $Z=37$) using \texttt{run\_dft\_single\_point.py}. Then use the new \texttt{compute\_electron\_density\_at\_nuclei} tool to evaluate $|\psi(0)|^2$ at the Rb nucleus. This contact density feeds into the Fermi contact hyperfine interaction, from which the hyperfine splitting frequency $\omega$ (${\sim}6.835\,\text{GHz}$) can be derived. The new tool is general-purpose and reusable for any hyperfine coupling, NMR chemical shift, or M\"{o}ssbauer spectroscopy calculation.

\textbf{Step 2: Simulate Lindblad dynamics with QuTiP.}
Use \texttt{qutip\_lindblad\_evolution.py} to simulate the two-level system $H = (\omega/2)\,\sigma_z$ with collapse operators modeling T$_1$ decay ($\sqrt{\gamma_1}\,\sigma_-$) and pure dephasing ($\sqrt{\gamma_\varphi/2}\,\sigma_z$). Run two simulations: (a) starting from $|e\rangle$ and (b) starting from $(|e\rangle+|g\rangle)/\sqrt{2}$.

\textbf{Step 3: Visualization.}
Use \texttt{plot\_line\_comparison.py} to create publication-quality plots showing excited-state population $P_e(t)$ and coherence $|\rho_{01}(t)|$ as functions of time for both initial conditions, overlaying analytical solutions for validation.
\end{quote}

\subsubsection*{Existing Tools (4)}

\begin{itemize}
  \item \texttt{run\_dft\_single\_point.py} (from quantum chemistry)
  \item \texttt{run\_single\_point\_hf.py} (from quantum chemistry)
  \item \texttt{qutip\_lindblad\_evolution.py} (from quantum dynamics)
  \item \texttt{plot\_line\_comparison.py} (from quantum dynamics)
\end{itemize}

\subsubsection*{New Tool Requirements (1)}

\begin{itemize}
  \item \texttt{compute\_electron\_density\_at\_nuclei} --- Compute the total electron density at nuclear positions from a converged PySCF SCF object. Uses the converged density matrix and evaluates basis functions at the specified nuclear coordinates via \texttt{pyscf.dft.numint.eval\_ao} and \texttt{eval\_rho}. Supports both restricted (RHF/RKS) and unrestricted (UHF/UKS) SCF objects.
\end{itemize}

\subsection*{iteration\_1 / Task Execution}

\begin{quote}
Let me understand the problem and plan my approach.

I need to:
\begin{enumerate}
\item Compute electron density at Rb nucleus using PySCF
\item Use that to derive hyperfine frequency (6.834 GHz for Rb-87)
\item Simulate Lindblad dynamics with QuTiP for two initial states
\item Plot results
\end{enumerate}

Let me first check what tools are available in the tools/ directory.
\end{quote}

\noindent\textbf{Step 1:} \texttt{Bash} --- `ls tools/ 2>/dev/null`

\noindent\textbf{Step 2:} \texttt{Bash} --- `pwd \&\& ls `

\noindent\textbf{Step 3:} \texttt{Bash} --- `ls tools/molecular\_electronic\_structure\_tools/ \&\& echo "--...`

\noindent\textbf{Step 4:} \texttt{Bash} --- `ls tools/qutip\_quantum\_dynamics\_tools/qutip\_quantum\_dynami...`

\noindent\textbf{Step 5:} \texttt{Read} --- `tools/compute\_electron\_density\_at\_nuclei.py`

\noindent\textbf{Step 6:} \texttt{Read} --- `tools/qutip\_quantum\_dynamics\_tools/qutip\_quantum\_dynamics\_evolution/qutip\_lindblad\_evolution.py`

\noindent\textbf{Step 7:} \texttt{Read} --- `tools/molecular\_visualization\_analysis\_tools/scientific\_data\_visualization\_tools/plot\_line\_comparison.py`

\noindent\textbf{Step 8:} \texttt{Read} --- `tools/molecular\_electronic\_structure\_tools/dft\_energy\_calculations/run\_dft\_single\_point.py`

\begin{quote}
Now I have all the tools. Let me create the scripts.
\end{quote}

\noindent\textbf{Step 9:} \texttt{Write} --- `script\_1.py` (10,377 chars)

\begin{quote}
Now let me run it.
\end{quote}

\noindent\textbf{Step 10:} \texttt{ToolSearch} --- \{"query": "select:mcp\_submit\_job", "max\_results": 1\}

\noindent\textbf{Step 11:} \texttt{mcp\_submit\_job} --- \{"workspace\_dir": "/case\_study\_q2\_take7", "step": \{"command": "python script\_1.py"\}\}

\noindent\textbf{Step 12:} \texttt{Bash} --- `cat logs/20260415\_213934\_script\_1.err`

\begin{quote}
cc-pVDZ doesn't support Rb. Let me use a different basis set.
\end{quote}

\noindent\textbf{Step 13:} \texttt{Edit} --- `script\_1.py`

\noindent\textbf{Step 14:} \texttt{mcp\_submit\_job} --- \{"workspace\_dir": "/case\_study\_q2\_take7", "step": \{"command": "python script\_1.py"\}\}

\noindent\textbf{Step 15:} \texttt{Bash} --- `cat logs/20260415\_213944\_script\_1.out | grep -E "\textasciicircum{}(=|Part|...`

\begin{quote}
Everything succeeded. Let me check the plots and write the report.
\end{quote}

\noindent\textbf{Step 16:} \texttt{Read} --- `img/population\_decay.png`

\noindent\textbf{Step 17:} \texttt{Read} --- `img/coherence\_decay.png`

\begin{quote}
The plots look good but I notice the numerical curves for the $|1\rangle$ initial case don't match the analytical ones. The issue is that the analytical solution for population starting from $|1\rangle$ should decay, but the numerical shows it staying at 1 then decaying. Wait --- looking more carefully, the numerical $|1\rangle$ curve (solid blue) stays near 1 while analytical (dashed blue) decays. That's wrong. The problem is the Hamiltonian convention --- \texttt{sigmaz} has $|0\rangle$ as +1 eigenstate and $|1\rangle$ as -1, so \texttt{basis(2,1)} i\ldots{} (+746 chars)
\end{quote}

\noindent\textbf{Step 18:} \texttt{Edit} --- `script\_1.py`

\noindent\textbf{Step 19:} \texttt{Edit} --- `script\_1.py`

\begin{quote}
Now update the analytical labels to match:
\end{quote}

\noindent\textbf{Step 20:} \texttt{Edit} --- `script\_1.py`

\begin{quote}
Wait, let me think about the analytical solution more carefully. For the superposition state ($|0\rangle$+$|1\rangle$)/$\sqrt{2}$, the initial excited population is 0.5. Under T1 decay, Pe(t) = 0.5 * exp(-t/T1). Let me fix that.
\end{quote}

\noindent\textbf{Step 21:} \texttt{Edit} --- `script\_1.py`

\begin{quote}
Also update the plot labels:
\end{quote}

\noindent\textbf{Step 22:} \texttt{Edit} --- `script\_1.py`

\noindent\textbf{Step 23:} \texttt{Edit} --- `script\_1.py`

\noindent\textbf{Step 24:} \texttt{mcp\_submit\_job} --- \{"workspace\_dir": "/case\_study\_q2\_take7", "step": \{"command": "python script\_1.py"\}\}

\noindent\textbf{Step 25:} \texttt{Read} --- `img/population\_decay.png`

\noindent\textbf{Step 26:} \texttt{Read} --- `img/coherence\_decay.png`

\begin{quote}
The plots look correct now --- numerical and analytical curves overlap perfectly. Now let me write the report.
\end{quote}

\noindent\textbf{Step 27:} \texttt{Write} --- `report.md` (5,001 chars)

\begin{quote}
The solution is complete. Here's a summary:

\textbf{Script}: \texttt{script\_1.py} --- runs all three phases in one script.

\textbf{Results}:
\begin{enumerate}
\item \textbf{PySCF DFT} (B3LYP/def2-SVP): Computed electron density at Rb nucleus = 2.655 e/bohr\textasciicircum{}3. The computed Fermi contact coupling overestimates the experimental value because it includes all core electrons, not just the valence 5s electron. The experimental hyperfine splitting (6.835 GHz) is used for dynamics.
\end{enumerate}

\begin{enumerate}
\item \textbf{Lindblad dynamics} (T1=10, T2=5 in units of 1/omega):
\end{enumerate}
   -\ldots{} (+280 chars)
\end{quote}

\subsection*{iteration\_1 / Evaluation}

\begin{quote}
Let me read the question and report files.
\end{quote}

\noindent\textbf{Step 1:} \texttt{Read} --- `question.md`

\noindent\textbf{Step 2:} \texttt{Read} --- `report.md`

\noindent\textbf{Step 3:} \texttt{Glob} --- `*.py` in `/case\_study\_q2\_take7`

\noindent\textbf{Step 4:} \texttt{Glob} --- `img/*` in `/case\_study\_q2\_take7`

\noindent\textbf{Step 5:} \texttt{Glob} --- `intermediate\_*.json` in `/case\_study\_q2\_take7`

\begin{quote}
Let me check the script and outputs to verify correctness.
\end{quote}

\noindent\textbf{Step 6:} \texttt{Read} --- `script\_1.py`

\noindent\textbf{Step 7:} \texttt{Read} --- `intermediate\_1.json`

\noindent\textbf{Step 8:} \texttt{Read} --- `intermediate\_2.json`

\begin{quote}
Let me verify the plots exist and check them visually.
\end{quote}

\noindent\textbf{Step 9:} \texttt{Read} --- `img/population\_decay.png`

\noindent\textbf{Step 10:} \texttt{Read} --- `img/coherence\_decay.png`

\begin{quote}
Everything checks out. The task is fully complete.

\textbf{Summary of evaluation:}

\begin{itemize}
\item \textbf{PySCF electron density}: Computed correctly (2.655 e/bohr³), SCF converged, with proper discussion of why total density overestimates hyperfine coupling
\item \textbf{Hamiltonian}: Correctly constructed as H = ($\omega$/2)$\sigma$\_z using experimental Rb-87 frequency
\item \textbf{Lindblad dynamics}: Both initial states (excited and superposition) simulated with correct collapse operators (T1 relaxation + pure dephasing)
\item \textbf{Plots}: Publicati
\end{itemize}\ldots{} (+215 chars)
\end{quote}

\noindent\textbf{Step 11:} \texttt{Write} --- `evaluation.json` (225 chars)

\begin{quote}
\texttt{evaluation.json} created. All requirements are met --- the report covers PySCF electron density computation, Lindblad dynamics for both initial states, publication-quality plots with analytical comparison, and thorough physical analysis.
\end{quote}

\subsection{Tool Generation Trace: \texttt{run\_quantum\_subspace\_expansion}}

The following trace illustrates the tool generation and iterative review process for the QSE tool forged during Case Study~1. The tool generator (Claude Sonnet~4.6 via Claude Code) first explores the CUDA-Q API, implements the tool, and submits it for automated review. The reviewer identifies three algorithmic bugs, which are fixed and re-verified in a second review iteration.

\subsection*{Tool Generation Iter 1}

\begin{quote}
Let me start by exploring the cudaq library to understand its API for state simulation and Pauli operators.
\end{quote}

\noindent\textbf{Step 1:} \texttt{ToolSearch} --- \{"query": "select:TodoWrite", "max\_results": 1\}

\begin{quote}
Let me explore the cudaq library to understand its API.
\end{quote}

\noindent\textbf{Step 2:} \texttt{TodoWrite} --- \{"todos": [\{"content": "Search for cudaq APIs relevant to QSE", "status": "in\_progress", "activeForm": "Searching for cudaq APIs"\}, \{"content": "Plan implementation", "status": "pending", "activeForm"...

\noindent\textbf{Step 3:} \texttt{Read} --- `/.venv/lib/python3.11/site-packages/.maestra/indexes/cudaq.md`

\noindent\textbf{Step 4:} \texttt{Grep} --- `class State|def get\_state|from\_data|StateData` in `/.venv/lib/python3.11/site-packages/cudaq`

\noindent\textbf{Step 5:} \texttt{Bash} --- `cd  \&\& python -c "import cudaq; help(cudaq.State)" 2>\&1 | head -60`

\noindent\textbf{Step 6:} \texttt{Bash} --- `cd  \&\& python -c "
import cudaq
\# Check spin operator API
print('=== spin ===')...`

\noindent\textbf{Step 7:} \texttt{Bash} --- `cd  \&\& python -c "
import cudaq
\# Check observe and get\_state
help(cudaq.observ...`

\noindent\textbf{Step 8:} \texttt{Bash} --- `cd  \&\& python -c "
import cudaq
import numpy as np

\# Test basic state creation...`

\noindent\textbf{Step 9:} \texttt{Bash} --- `cd  \&\& python -c "
import cudaq
import numpy as np

\# Test overlap
sv1 = np.arr...`

\begin{quote}
Good. Now I have everything I need. The approach:

\begin{enumerate}
\item Build Hamiltonian as \texttt{cudaq.SpinOperator} from Pauli terms
\item Create reference state via \texttt{cudaq.State.from\_data()}
\item Generate excitation operators (single/double) as Pauli strings, apply them to reference state as matrices
\item Build H and S matrices using \texttt{state.overlap()} and matrix-vector products
\item Solve generalized eigenvalue problem with \texttt{scipy.linalg.eigh}
\end{enumerate}
\end{quote}

\noindent\textbf{Step 10:} \texttt{TodoWrite} --- \{"todos": [\{"content": "Search for cudaq APIs relevant to QSE", "status": "completed", "activeForm": "Searching for cudaq APIs"\}, \{"content": "Plan implementation", "status": "completed", "activeForm"...

\noindent\textbf{Step 11:} \texttt{Write} --- `/tool\_smith/task\_0f1ba6682313/run\_quantum\_subspace\_expansion.py` (12,413 chars)

\begin{quote}
Let me verify it imports and runs:
\end{quote}

\noindent\textbf{Step 12:} \texttt{Bash} --- `cd  \&\& python -c "
from case\_study\_q1.tool\_smith.task\_0f1ba6682313.run\_quantum\_...`

\begin{quote}
Works correctly. Now write the report.
\end{quote}

\noindent\textbf{Step 13:} \texttt{Write} --- `/tool\_smith/task\_0f1ba6682313/report.md` (1,962 chars)

\noindent\textbf{Step 14:} \texttt{TodoWrite} --- \{"todos": [\{"content": "Search for cudaq APIs relevant to QSE", "status": "completed", "activeForm": "Searching for cudaq APIs"\}, \{"content": "Plan implementation", "status": "completed", "activeForm"...

\begin{quote}
Done. The tool \texttt{run\_quantum\_subspace\_expansion.py} and \texttt{report.md} are saved. Key points:

\begin{itemize}
\item \textbf{Pydantic I/O models} with full validation (\texttt{PauliTerm}, \texttt{ExcitationOperator}, input/output models)
\item \textbf{Auto-generates} single and double qubit excitation operators from occupied/virtual orbital indices when not provided
\item \textbf{Builds subspace} by applying excitation operator matrices to the reference statevector
\item \textbf{Solves} the generalized eigenvalue problem \texttt{Hc = ESc} via canonical orthogonalization
\end{itemize}\ldots{} (+230 chars)
\end{quote}

\subsection*{Tool Review Iter 1}

\begin{quote}
Let me read the tool code first.
\end{quote}

\noindent\textbf{Step 1:} \texttt{Read} --- `/tool\_smith/task\_0f1ba6682313/run\_quantum\_subspace\_expansion.py`

\begin{quote}
Let me review the algorithm:

\begin{enumerate}
\item \textbf{Basis vectors not normalized}: Line 264-266 checks norm > 1e-12 but doesn't normalize. The comment on line 257 says "(normalized)" but normalization isn't done. This is actually fine for QSE - you don't need to normalize because you solve the generalized eigenvalue problem H\textit{c = E}S*c which handles non-orthonormal bases.
\end{enumerate}

\begin{enumerate}
\item \textbf{Using \texttt{.real} on overlap matrix}: Line 287 \texttt{np.linalg.eigh(S\_proj.real)} - this discards imaginary parts. For a Hermitian matrix, the
\end{enumerate}\ldots{} (+1152 chars)
\end{quote}

\begin{quote}
I see several algorithm issues. The main problems are discarding imaginary parts of complex matrices and using transpose instead of conjugate transpose. Let me fix them.
\end{quote}

\noindent\textbf{Step 2:} \texttt{Edit} --- `/tool\_smith/task\_0f1ba6682313/run\_quantum\_subspace\_expansion.py`

\begin{quote}
Now let me write and run a minimal test:
\end{quote}

\noindent\textbf{Step 3:} \texttt{Bash} --- `python -c "
import numpy as np

\# Test: 2-qubit Heisenberg-like Hamiltonian H = 0.5*ZI - 0.5*IZ + 0.25*XX + 0.25*YY
\# Ex...`

\begin{quote}
The eigenvectors are now complex (correct!), but the output model expects \texttt{list[list[float]]}. I need to output the real parts of eigenvectors (they're defined up to a phase anyway), or better, output their magnitudes. Let me fix the output conversion:
\end{quote}

\noindent\textbf{Step 4:} \texttt{Edit} --- `/tool\_smith/task\_0f1ba6682313/run\_quantum\_subspace\_expansion.py`

\noindent\textbf{Step 5:} \texttt{Bash} --- `python -c "
import numpy as np
I2 = np.eye(2, dtype=complex)
X = np.array([[0,1],[1,0]], dtype=complex)
Y = np.array([[0...`

\begin{quote}
The QSE correctly finds the ground and excited state energies. The subspace is only 2D (reference + one single excitation) for this 2-qubit system, so it captures 2 of the 4 eigenvalues, which is expected.

Now let me also verify the eigenvalues are exact for this case --- -1.118 and 1.118 match the exact values perfectly.

Let me also quickly check that the eigenvector output conversion doesn't lose critical sign info. The eigenvectors from \texttt{eigh} on the real-valued \texttt{H\_orth} will be real, but \texttt{X}\ldots{} (+305 chars)
\end{quote}

\noindent\textbf{Step 6:} \texttt{Write} --- `/tool\_smith/task\_0f1ba6682313/review\_iter\_1.json` (1,413 chars)

\begin{quote}
Review complete. Three algorithm correctness bugs were found and fixed:

\begin{enumerate}
\item \textbf{\texttt{S\_proj.real} $\rightarrow$ \texttt{S\_proj}}: The overlap matrix eigendecomposition was discarding imaginary parts, producing wrong canonical orthogonalization when basis vectors are complex (which happens whenever Y-gate excitation operators are applied).
\end{enumerate}

\begin{enumerate}
\item \textbf{\texttt{H\_proj.real} $\rightarrow$ \texttt{H\_proj}}: Same issue for the projected Hamiltonian matrix.
\end{enumerate}

\begin{enumerate}
\item \textbf{\texttt{X.T} $\rightarrow$ \texttt{X.conj().T}}: The canonical orthogonalization transform must use conjugate transpose
\end{enumerate}\ldots{} (+161 chars)
\end{quote}

\subsection*{Tool Review Iter 2}

\begin{quote}
Let me read the tool code first.
\end{quote}

\noindent\textbf{Step 1:} \texttt{Read} --- `/tool\_smith/task\_0f1ba6682313/run\_quantum\_subspace\_expansion.py`

\begin{quote}
The previous review fixed three issues. Let me verify they're addressed:

\begin{enumerate}
\item \texttt{np.linalg.eigh(S\_proj)} - Yes, line 287, using full complex matrix $\checkmark$
\item \texttt{X.conj().T @ H\_proj @ X} - Yes, line 298 $\checkmark$
\item \texttt{.real} on eigenvector output - Yes, line 313 $\checkmark$
\end{enumerate}

Now let me check for remaining algorithm issues:

\begin{enumerate}
\item The basis vectors are NOT normalized before building H\_proj and S\_proj (line 266 - no normalization). This is fine - the overlap matrix S handles non-orthonormal bases. The generalized eigenvalue probl
\end{enumerate}\ldots{} (+867 chars)
\end{quote}

\begin{quote}
The previous fixes are all properly applied. Let me run a quick test to verify correctness.
\end{quote}

\noindent\textbf{Step 2:} \texttt{Bash} --- Test QSE tool correctness: `python -c "
import numpy as np
from run\_quantum\_subspace\_expansion import *

\# 2...`

\begin{quote}
The QSE finds 2 of 4 eigenvalues correctly (subspace is only dimension 2 with 1 occupied + 1 virtual orbital, so singles-only gives 2 states). The energies match exact diagonalization. The code is correct.
\end{quote}

\noindent\textbf{Step 3:} \texttt{Write} --- `/tool\_smith/task\_0f1ba6682313/review\_iter\_2.json` (25 chars)

\begin{quote}
The tool is approved. All three previously identified issues have been properly fixed:
\begin{enumerate}
\item \texttt{np.linalg.eigh(S\_proj)} uses the full complex overlap matrix (line 287)
\item \texttt{X.conj().T @ H\_proj @ X} uses conjugate transpose (line 298)
\item Eigenvector output uses \texttt{.real} for the float-typed model (line 313)
\end{enumerate}

Verified with a 2-qubit Hamiltonian --- QSE energies match exact diagonalization to machine precision.
\end{quote}


\end{document}